\documentclass[runningheads]{llncs}

\usepackage[mobile]{eccv}

\usepackage{eccvabbrv}
\usepackage{graphicx}
\usepackage{booktabs}
\usepackage[accsupp]{axessibility}  %

\usepackage{microtype}
\usepackage{multirow} 
\usepackage{tcolorbox}

\definecolor{bronze}{rgb}{1, 1, 0.75}
\definecolor{silver}{rgb}{1, 0.875, 0.75}
\definecolor{gold}{rgb}{1, 0.75, 0.75}

\newcommand{\tgold}[1]{\colorbox{gold}{{#1}}}
\newcommand{\tsilver}[1]{\colorbox{silver}{{#1}}}
\newcommand{\tbronze}[1]{\colorbox{bronze}{{#1}}}

\usepackage{xcolor}
\usepackage{colortbl}
\newcommand{\gold}[1]{\cellcolor{red!25}{#1}}
\newcommand{\silver}[1]{\cellcolor{orange!25}{#1}}
\newcommand{\bronze}[1]{\cellcolor{yellow!25}{#1}}

\newcommand{\method}{{UMBRAE}} %
\newcommand{\methodSubject}[1]{{\method{}-#1}}
\newcommand{\bench}{{BrainHub}}
\newcommand{\methodColor}{\textcolor{violet}{\method}}
\newcommand{\benchColor}{\textcolor{olive}{\bench}}

\newcommand{\condparagraph}[1]{\vspace{0.25em}\noindent\textbf{#1}\enspace}

\newcommand{\revise}[1]{\textcolor{black}{#1}}

\usepackage[breaklinks,colorlinks,citecolor=eccvblue]{hyperref}

\usepackage{orcidlink}

\begin{document}

\title{UMBRAE: Unified Multimodal Brain Decoding}

\titlerunning{UMBRAE}

\author{Weihao Xia\inst{1} \quad
Raoul de Charette\inst{2} \quad
Cengiz Oztireli\inst{3}\quad
Jing-Hao Xue\inst{1} 
}

\authorrunning{W.~Xia et al.}

\institute{$^{1}$University College London  \ \ \ \ $^{2}$Inria \ \ \ \ $^{3}$University of Cambridge}

\maketitle

\begin{abstract}

We address prevailing challenges of the brain-powered research, departing from the observation that the literature hardly recover accurate spatial information and require subject-specific models.
To address these challenges, we propose UMBRAE, a unified multimodal decoding of brain signals.
First, to extract instance-level conceptual and spatial details from neural signals, we introduce an efficient universal brain encoder for multimodal-brain alignment and recover object descriptions at multiple levels of granularity from subsequent multimodal large language model (MLLM).
Second, we introduce a cross-subject training strategy mapping subject-specific features to a common feature space.
This allows a model to be trained on multiple subjects without extra resources, even yielding superior results compared to subject-specific models. Further, we demonstrate this supports weakly-supervised adaptation to new subjects, with only a fraction of the total training data.
Experiments demonstrate that \method{} not only achieves superior results in the newly introduced tasks but also outperforms methods in well established tasks. 
To assess our method, we construct and share with the community a comprehensive brain understanding benchmark \bench{}. Our code and benchmark are available at \protect{\small \url{https://weihaox.github.io/UMBRAE}}.

\keywords{Multimodal Brain Decoding \and Universal Brain Encoder \and Cross-Subject Training \and Weakly-Supervised Adaptation 
}
\end{abstract}

\begin{figure}[!th]
\vspace{-2.9em}
	\centering
	\includegraphics[width=\linewidth]{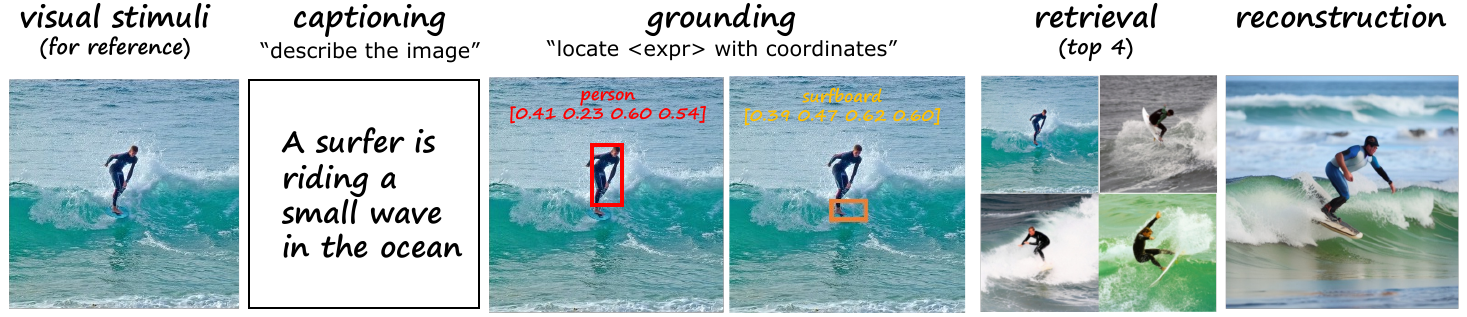}\vspace{-.3em}
	\caption{\textbf{\textbf{Multimodal Decoding.}} By aligning brain features with MLLMs, \method{} decodes multimodal cues from brain signals, which allows multiple downstream tasks.}
	\label{fig:teaser}
\vspace{-3.1em}
\end{figure}

\section{Introduction}
\label{sec:intro}

Typically, artificial intelligence research relies on intermediate modalities to interpret human intentions, such as language~\cite{touvron2023llama,chen2023shikra}, gaze~\cite{admoni2017social,xia2020controllable}, facial expression~\cite{cui2021empathic}, and action~\cite{hussein2017imitation}.
These modalities, however, are indirect channels of communication with humans and may be highly inaccurate for people with cognitive or physical disabilities or even locked-in patients who are conscious but unable to communicate through speech, limb, or facial movements~\cite{laureys2005locked}.
In this context, the potential for direct interpretation of neural signals stands out as a promising prospect.
The brain imaging literature has recently advanced, decoding neural signals into various forms such as image~\cite{scotti2023reconstructing}, video~\cite{chen2024cinematic}, or text~\cite{tang2023semantic} to read intentions~\cite{chaudhary2022spelling,lee2024noir}. 
This deepens our understanding of the brain which neural activity is not directly comprehensible to humans. 

However, there are remaining challenges in brain-powered research. First, decoding brain signals into a single modality results in a \revise{lossy representation} of the brain activities. On the one hand, text fails to preserve the peculiar appearance of a texture or the spatial location of an object. On the other hand, visual decoding~\cite{xia2023dream,ozcelik2023brain,scotti2023reconstructing} addresses the underdetermined problem of pixel-wise reconstruction and lacks explicitation of the scene structure.
Consider, for instance, the scenario where a person uses thoughts to control a robotic arm to retrieve an apple from a fruit bowl on a table.
The first task is to recognize the apple amidst similar visual concepts and then locate its exact position. 
But current methods lack such fine-grained decoding capability to interpret object categories, visual concepts, and their relationships.
The second challenge pertains to the subject-specific patterns of brain activities~\cite{allen2022massive}. 
Therefore, current methods typically train a model for each subject to cope with distinctive brain patterns. Decoding brain signals across multiple subjects presents challenges due to the structural and functional differences among individual brains.

Hence, we instead propose to decode a robust multimodal representation which serves as proxy for downstream tasks, such as textual or visual decoding. 
Our method allows brain decoding at different granularities, through prompting, which unravels unprecedented brain-machine interface for locked-in patients~\cite{laureys2005locked} that typically requires iterative feedback.
To evaluate our novel tasks, we extend the popular Natural Scenes Dataset~(NSD)~\cite{allen2022massive} with multimodal ground truth, which constitutes a new brain understanding benchmark. 
Both code and benchmark will be made publicly available.
Our contributions summarize as follows:

\begin{itemize}
	\item We introduce \methodColor, aiming at
     \underline{u}nified \underline{m}ultimodal \underline{bra}in d\underline{e}coding.
 Our method relies on a universal brain encoder and a frozen multimodal large language model seeking to align brain signals with images. We also propose a cross-subject training strategy to learn a universal representation across subjects, as opposed to the standard subject-specific training. Furthermore, it allows the novel weakly-supervised adaptation, enabling the training of a model for any arbitrary subject with minimal training data.

	\item We construct \benchColor, a multimodal \underline{brain} \underline{u}nderstanding \underline{b}enchmark extending NSD~\cite{allen2022massive}. The benchmark pairs fMRI with semantic concepts and spatial localization in visual stimuli, offering tasks and metrics for evaluation.
 
    \item Our method achieves better or on par performance compared to state-of-the-art methods on a variety of tasks including brain captioning, retrieval, and visual decoding. It is also the first one to enable direct brain grounding, performing on par with natural baselines while being at least 10 times faster.
\end{itemize}

\section{Related Works}
\label{sec:related_works}

\condparagraph{Brain-Conditioned Generation.}
Generative vision models conditioned on brain signals~\cite{ozcelik2023brain,takagi2023high,xia2023dream,scotti2023reconstructing,lin2022mind} have recently achieved unparalleled performance in decoding visual stimuli from corresponding brain responses.
Generally, these methods map brain responses, captured in the form of functional magnetic resonance imaging (fMRI), to more common modalities suitable for feeding into pretrained vision-language models~\cite{karras2020analyzing,rombach2022high,xu2022versatile} for subsequent image reconstruction.
For example, Lin~\etal~\cite{lin2022mind} project fMRI data to a CLIP~\cite{radford2021learning} common space and reconstruct images through a finetuned StyleGAN2~\cite{karras2020analyzing}.
Takagi and Nishimoto~\cite{takagi2023high} utilize the ridge regression to link fMRI signals with CLIP text embeddings and the latent space of Stable Diffusion~\cite{rombach2022high} (SD). 
Xia~\etal~\cite{xia2023dream} extract semantics, depth, and color cues, and reconstruct images using a depth-color-conditioned SD. 
Rather than relying solely on textual embeddings, several methods~\cite{han2024onellm,takagi2023improving} aim to obtain explicit descriptions for the visual stimuli.
{In contrast, our method decodes brain responses into various human-readable textual and visual cues, which can also flexibly serve as inputs for generative models.}

\condparagraph{Multimodal Large Language Models.} 
Expanding Large Language Models (LLMs) to encompass other modalities, such as images, has garnered considerable attention recently.
These models typically comprise three components: a frozen image encoder, a trainable adapter, and a frozen or finetuned LLM. The adapter's role is to bridge the gap from image features to the LLM, which can be implemented as a linear layer~\cite{chen2023shikra}, a multilayer perceptron (MLP)~\cite{liu2023visual}, or a lightweight transformer~\cite{jaegle2021perceiver}. 
In addition to vision-focused LLMs, recent studies aim to expand the boundaries of LLMs to include other modalities, making it possible to unify multiple modalities within a single LLM.
Brain signals, as an emerging modality, have also recently been incorporated, for example in OneLLM~\cite{han2024onellm}, but like training with other modalities, all these methods require massive amounts of data and abundant computational resources. {In contrast, we demonstrate a simple yet effective way to align brain signals with images.}

\condparagraph{Multimodal-Brain Alignment.} 
The prevailing practice for brain alignment is to map the neural modality into a common latent space~\cite{scotti2023reconstructing,ozcelik2023brain,xia2023dream,han2024onellm}, which can be divided into two lines of works: discriminative alignment and generative alignment. 
Considering the scarcity of data, methods in the first category aligns the brain modality within a pretrained embedding space, such as CLIP \cite{radford2021learning}, through direct regression~\cite{takagi2023improving,takagi2023high,ozcelik2023brain}, contrastive learning \cite{xia2023dream}, or diffusion prior \cite{scotti2023reconstructing}.
The second is garnering significant attention in the field of MLLMs. For instance, OneLLM~\cite{han2024onellm} adopts generative training to learn the alignment of multimodal inputs, including brain signals, thereby connecting a universal encoder with an LLM.
However, such alignments between brain signals and images or text are  trained per subject, resulting in one model for each subject. {In contrast, we align the brain modality with image features to recover both semantic and spatial cues and achieve cross-subject multimodal-brain alignment to leverage user diversity.}

\begin{figure}[t]
	\centering
	\includegraphics[width=0.95\linewidth]{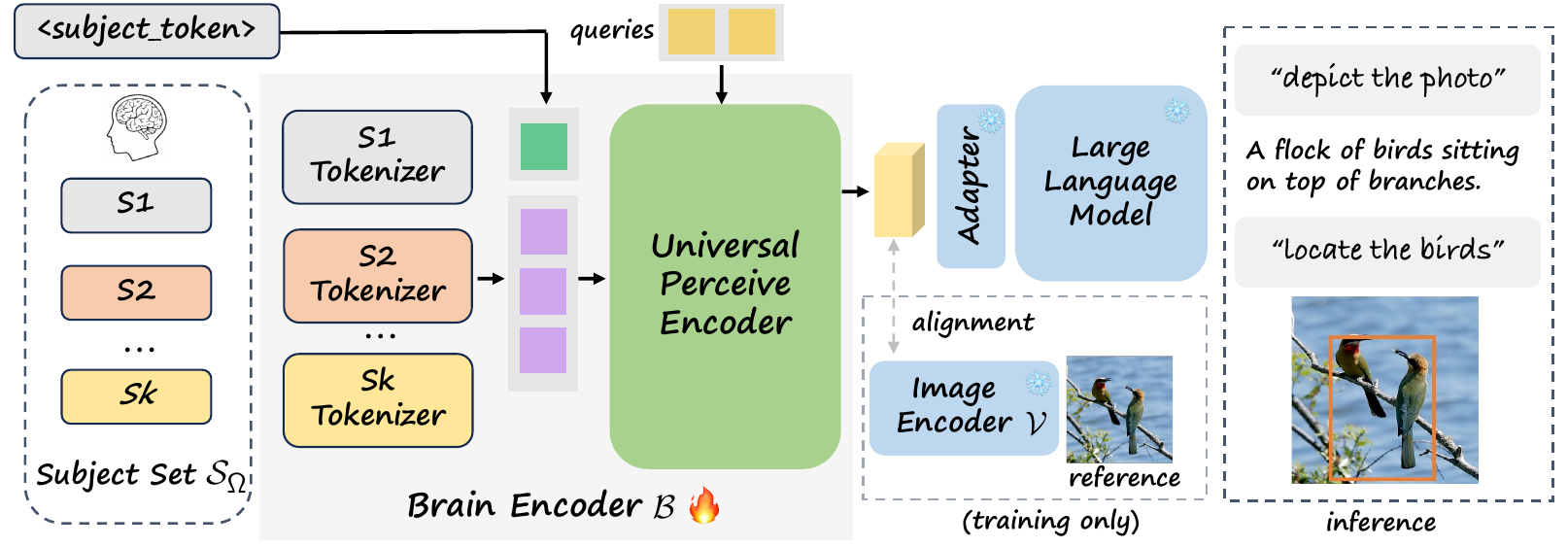}
	\caption{\textbf{Overview of \method{}.}
		Our brain encoder includes subject-specific tokenizers and a universal perceive encoder (\cref{subsec:model_arch}).
		Neural signals (fMRI) from multiple subjects 
        are mapped into a common feature space, enabling cross-subject training and weakly-supervised adaptation (\cref{subsec:training_strategy}).
		The brain encoder learns to align neural signals with image features (\cref{subsec:binding_feature}). 
		During inference, the learned encoder interacts with MLLMs and performs brain understanding tasks according to given prompts (\cref{subsec:inference}).
	}
	\label{fig:model_arch}
\end{figure}

\section{\method{}}
\label{subsec:method}

Our method is designed to address two shortcomings of the brain decoding literature. 
First, instead of learning a unimodal decoding (text or image), we learn a brain encoder that aligns brain features with pretrained multimodal space thus benefiting from all the MLLMs downstream tasks for multimodal decoding. 
Second, we observe that prior works train per-subject models owing to neuroscience showing inter-subject variability of the brain activities~\cite{allen2022massive}. Instead, we intuit that inter-subject patterns present and learn a subject-unified representation by training jointly across-subjects. Beside better performance, this allows adaptation to novel subjects with minimal training data.

The overview of our method, named \method{}, is in~\cref{fig:model_arch}. 
{The acronym stands for \underline{u}nified \underline{m}ultimodal \underline{bra}in d\underline{e}coding
and signifies the process of unveiling encoded information hidden within the `shadows' of brain signals.}
We rely mainly on a flexible brain encoder architecture (\cref{subsec:model_arch}) and a cross-subject training strategy (\cref{subsec:training_strategy}) to map brain responses from different subjects, each with variable length, into a common feature space.
In our experiments, we observe that simply binding brain modality with image features enables the recovery of both semantic and spatial cues (\cref{subsec:binding_feature}).
We then conduct brain prompting interface by inputting the standardized brain representations from diverse subjects into MLLMs for downstream tasks (\cref{subsec:inference}).

\subsection{Architecture}
\label{subsec:model_arch}

\revise{Our brain encoder, based on a lightweight transformer architecture~\cite{vaswani2017attention,jaegle2021perceiver}, accommodates variable-length brain response inputs.
This is important for cross-subject training as fMRI data} are variable across subjects. 
Hence, our architecture comprises subject-specific tokenizers which aim is to extract and map the subject-specific characteristics, along with a universal perceive encoder designed to capture subject-agnostic knowledge, later aligned with image features.

\condparagraph{Subject Tokenizer.} 
The subject-specific tokenizer projects the input brain signal $s \in \mathbb{R}^{1 \times L_{s}}$ selected from the subject set $\mathcal{S}_{\Omega}$, with arbitrary length $L_{s}$,
into a fixed-length sequence of brain tokens $\mathbf{x} \in \mathbb{R}^{L \times D}$. Here, $L$ is the sequence length and $D$ is the token dimension. 
Considering inter-subject variability in brain patterns~\cite{allen2022massive}, we design a separate tokenizer for each subject.
Besides, we introduce a learnable subject-specific token 
$\left\{\mathbf{s}_k\right\}_{k=1}^K$
to switch between subjects, where $K$ is the total number of subjects and
$\mathbf{s}_k \in \mathbb{R}^{M \times D}$ contains $M$ tokens of dimension $D$. 
Then, we prepend subject-specific tokens $\mathbf{s}_k$ to the predicted brain tokens $\mathbf{x}$
and encode them with the following universal perceive encoder.

\condparagraph{Universal Perceive Encoder.}
The universal perceive encoder seeks to project all brain tokens $\mathbf{x}$ into a common space.
We utilize here a transformer-based architecture~\cite{jaegle2021perceiver} which uses cross-attention modules to project the input tokens into a latent bottleneck where
the key $K$ and value $V$ are projections of the input tokens, while $Q$ is the projection of learnable latent queries. 

The subject-specific tokenizers are expected to capture specific information for each subject, including structural and functional differences among individual brains; and the universal perceive encoder aims to extract common knowledge across different subjects, encompassing categories, semantics, textures, and geometries of various objects and scenes. 
We now detail the training strategy for cross-subject alignment.

\subsection{Cross-Subject Alignment}
\label{subsec:training_strategy}
 For cross-subject alignment, it is crucial to ensure that examples from each subject are uniformly sampled. This enables the model to avoid subject preference and prevent catastrophic forgetting.
Therefore, we adopt a sampling strategy to ensure that $\theta$ percent of samples in a batch are from the same subject.
Considering $\mathcal{S}_{\Omega}$ being the union over $K$ subjects training data $\mathcal{S}_{\Omega}=\bigcup_{k \in\{1,2, \ldots, K\}} \mathcal{S}_k$, we select data samples from a subject $\mathcal{S}_k$ with probability:
\begin{equation}
   p_k=\frac{\left\|\mathcal{S}_k\right\|}{\sum_{n=1}^K\left\|\mathcal{S}_n\right\|},
\end{equation}
where $\|\cdot\|$ denotes cardinality (\ie, the number of data samples).
To construct a batch with $B$ number of data samples, we select a subject $S_k$ with probability $p_k$ and conduct random sampling to yield $\theta \times B$ training samples. The remaining $(1-\theta) \times B$ examples are uniformly sampled from other subjects, \ie, $\mathcal{S}_{\Omega} \backslash \mathcal{S}_k$.
This batch sampling strategy significantly benefits from user diversity as it allows the model to focus primarily on intra-subject training while being exposed to different subjects to improve inter-subject discrimination.
We latter demonstrate that this \revise{cross-subject alignment} enhances performance without incurring extra training costs compared to training with a single subject, at no additional training time.

\subsection{Multimodal Alignment}
\label{subsec:binding_feature}

Previous multimodal methods learn to map multiple modalities into a common latent space, typically through contrastive pretraining, using either images~\cite{girdhar2023imagebind} or text~\cite{han2024onellm} as the binding modality. 
\revise{Instead}, we align brain representations with image features from a pretrained image encoder using element-wise reconstruction.

Given a brain response $s \in \mathbb{R}^{1 \times L_s}$ and the corresponding visual stimulus $v \in \mathbb{R}^{W \times H \times C}$, the source brain encoder $\mathcal{B}$ and target visual image encoder $\mathcal{V}$ encode brain signals and images into features denoted $\mathbf{b}$ and $\mathbf{v}$, respectively.
We train the brain encoder $\mathcal{B}$ to minimize the distance between brain features and image features, aiming for a close approximation $\mathcal{B}(b) \approx \mathcal{V}(v) $ through:
\begin{equation}
\left.\mathcal{L}_\text{rec}=\mathbb{E}_{\mathbf{b} \sim \mathbf{B}, \mathbf{v} \sim \mathbf{V}}[\| \mathcal{V}(v)-\mathcal{B}(b)) \|_2^2\right].  
\end{equation}

Our brain encoder $\mathcal{B}$ learns the alignment between source brain space $\mathbf{B}$ and target image space $\mathbf{V}$. Different from previous methods, we align the brain signals with the intermediate image features from a pretrained image encoder, thus achieving semantic and spatial alignment for the brain representation.
\revise{MLLMs~\cite{chen2023shikra,liu2023visual} show such features provide sufficient visual cues for finetuning LMMs. Furthermore, aligning with intermediate image features allows direct input of aligned brain representations into the MLLM.}

\subsection{Brain Prompting Interface}
\label{subsec:inference}

After alignment, brain features from the brain encoder $\mathcal{B}$ are fed into the MLLM's adapter, to retrieve the mapped visual embeddings \texttt{<image>}.
These embeddings are then concatenated with a user instruction prompt and inputted into the finetuned LLM.
Thus, our brain encoder inherits from multimodal capabilities of the MLLM, allowing tasks to be used in a prompting fashion using template:

\begin{tcolorbox}[left=0.2em, right=0.2em, top=0.2em, bottom=0.2em,colback=white,colframe=black!30!white]
	\centering
	\small
	\texttt{system message.}
	user: \textcolor{VioletRed}{\texttt{<instruction>}} \texttt{<image>} 
	assistant: \textcolor{DarkOrchid}{\texttt{<answer>}}
\end{tcolorbox}
\noindent{}The tags \textcolor{VioletRed}{\texttt{<instruction>}} and \textcolor{DarkOrchid}{\texttt{<answer>}} serve as placeholders for human instructions and assistant answers.
We use variable templates for different tasks. Specifically, brain captioning uses `\texttt{Describe this image <image> as simply as possible.}'; brain grounding, `\texttt{Locate <expr> in <image> and provide its coordinates, please.}', where \texttt{<expr>} is the expression. More templates for different supported tasks can be found in the appendix.%

Our primary focus in this study is on brain captioning and grounding, which reflects the capabilities of brain signals in concept recognition and spatial localization.
They are often referred to as image captioning and visual grounding in the multimodal learning literature.
However, in this context, brain signals are used as the input rather than images.
Our method also supports other instruction-following capabilities, such as conversation, detail description, and complex reasoning. 
Our method is model-agnostic, allowing for the use of any image encoders, LLMs, and MLLMs according to specific needs.

\section{Experiments}

\subsection{Implementation Details}

\condparagraph{Architecture.} We use the pretrained CLIP ViT-L/14~\cite{radford2021learning} as the visual encoder and Vicuna-7B/13B~\cite{chiang2023vicuna} as the LLM, consistent with the setup in Shikra~\cite{chen2023shikra} and LLaVA~\cite{liu2023visual}. 
The target image features are obtained from the second last layer of the transformer encoder, denoted as $\mathbf{T} \in \mathbb{R}^{16 \times 16 \times 1024}$, which are then converted to $\mathbf{T}^{\prime} \in \mathbb{R}^{256 \times D}$ for further processing by the adapter and LLM. The dimension $D$ is 4,096 for Vicuna-7B and 5,120 for Vicuna-13B.
The learnable tokens for each subject are of dimensions $\mathbb{R}^{5 \times 1024}$.

\condparagraph{Training Details.} 
Our models are trained on a single A100 GPU for 240 epochs with a global batch size of 256. It takes around 12 hours to complete.
We use AdamW~\cite{loshchilov2017decoupled} as the optimizer with $\beta_1=0.9$, $\beta_2=0.95$, and weight decay of $0.01$.
For the learning rate scheduler, we use one-cycle~\cite{smith2019super} with an initial learning rate of 3e-4. We set $\theta=0.5$, meaning that in each batch of 256 samples, 128 come from each of two subjects. The selection probabilities are identical for each subject, as they contain the same number of training data.
Following visual decoding studies \cite{takagi2023high,scotti2023reconstructing,xia2023dream}, we use the standard train and test splits for the four subjects (S1, S2, S5, S7). Specifically, each subject contains 24,980 training samples. For testing, we report the average of the three same-image repetitions, totaling 982 samples per subject. 
Note that the above studies train a subject-specific model for each of the four subjects, while we train one brain encoder for them all.

\subsection{\bench}
\label{subsec:benchmark}
For evaluation, we construct a multimodal brain understanding benchmark, \bench{}, to further analyze the 
information contained in brain signals.
Specifically, we extend the NSD~\cite{allen2022massive}, a popular dataset comprising brain responses of subjects viewing visual stimuli (images) sourced from  Microsoft Common Objects in Context (COCO)~\cite{lin2014microsoft}.
NSD provides (fMRI, image) pairs which is sufficient for visual decoding. 
However, we aim to explore the ability to process brain signals for identifying visual concepts, recognizing and localizing instances, as well as extracting spatial relationships among multiple exemplars.
Specifically, we process the corresponding COCO images for each fMRI sample and extract relevant labels for the following tasks and metrics:  

\begin{itemize}
    \item \textbf{Brain Captioning} aims at  textually describing the primary content of a given brain response.
    Ground truth captions are retrieved from COCO~\cite{lin2014microsoft}, and evaluation of inferred captions uses five standard metrics: BLEU-$k$~\cite{papineni2002bleu}, METEOR~\cite{banerjee2005meteor}, ROUGE-L~\cite{lin2004rouge}, CIDEr~\cite{vedantam2015cider}, and SPICE~\cite{anderson2016spice}, in addition to two CLIP-based scores~\cite{radford2021learning}, namely CLIP-S and RefCLIP-S~\cite{hessel2021clipscore}. %
    \item \textbf{Brain Grounding} is the counterpart of visual grounding~\cite{chen2023shikra,liu2023visual} and seeks to recover spatial locations from brain signals by inferring coordinates. 
    Given that identified classes might be named differently, or simply absent from ground truth labels, we evaluate bounding boxes through the task of referring expression comprehension~\cite{yu2016modeling}, using accuracy and intersection over union~(IoU) as the evaluation metrics.
    \item \textbf{Brain Retrieval} is to search for pertinent results in response to a provided query from a large database, often considered as a form of fine-grained, instance-level classification. 
    The evaluation metric used is accuracy.
    \item \textbf{Visual Decoding} refers to the capability to reconstruct the visual stimuli associated with the fMRI data. We include it here for consistency with the extensive literature on visual decoding~\cite{scotti2023reconstructing,ozcelik2023brain}.
\end{itemize}

\setlength{\tabcolsep}{4pt}
\setlength{\fboxrule}{0pt} 
\setlength{\fboxsep}{2pt}
\begin{table}[t]
\caption{\textbf{Brain Captioning}. `\methodSubject{S1}' refers to our model trained with a single subject (S1 here) only, while `\method{}' denotes the model with cross-subject training.
`Shikra-w/img' refers to the image captioning result from Shikra~\cite{chen2023shikra} using the ground truth image as input, serving as an approximate upper bound.
The colors represent the \tgold{best}, \tsilver{second-best}, and \tbronze{third-best} performance. Brain captioning results for the other subjects are provided in the appendix. 
}
\label{tab:brain_captioning}
\vspace{-2.5mm}
\centering
\resizebox{\textwidth}{!}{
\begin{tabular}{@{}lcccccccccc@{}}
\toprule
Method & BLEU1 & BLEU2 & BLEU3 & BLEU4 & METEOR & ROUGE & CIDEr  & SPICE  & CLIP-S & RefCLIP-S\\
\midrule
Shikra-w/img~\cite{chen2023shikra} & 82.38 & 69.90 & 58.63 & 49.66 & 35.60 & 65.49 & 161.43 & 27.62 & 80.60   & 85.92  \\
\midrule
SDRecon~\cite{takagi2023high}        & 36.21 & 17.11 & 7.72 & 3.43 & 10.03 & 25.13  & 13.83 & 5.02 & 61.07 & 66.36   \\
OneLLM~\cite{han2024onellm} & 47.04 & 26.97 & 15.49 & 9.51 & 13.55 & 35.05  & 22.99 & 6.26 & 54.80    & 61.28   \\
UniBrain~\cite{mai2023unibrain} &  -  & -  & - & - & \bronze{16.90}  & 22.20  &  - & -  & -  &  -  \\
BrainCap~\cite{ferrante2023brain} & \bronze{55.96} & \bronze{36.21} & \bronze{22.70} & \bronze{14.51} & 16.68 & \bronze{40.69} & \bronze{41.30} & \bronze{9.06} & \bronze{64.31} & \bronze{69.90} \\
\midrule
\methodSubject{S1}   & \silver{57.63} & \silver{38.02} & \silver{25.00}  & \silver{16.76}   & \silver{18.41}   &  \silver{42.15}  & \silver{51.93} & \silver{11.83} & \silver{66.44} & \silver{72.12}   \\
\method{}  & \gold{59.44} & \gold{40.48} & \gold{27.66} & \gold{19.03} & \gold{19.45} & \gold{43.71}  & \gold{61.06} & \gold{12.79} &  \gold{67.78} & \gold{73.54}  \\
\bottomrule
\end{tabular}
}
\end{table}
\setlength{\tabcolsep}{1.4pt}

\subsection{Brain Captioning}
\label{subsec:captioning}

\cref{tab:brain_captioning} provides an evaluation of our brain captioning for subject 1 (S1), with respect to SOTA baselines being SDRecon~\cite{takagi2023improving}, BrainCap~\cite{ferrante2023brain} and OneLLM~\cite{han2024onellm}.
From the latter table, \method{} outperforms all baselines by a significant margin on all metrics. 
SDRecon poor performance 
results from its limited limited vocabulary, and the use redundant or meaningless words in its captioning, such as `\textit{person and person with person person wearing a tie shirt person person, women's clothing.}', which impacts the quality metrics negatively.
BrainCap~\cite{ferrante2023brain} follows a similar pipeline but replaces the captioning model, which performs better.
OneLLM~\cite{han2024onellm} learns a unified encoder for multimodal-text alignment which improves the caption quality but deteriorates the CLIP similarity score, as it merely aligns with texts.
In contrast, %
the alignment with image features of \method{} preserves more accurate semantic and spatial cues decoded from the brain signals. Moreover, the use of LLMs helps generate sentences that are fluent, complete, and rich in information. 
Interestingly, we note that the performance of \method{} (trained on S1, S2, S5, S7) exceeds those when trained only on data from S1 (\methodSubject{S1}), demonstrating the ability to learn from cross-subject patterns.
As a approximate upper bound, we also report `Shikra-w/img' which, similar to us, utilizes Shikra~\cite{chen2023shikra} for captioning though here using the ground truth image (visual stimuli). 
Results for other subjects are provided in the appendix.

\begin{figure}[t]
    \centering
    \includegraphics[width=0.95\linewidth]{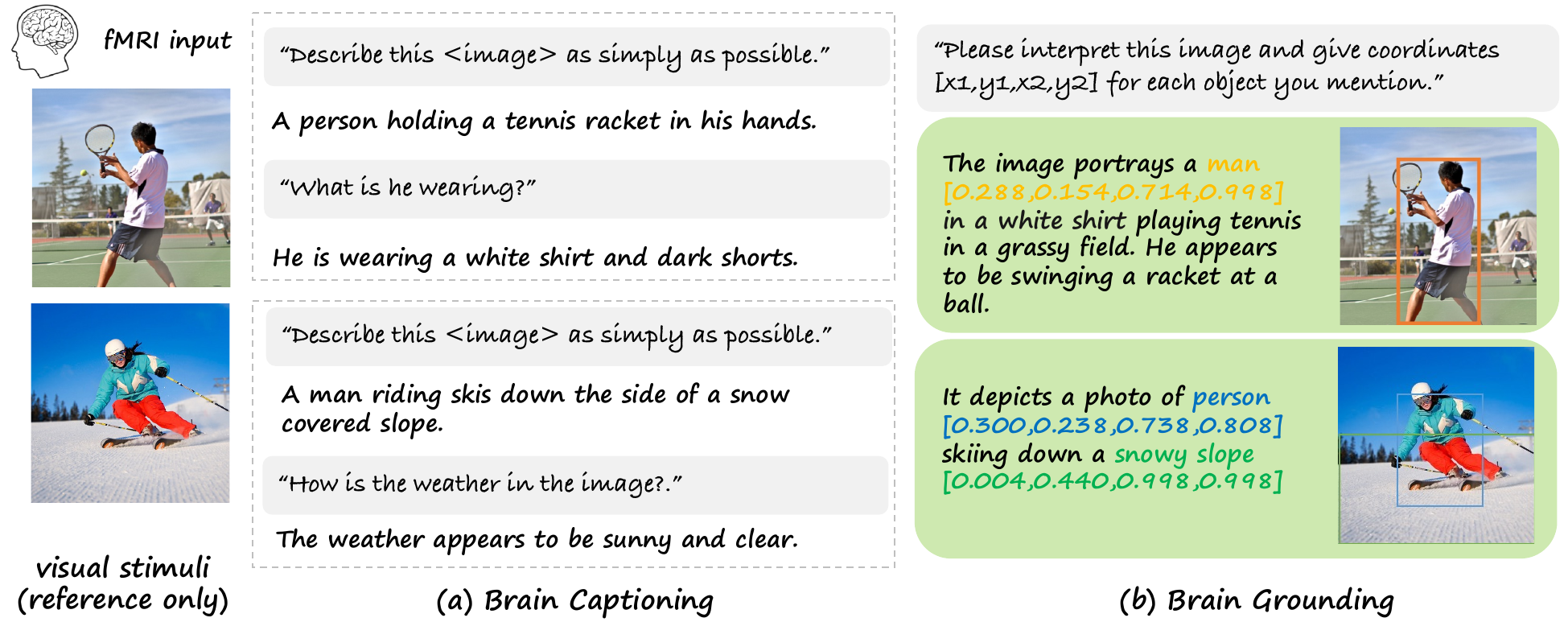}
    \caption{\textbf{Example Results.}
    Our method inherits the multimodal capability from MLLMs and thus supports multiple  brain captioning and grounding tasks.
    Different task prompts for the same input brain signal result in unique outcomes.
    }
    \label{fig:demo_example}
\end{figure}

\subsection{Brain Grounding}
\label{subsec:grounding}

Our method showcases grounding capabilities across various settings by adapting to the corresponding instructions, which is illustrated in~\cref{fig:demo_example}.
For instance, we conduct spotting captioning, a task aimed at generating a description of the image along with bounding boxes for the mentioned items, using the instruction `{\small \texttt{Please interpret this image and provide coordinates [x1,y1,x2,y2] for each object you mention}}'. We can also perform referring expression comprehension using `{\small \texttt{could you find and tell me the coordinates of <expr>?}}'.
For evaluation, we detect queried objects and report the accuracy and IoU. The accuracy metric `acc@m' measures the percentage of correctly labeled instances with an IoU greater than the threshold $m$.
Results of acc@0.5 are reported in~\cref{tab:brain_grounding} and examples are depicted in~\cref{fig:demo_example}. More results are in the appendix.

\begin{table}[t]
\caption{\textbf{Brain Grounding.} `\methodSubject{Sx}' refers to our model trained with a single subject only, while `\method{}' denotes the model with cross-subject training.
`Shikra-w/img' refers to the \textit{visual} grounding result from Shikra~\cite{chen2023shikra} using the ground truth visual stimuli (images) as input.
Similarly, `Shikra-w/method' provides visual grounding results using images produced by visual decoding methods~\cite{ozcelik2023brain,scotti2023reconstructing,xia2023dream}.
We highlight \tgold{best}, \tsilver{second-best}, and \tbronze{third-best} performance per subject. 
}
\vspace{-2.5mm}
\label{tab:brain_grounding}
\centering
\setlength{\tabcolsep}{4pt}
\resizebox{\textwidth}{!}{
\begin{tabular}{@{}l|c|cc|cc|cc|cc|cc|c@{}}
\toprule
\multirow{2}{*}{Method} & \multirow{2}{*}{Eval} & \multicolumn{2}{c|}{All} & \multicolumn{2}{c|}{Salient}  & \multicolumn{2}{c|}{Salient Creatures} & \multicolumn{2}{c|}{Salient Objects}  & \multicolumn{2}{c|}{Inconspicuous} & \multirow{2}{*}{Time (s)}  \\
~ & ~ & acc@0.5  & IoU      & acc@0.5      & IoU      & acc@0.5       & IoU       & acc@0.5       & IoU       & acc@0.5       & IoU    & ~ \\
\midrule
Shikra-w/img~\cite{chen2023shikra} & * & 51.96  & 47.22 & 62.92   & 56.44 & 66.71   & 59.34  & 58.79   & 53.27 & 38.29   & 35.71  & 0.96 \\
\midrule
Shikra-w/BrainDiffuser~\cite{ozcelik2023brain} & \multirow{5}{*}{S1} & \silver{17.49} & \silver{19.34} & \silver{27.18} & \bronze{27.46} & \silver{38.71} & \silver{34.63} & 14.62  & 19.66 & \gold{5.39}  & \silver{9.20}   & 16.4 \\
Shikra-w/MindEye~\cite{scotti2023reconstructing} & ~ & 15.34 & 18.65 & 23.83 & 26.96 & 29.29  & 31.64 & \bronze{17.88} & \silver{21.86} & \bronze{4.74}  & \bronze{8.28} & 16.4 \\
Shikra-w/DREAM~\cite{xia2023dream} & ~ & {16.21} & 18.65 & {26.51} & 27.35 & \bronze{34.43} & \bronze{33.85} & \bronze{17.88} & 20.28 & 3.35 & 7.78 & \bronze{10.5} \\
Shikra-w/\method{} & ~ & \bronze{16.83} & \bronze{18.69} & \bronze{27.10} & \silver{27.55} & {34.14} & {33.65} & \silver{19.44} & \bronze{20.92}  & 4.00 & 7.64 & 16.4\\
\methodSubject{S1} & ~ & {13.72} & {17.56} & {21.52} & {25.14} & {26.00} & {29.06} & {16.64} & {20.88} & {4.00} & {8.08} & \gold{0.92} \\
\method{} & ~ & \gold{18.93} & \gold{21.28}  &  \gold{30.23} & \gold{30.18} & \gold{39.57}    & \gold{36.64} & \gold{20.06} & \gold{23.14}   & \silver{4.83}  & \gold{10.18} & \gold{0.92} \\
\midrule
\methodSubject{S2} & \multirow{2}{*}{S2}    & {15.21} & {18.68} & {23.60} & {26.59} & {27.86} & {30.51} & \gold{18.97} & \gold{22.32} & {4.74} & {8.81} & - \\
\method{} & ~                               & \gold{18.27} & \gold{20.77} & \gold{28.22} & \gold{29.19} & \gold{38.29} & \gold{36.13} & {17.26} & {21.63} & \gold{5.86} & \gold{10.25} & - \\
\midrule
\methodSubject{S5} & \multirow{2}{*}{S5} & {14.72} & {18.45} & {22.93} & {26.34} & {26.86} & {29.84} & {18.66} & {22.52} & {4.46} & {8.60} & - \\
\method{} & ~ & \gold{18.19} & \gold{20.85} & \gold{28.74} & \gold{30.02} & \gold{36.71} & \gold{36.25} & \gold{20.06} & \gold{23.23} & \gold{5.02} & \gold{9.41} & - \\
\midrule
\methodSubject{S7} & \multirow{2}{*}{S7} & {13.60} & {17.83} & {21.07} & {25.19} & {24.57} &  {28.90} & {17.26} & {21.15} & {4.28} & {8.64} & - \\
\method{} & ~ & \gold{16.74} & \gold{19.63} & \gold{25.69} & \gold{27.90} & \gold{33.14} & \gold{33.42} & \gold{17.57} & \gold{21.89} & \gold{5.58} & \gold{9.31} & - \\
\bottomrule
\end{tabular}
}\\
{\tiny * The subjects test sets use the same reference images making `Shikra-w/img' identical for all subjects. }
\end{table}

In~\cref{tab:brain_grounding} we again report an approximate upper bound `Shikra-w/img' being the \textit{visual} grounding using the ground truth image.
Given the absence of prior brain grounding baselines, we construct natural baselines by combining Shikra with the images from SOTA visual decoding methods~\cite{ozcelik2023brain,scotti2023reconstructing,xia2023dream}, referred as `Shikra-w/{method}'. We also report `Shikra-w/\method{}' using our own visual decoding later described in~\cref{subsec:decoding}.
Being the first to attempt decoding spatial information from brain signals, our method `\method{}' performs roughly on par with our constructed baselines while being at least x10 {faster for grounding noting that speed is a critical characteristics for brain controlled applications}. 

In addition to metrics for all classes denoted `All', we inspire from neuroscience exploring the salience-processing systems in the human brain~\cite{uddin2015salience} for more detailed evaluation. Specifically, we group the 80 classes of COCO~\cite{lin2014microsoft} according to their prominence into: `Salient', being the union of `Salient Creatures' (people and animals) and `Salient Objects' (\eg, car, bed, table), and `Inconspicuous' (\eg, backpack, knife, toothbrush). We report the detailed mapping in the appendix. 
It is interesting to note that \method{} is outperformed on Salient Creatures but not on Salient Objects and Inconspicious elements. This suggests that visual decoding effectively reconstruct the salient creatures in the image space, arguably because the subject focuses on the latter.

Experimentally, we also notice that images containing few salient objects exhibit better performance compared to cluttered scenes, and easy background also lead to better grounding.
Conversely, we note that localization suffers when images are filled with numerous inconspicuous objects.
We argue that inconspicuous objects in the image may not draw the subject's attention, or that relevant brain activities may not be effectively captured during experiments~\cite{allen2022massive}.
Our categorization and observation also align with the semantic selectivity found in the higher visual cortex of the human brain~\cite{kanwisher1997fusiform,desimone1984stimulus,puce1996differential}, which contains specialization of certain regions that respond selectively to specific semantic categories of visual stimuli, such as faces, bodies, words, food, and places. 
The results demonstrate that our method performs well in relevant cases.

\subsection{Brain Retrieval}
\label{subsec:retrieval}

The retrieval evaluation demonstrates the amount of image-specific information contained in the brain embedding. 
Following~\cite{scotti2023reconstructing}, we conduct three experiments: forward retrieval, backward retrieval, and exemplar retrieval.
The \textit{forward} retrieval computes accuracy of identifying the correct paired CLIP image embedding from 300 brain embeddings. Conversely, the \textit{backward} retrieval finding the correct brain embedding from 300 image embeddings.
For a fair comparison, we modify the output dimension and proceed to optimize the encoder and embedding using an InfoNCE~\cite{oord2018representation} loss.
We follow the same procedure as in \cite{lin2022mind} for calculating the retrieval metrics reported in \cref{tab:retrieval}.
The \textit{exemplar} retrieval aims to find the exact original image within the 982 test images. Our method outperforms current methods with accuracy percentages of $94.2\%$, $91.3\%$, and $93.8\%$ on forward, backward, and exemplar retrieval experiments, respectively. 
These results demonstrate the ability to distinguish among misconstruable exemplars and suggest the fine-grained, image-specific information retained in the predicted brain embeddings.
\begin{table}[t]
\setlength{\fboxrule}{0pt} 
\setlength{\fboxsep}{2pt}
\setlength{\tabcolsep}{4pt}
    \centering
    \caption{\textbf{Brain Retrieval.}
    We report \textit{forward}, \textit{backward}, and \textit{exemplar} retrieval metrics~\cite{scotti2023reconstructing}, showing that our method significantly outperforms the baselines. We also compare the floating-point operations (FLOPs), multiply-accumulate operations (MACs), and model parameters (Params).    
    `\method{}' denotes the model with cross-subject training. 
    Colors represent the \tgold{best} and \tsilver{second-best} performance.
    }
    \label{tab:retrieval}
    \vspace{-2.5mm}
    \resizebox{0.85\textwidth}{!}
    {
    \begin{tabular}{lcccccc}
    \toprule
        Method & Forward & Backward & Exemplar & FLOPs (G) & MACs (G) & Params (M) \\ 
        \midrule
        MindReader~\cite{lin2022mind} & $11.0\%$ & $49.0\%$ & $\backslash$  & $\backslash$ & $\backslash$ & $\backslash$ \\ 
        BrainDiffuser~\cite{ozcelik2023brain} & $21.1\%$ & $30.3\%$ & $\backslash$  & $\backslash$ & $\backslash$ & $\backslash$ \\ 
        MindEye~\cite{scotti2023reconstructing} & \silver{$93.6\%$} & \silver{$90.1\%$} & \silver{93.2\%} & \gold{52.27} & \gold{26.13} & \silver{1,003.64} \\ 
        \midrule
        \method{} & \gold{$94.2\%$} & \gold{$91.3\%$}  & \gold{93.8\%}
 & \silver{67.48} & \silver{33.72}  & \gold{146.24} \\ 
        \bottomrule
    \end{tabular}
    }
\end{table}

\begin{table}[t]
\setlength{\fboxrule}{0pt} 
\setlength{\fboxsep}{2pt}
\setlength{\tabcolsep}{4pt}
	\centering
	\caption{\textbf{Visual Decoding.}
 Following the standard evaluation metrics~\cite{ozcelik2023brain}, our method performs on par or better than the SOTA methods~\cite{lin2022mind,takagi2023high,ozcelik2023brain,scotti2023reconstructing,xia2023dream}. 
 Colors represent the \tgold{best}, \tsilver{second-best}, and \tbronze{third-best} performance. 
 }
    \vspace{-2.5mm}
	\label{tab:visual_decoding}
	\resizebox{\textwidth}{!}{
		\begin{tabular}{l|cccc|cccc}
			\toprule
			\multirow{2}{*}{Method}  & \multicolumn{4}{c|}{Low-Level} & \multicolumn{4}{c}{High-Level}\\
			~ & PixCorr $\uparrow$ & SSIM $\uparrow$ & AlexNet(2) $\uparrow$ & AlexNet(5) $\uparrow$ & Inception $\uparrow$ & CLIP $\uparrow$  & EffNet-B $\downarrow$ & SwAV $\downarrow$\\
			\midrule
			MindReader~\cite{lin2022mind}  &- &- &- &-  & 78.2\% &- &- &-\\
			SDRecon~\cite{takagi2023high} &- &- &  83.0\% & 83.0\%  & 76.0\% & 77.0\% & - & -\\
			BrainDiffuser~\cite{ozcelik2023brain}  & .254 & \gold{.356} & \bronze{94.2\%} & 96.2\%  & {87.2\%} & {91.5\%} & {.775} & {.423} \\            MindEye~\cite{scotti2023reconstructing} & \gold{.309} & .323 & \silver{94.7\%} & \gold{97.8\%} & \gold{93.8\%} & \gold{94.1\%} & \gold{.645} & \gold{.367} \\
            DREAM~\cite{xia2023dream} & \bronze{.274}    & \bronze{.328} & 93.9\%      & \bronze{96.7\%}      & \silver{93.4\%}     & \gold{94.1\%} & \gold{.645}     & \bronze{.418} \\
            \midrule
            \method{} & \silver{.283}    & \silver{.341} & \gold{95.5\%}      & \silver{97.0\%}      & \bronze{91.7\%}     & \silver{93.5\%} & \silver{.700}     & \silver{.393} \\
             \bottomrule
		\end{tabular}
	}
\end{table}

\subsection{Visual Decoding}
\label{subsec:decoding}

Although this is not our primary purpose, to show the versatile capabilities of our method, we conduct experiments on the visual decoding task and compare with SOTAs~\cite{lin2022mind,takagi2023high,ozcelik2023brain,scotti2023reconstructing} using recognized metrics.
While our method is not specifically tailored for this task, the textual and visual outputs it generates can be used as cues for fMRI-to-image reconstruction. 
Our results in~\cref{tab:visual_decoding} is utilizing the Versatile Diffusion~\cite{xu2022versatile} to reconstruct the image based on the decoded text and CLIP image embedding obtained in~\cref{subsec:retrieval}.
We employ the same evaluation metrics as used in~\cite{ozcelik2023brain}.
Specifically, PixCorr calculates the pixel-level correlation between the ground-truth and reconstructed images.
SSIM~\cite{wang2004image} measures the textural and structural similarity instead of pixel-wise differences.
Two-way comparisons are conducted using the second and fifth layers of AlexNet~\cite{krizhevsky2017imagenet}, the last pooling layer of Inceptionv3~\cite{szegedy2016rethinking}, and the last layer of CLIP ViT-L/14 image encoder~\cite{radford2021learning}. 
EffNet-B and SwAV are distances from EfficientNet~\cite{tan2019efficientnet} and SwAV-ResNet50~\cite{caron2020unsupervised}. The first four metrics focus on low-level characteristics, whereas the remaining metrics are concerned with higher-level measurements.

Results in \cref{tab:visual_decoding} demonstrate that our method performs comparably or better than state-of-the-art methods without any specific designs tailored for this task.
Moreover, with access to common conditions like texts, image embeddings, and bounding boxes, we can leverage a wide range of pretrained image generation models. These models encompass text-to-image (\eg, SD~\cite{rombach2022high}, SD-XL~\cite{podell2023sdxl}), layout-to-image~\cite{li2023gligen}, and multiple-condition~\cite{xu2022versatile}.
Details are in the appendix.

\subsection{Weakly-Supervised Adaptation}
\label{subsec:few_shot_learning}

\begin{figure}[!t]
    \centering
    \resizebox{1.0\linewidth}{!}{%
    \setlength{\tabcolsep}{2pt}%
    \begin{tabular}{ccccc}
	\multicolumn{3}{c}{\scriptsize\textbf{Captioning}} & \multicolumn{2}{c}{\scriptsize\textbf{Grounding}}\\
	\cmidrule(lr){1-3}\cmidrule(lr){4-5}
	 \begin{subfigure}[b]{0.18\textwidth}
		\includegraphics[width=\textwidth]{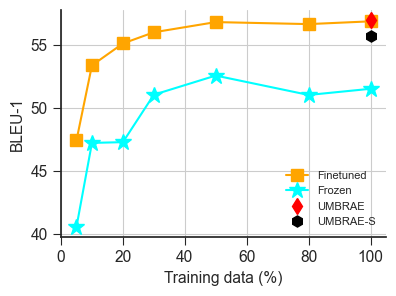}
	\end{subfigure}&%
	\begin{subfigure}[b]{0.18\textwidth}
		\includegraphics[width=\textwidth]{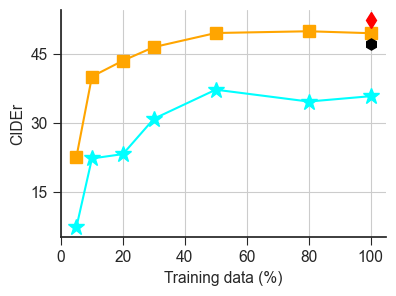}
	\end{subfigure}&
	\begin{subfigure}[b]{0.18\textwidth}
		\includegraphics[width=\textwidth]{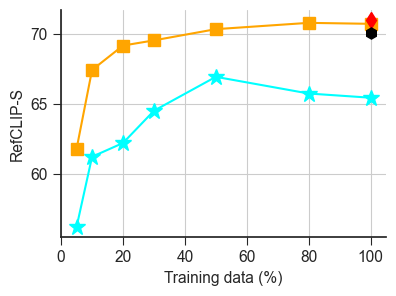}
	\end{subfigure}&%
	\begin{subfigure}[b]{0.18\textwidth}
		\includegraphics[width=\textwidth]{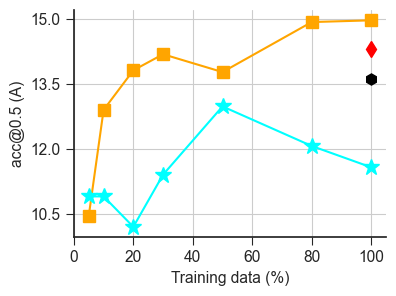}
	\end{subfigure}&%
	\begin{subfigure}[b]{0.18\textwidth}
		\includegraphics[width=\textwidth]{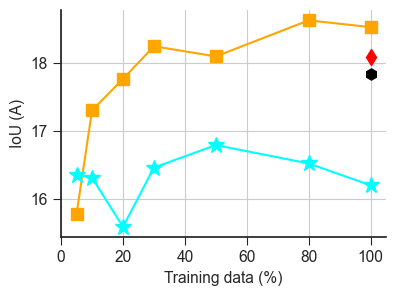}
	\end{subfigure}\\%
 	   \begin{subfigure}[b]{0.18\textwidth}
		       \includegraphics[width=\textwidth]{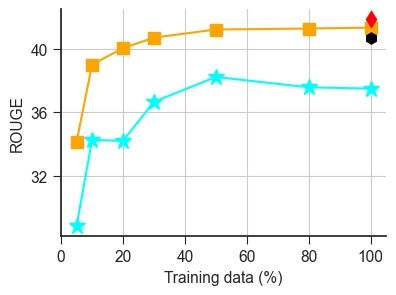}
		   \end{subfigure}&%
	   \begin{subfigure}[b]{0.18\textwidth}
		       \includegraphics[width=\textwidth]{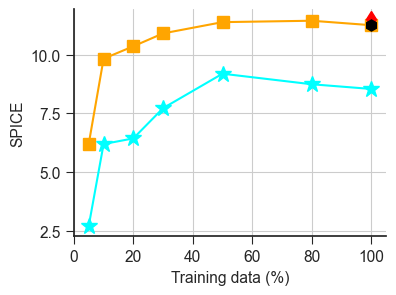}
		   \end{subfigure}&%
	   \begin{subfigure}[b]{0.18\textwidth}
		\includegraphics[width=\textwidth]{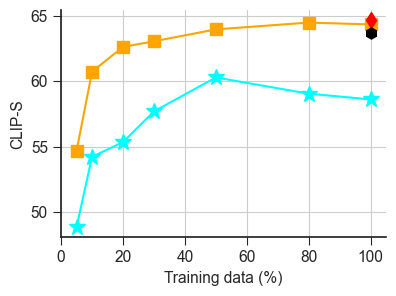}
		   \end{subfigure}&%
	   \begin{subfigure}[b]{0.18\textwidth}
		 \includegraphics[width=\textwidth]{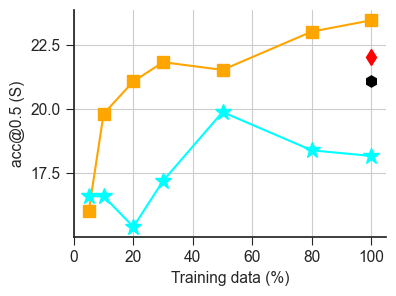}
		   \end{subfigure}&%
	   \begin{subfigure}[b]{0.18\textwidth}
		       \includegraphics[width=\textwidth]{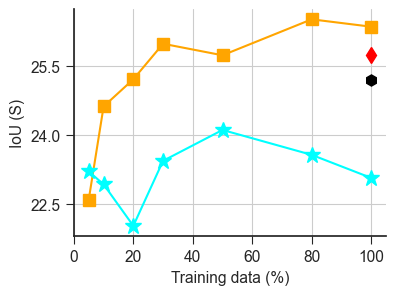}
		   \end{subfigure}
\end{tabular}%
	}
    \caption{\textbf{Weakly-Supervised Subject Adaptation}. This model for S7 is trained or finetuned on a pretrained model (trained on S1, S2, and S5) using varying ratios (0.05, 0.1, 0.2, 0.3, 0.5, 0.8, 1.0) of training data. The model achieves comparable performance using only 30\% of the data and obtains better results when increasing the ratio of used training samples to 50\%, compared to the model trained on the full dataset of S7.
    }
    \label{fig:few_shot}
\end{figure}

Capturing brain signals, such as high-resolution fMRI, requires specialized equipment and professional personnel, making it challenging to collect on a large scale.
A benefit of our cross-subjects training is to allow subject adaptation with minimal training data.
To evaluate this emerging property, we train our brain encoder with subjects S1, S2, and S5 and seek to adapt the trained to a new subject S7 using various amount of training data. 
For ablation, we explore two settings where we train a new tokenizer for S7 with the universal perceive encoder being either \textit{Frozen} or \textit{Finetuned}. 

Plots in~\cref{fig:few_shot} report `Frozen' and `Finetuned' adaptation with variable amount of S7 data. Additionally, we report `\method{}' when trained with all training data of \{S1,~S2,~S5,~S7\} as as well as `\methodSubject{S7}' trained on all S7 data only. 
Notably, compared to `\methodSubject{S7}', our `Finetuned' adaptation achieves comparable performance using only {30\%} of the data and often better when using more than 50\%. 
Training only the tokenizer while keeping the pretrained backbone encoder frozen generally resulted in lower performance compared to fine-tuning the backbone together.
This could be because the backbone encoder did not adequately incorporate the subject discrepancy in S7. 
Please consult the appendix for extra results and discussions on other subjects.

\vspace{-0.1in}
\section{Ablation Study}

\subsection{Architectural Improvements} 

\condparagraph{MLP-based Encoder \textit{vs.} Our Encoder.} 
The MLP-based brain encoder is adapted from~\cite{scotti2023reconstructing} with slight adjustments to match the desired output dimension. This deep MLP backbone amounts to 1,003.64 million parameters per subject.
In comparison, our model needs only 112.63 million parameters for a single subject and 146.24 million for all four subjects. 
This translates to an 88.78\% reduction in parameters for a single subject and a 96.36\% reduction for all four subjects, respectively.
Our single-subject encoder (denoted as \methodSubject{Sx}) surpasses the MLP-based architecture~\cite{scotti2023reconstructing} in captioning (\cref{tab:brain_captioning}), grounding (\cref{tab:brain_grounding}), and retrieval (\cref{tab:retrieval}) tasks by significant margins. The universal encoder (denoted as \method{}) 
achieves even greater performance improvements.

\condparagraph{Single \textit{vs.} Cross-Subject Design.} 
The universal encoder differs from the single-subject encoder solely in the addition of subject-specific tokenizers (\cref{subsec:model_arch}), and its training only varies in the batch sampling strategy (\cref{subsec:training_strategy}), enabling the training of diverse subjects within one model. Additionally, the resources required are basically the same as those for training a single-subject model, eliminating the necessity of extra training time or computational resources. 
This cross-subject design benefits from user diversity, achieving superior performance compared to focusing on a single subject. Results in ~\cref{tab:brain_captioning,tab:brain_grounding,tab:retrieval} show that the cross-subject design surpasses its single-subject counterpart across almost metrics.

\setlength{\fboxrule}{0pt} 
\setlength{\fboxsep}{2pt}
\setlength{\tabcolsep}{4pt}
\begin{table}[t]
\caption{\textbf{Ablation Study.} `MLP' refers to the MLP-based brain encoder~\cite{scotti2023reconstructing}, while `Enc-S' and `Enc-U' represent our transformer-based encoders for single and multiple subjects, respectively. `Dim.' means the output dimension of the brain encoder. 
The output needs to be passed to the adapter for further processing if it is 1024; otherwise, it is directly inputted into the LLM. The adapter has three training settings: `Pretrained' means freezing the weights, `Finetuned' means finetuning based on the pretrained weights, and `Joint' means training with the encoder from scratch. 
`Loss Type' refers to loss functions (MSE, NCE, or both) applied to the outputs from the encoder (E.) or the adapter (A.).
}
\vspace{-2.5mm}
\label{tab:ablation}
\centering
\resizebox{\textwidth}{!}{
\begin{tabular}{@{}lccc|ccccc|cccc@{}}
\toprule
\multicolumn{4}{c|}{Different Ablation Configurations}  & \multicolumn{5}{c|}{Captioning} &
\multicolumn{4}{c}{Grounding} \\ 
&&&&&&&&&\multicolumn{2}{c}{All}&\multicolumn{2}{c}{Salient}\\
Arch. & Dim. & Adapter & Loss Type &  BLEU1  & CIDEr & SPICE & CLIP-S & RefCLIP-S  &  acc@0.5       & IoU      & acc@0.5       & IoU  \\
\midrule
MLP & 1024 & Pretrained & MSE (E.) & \bronze{55.04} & \bronze{46.24} & \bronze{10.80} & \bronze{64.75} & \bronze{70.59} & 13.44 & 17.54 & 20.55 & 24.68 \\
MLP & 1024 & Finetuned & MSE (A.) & 54.02 & 43.24 & 10.35 & 64.09 & 70.02 & 13.56 & \bronze{17.91} & 20.92 & \silver{25.54}\\
Enc-S & 1024 & Pretrained & MSE (E.) & \silver{57.63} & \silver{51.93} & \silver{11.83} & \silver{66.44} & \silver{72.12} & \silver{13.72} & 17.56 & \silver{21.52} & 25.14 \\
Enc-S & 4096 & N/A & MSE (A.)  & 52.06 & 36.40 & 9.06 & 62.30 & 68.27 & 13.31 & 17.04 & 20.85 & 24.78\\ 
Enc-S & 1024 & Joint & MSE (A.) & 55.02 & 43.53 & 10.48 & 64.00 & 70.01 & \silver{13.72} & 17.57 & \bronze{21.44} & 25.15 \\
Enc-S & 1024 & Joint & MSE (E.) NCE (A.)  & 27.09 & 3.16 & 1.27 & 52.69 & 59.08 & 8.72 & 11.40 & 13.78 & 16.26 \\
Enc-S & 1024 & Joint & MSE (A.) NCE (A.)  & 51.69 & 34.09 & 8.71 & 62.27 & 68.05 & \bronze{13.68} & \silver{18.07} & 21.07 & \bronze{25.45} \\
Enc-U & 1024 & Pretrained & MSE (E.) & \gold{59.44} & \gold{61.06} & \gold{12.79} & \gold{67.78} & \gold{73.54} & \gold{18.93} & \gold{21.28} & \gold{30.23} & \gold{30.18} \\
\bottomrule
\end{tabular}
}
\end{table}
\setlength{\tabcolsep}{1.4pt}

\vspace{-0.1in}
\subsection{Training Strategies} 

Current vision-language models typically comprise three main components: an image encoder, an adapter, and a large language model. 
Within this framework, there are several potential ways for multimodal-brain alignment.
For example, one could train the model to map brain responses to a pretrained semantic space through contrastive learning, which has been a common practice in previous methods. Alternatively, one could opt to fine-tune the adapter or train it jointly with the brain encoder, applying separate losses to each component. In this section, we delve into the motivation and rationale behind aligning with image features from the image encoder using a reconstruction loss. Further, in the appendix we ablate our sampling strategy.

\condparagraph{Pretrained \textit{vs.} Finetuned Adapter.} The adapter serves as the bridge connecting multimodal encoders~\cite{han2024onellm,radford2021learning} with the output space of MLLMs~\cite{touvron2023llama,chen2023shikra,liu2023visual}. As shown in~\cref{tab:ablation}, either finetuning the adapter or training it jointly with the brain encoder results in decreased performance, likely due to catastrophic forgetting that occurs when updating well-trained parameters that have been learned from a significantly larger volume of data.

\condparagraph{Image Feature \textit{vs.}  LLM Embedding.} 
In addition to aligning with image features, we also conduct experiments on learning to align brain responses with the pretrained LLM used.
Specifically, we explore three variants: (a) training the brain encoder and finetuning the adapter with different losses, (b) training the brain encoder and the adapter jointly, and (c) adjusting the output dimension of the brain encoder to align directly with the language embedding. As illustrated in \cref{tab:ablation}, all attempts yield less desirable results compared to simply aligning with image features.
Finetuning the entire model with new and sufficient data may indeed achieve better results, but acquiring such data can be challenging. However, experiments suggest that aligning with image features yields the best results when only image-brain pairs are available for training.

\condparagraph{Reconstruction \textit{vs.}  Contrastive Learning.} We further explore the effects of different loss functions on the aforementioned three variants, specifically a pixel-wise reconstruction loss (MSE) and a contrastive loss with MixCo augmentation (NCE)~\cite{oord2018representation}. We also test applying separate losses to outcomes from the encoder and the adapter when trained jointly. The findings show that applying an MSE loss to the image feature while keeping the adapter unchanged leads to the most favorable performance in both concept recognition and object localization. Conversely, employing contrastive learning significantly diminishes performance.

\vspace{-0.1in}
\section{Conclusion}
\vspace{-0.08in}

In this work, we propose a method that decodes multimodal explanations from brain signals. Specifically, we introduce a universal brain encoder for multimodal brain alignment, which enables the recovery of conceptual and spatial details using multimodal large language models. To overcome unique brain patterns among different individuals, we introduce a novel cross-subject training strategy. This enables brain signals from multiple subjects to be trained within the same model and allows weakly-supervised subject adaptation, facilitating the training of a model for a new subject in a data-efficient manner. For evaluation, we construct BrainHub, a brain understanding benchmark, to facilitate future research.

\vspace{.3em}\noindent{}{\scriptsize%
\textbf{Acknowledgements.} This work was supported by the Engineering and Physical Sciences Research Council [grant EP/W523835/1] and the UKRI Future Leaders Fellowship [grant G104084].\par%
}

\section*{Appendix}
\appendix

\setcounter{section}{0}
\setcounter{table}{0}
\setcounter{figure}{0}
\setcounter{equation}{0}

\begin{appendix}
\label{appendix}

In this appendix, we provide further insights, experiments and analyses. First, in \cref{subsec:supmat_prompts} we describe the tasks and associated prompts used in \method{} and in \cref{subsec:supmat_bench} describe the \bench{} benchmark construction. We then report in \cref{sec:supmat_results} additional experiments to showcase the superiority of \method{} on all tasks, and extend our ablation and analyses in \cref{sec:supmat_analysis}. Finally, in \cref{sec:supmat_discussion}, we discuss the method limitations and potential negative impacts.

\section{\method{}: Tasks and Example Prompts}
\label{subsec:supmat_prompts}

Our method inherits multimodal understanding capabilities of MLLMs, enabling the switch between different tasks through the use of various task prompts. Taking Shikra~\cite{chen2023shikra} as an example, we excerpts the task prompts used during their training process in~\cref{tab:supmat_prompt}. 
Three prompts for each task are randomly selected to provide readers an intuitive understanding.  During inference, users are not constrained to these predefined formats and are free to frame their queries in natural language, allowing for a broad spectrum of diverse and compelling task formats.

It should be noted that our method is compatible with nearly all tasks featured in Shikra~\cite{chen2023shikra} and LLaVA~\cite{liu2023visual}, with the exception of tasks that necessitate initially locating an object or scene as inputs, such as grounding captioning and referring expression generation.
Importantly, note that although the term `image' is used in the prompts we do \textit{not} utilize images as input in our brain understanding tasks. \textit{The reference images are only used for visualization purposes.} In practice, the prompt embedding will be concatenated with the brain features predicted from our brain encoder, utilizing brain signals as inputs. We report detailed results for each task of~\cref{tab:supmat_prompt} later in \cref{sec:supmat_results}.

\vspace{-10pt}
\begin{table*}
	\centering
	\caption{
		\textbf{Supported Task Prompts.} 
		The tags \textit{<image>}, \textit{<question>}, and \textit{<expr>} are placeholders, representing input images, questions in QA tasks, and expressions in the REC task. During inference, users are free to create diverse task formats according to actual needs. Q, A, C, 
		and ${C}^\text{Box}$ denote the \textbf{Q}uestion, \textbf{A}nswer, \textbf{C}hain of thoughts~(CoT), 
		and \textbf{C}oT with \textbf{Box}.
		CoT is delivering an answer along with the reasoning process.
		`Box' denotes coordinates of bounding boxes.
	}
	\label{tab:supmat_prompt}
	\resizebox{\textwidth}{!}{%
		\begin{tabular}{l|l}
			\toprule
			\textbf{Task} & 
			\textbf{Three randomly chosen examples from hundreds.}  \\
			\cmidrule(lr){1-1}\cmidrule(lr){2-2}
			\multirow{3}{*}{Captioning}
			& Describe this image <image> as simply as possible.  \\
			& What is the content of the image <image>? Please answer in short sentences.    \\
			& Summarize the content of the photo <image>.  \\
			\cmidrule(lr){1-1}\cmidrule(lr){2-2}
			\multirow{3}{*}{REC}
			& In the given <image>, could you find and tell me the coordinates of <expr>?  \\
			& I need the coordinates of <expr> in <image>, can you please assist me with that?  \\
			& Locate <expr> in <image> and provide its coordinates, please.\\
			\cmidrule(lr){1-1}\cmidrule(lr){2-2}
			\multirow{3}{*}{Spotting Cap.}
			& Can you provide a description of the image <image> and include the coordinates [x0,y0,x1,y1] for each mentioned object?  \\
			& Please explain what's happening in the photo <image> and give coordinates [xmin,ymin,xmax,ymax] for the items you reference.    \\
			& How would you describe the contents of the image <image>? Please provide the positions of mentioned objects in square brackets.  \\
			\cmidrule(lr){1-1}\cmidrule(lr){2-2}
			\multirow{3}{*}{$\text{Q} \rightarrow \text{A}$}
			& I want to know the answer to `<question>' Refer to the image <image> and give a clear response.  \\
			& Answer this question directly after referring to the image <image>: <question>    \\
			& Examine the image <image> and provide a brief answer for `<question>'  \\
			\cmidrule(lr){1-1}\cmidrule(lr){2-2}
			\multirow{3}{*}{$\text{Q}\rightarrow\text{CA}$}
			& Having a look at image <image>, can you tell me the answer to my question '<question>' and the logic leading to it?  \\
			& Please answer the following question '<question>' based on the image <image>, and describe your thought process \\
			& Upon analyzing the image <image>, please find the answer to my question '<question>' and provide a detailed explanation. \\
			\cmidrule(lr){1-1}\cmidrule(lr){2-2}
			\multirow{3}{*}{Q$\rightarrow\text{C}^\text{Box}$A}
			& <question> Please offer your reasoning process, and provide bounding boxes of mentioned objects within square brackets. Here is the picture <image>  \\
			& Please explain your reasoning and provide bounding boxes, denoted by square brackets, for the objects mentioned in the picture <image>. <question>  \\
			& Consider <image>, and then provide a well-reasoned answer to '<question>' Don't forget to mark relevant object locations using [x0,y0,x1,y1].\\
			\bottomrule
		\end{tabular}%
	}
\end{table*}

\section{Details on \bench}
\label{subsec:supmat_bench}
Our multimodal brain understanding benchmark, coined \bench, extends the popular NSD~\cite{allen2022massive} using COCO~\cite{lin2014microsoft} annotations.
Here, we first describe NSD~(\cref{subsec:supmat_bench_nsd}) and then elaborate the construction of \bench~(\cref{subsec:supmat_bench_cons}).

\subsection{Natural Scenes Dataset} %
\label{subsec:supmat_bench_nsd}

Natural Scenes Dataset~\cite{allen2022massive}~(NSD) stands as the largest publicly accessible fMRI dataset. It encompasses brain activity recordings from 8 subjects (participants) who viewed images passively for up to 40 hours inside an MRI machine. Each image was displayed for three seconds and repeated three times across 30-40 scanning sessions, yielding 22,000--30,000 fMRI response trials per subject. 

Following recent studies, we utilize the four subjects who finished all scanning sessions, that is: S1, S2, S5, and S7. 
As in~\cite{scotti2023reconstructing,takagi2023improving,xia2023dream}, we utilize preprocessed voxels corresponding to the `nsdgeneral' brain region. 
The latter region, described by the NSD authors, comprises the subset of voxels in the posterior cortex most responsive to the presented visual stimuli.
In the training set for each subject, there are 8,859 images and 24,980 fMRI trials (with each image tested up to three times). We compute the average response as per previous studies~\cite{scotti2023reconstructing}.
The test set comprises an additional 982 images and 2,770 fMRI trials common across four individuals.
Importantly, all images used during the fMRI recordings are from the COCO~\cite{lin2014microsoft} dataset, which we conveniently use to retrieve the original COCO labels to construct our \bench{} benchmark.

\vspace{-10pt}
\begin{table}
	\centering
	\caption{\textbf{\bench~Details.} The test set characteristics include the number of images, voxels, captions, bounding boxes, regions of interest (ROIs) in the fMRI data, subject references, and their corresponding dimensions. 
	}
	\label{tab:supmat_nsd_dataset}
     \vspace{-8pt}
	\resizebox{0.75\linewidth}{!}
	{
		\begin{tabular}{ccccccc}
			\toprule
			Images & Classes & Captions & Bounding boxes & ROIs  & Subject & Dimension \\ 
			\midrule
			\multirow{4}{*}{982} & \multirow{4}{*}{80} & \multirow{4}{*}{4,913} & \multirow{4}{*}{5,829} & \multirow{4}{*}{\begin{tabular}[c]{@{}c@{}}V1, V2, V3, hV4,\\ VO, PHC, MT,\\ MST, LO, IPS\end{tabular}} & S1 & 15,724   \\ 
			& &          &              &      & S2      & 14,278    \\ 
			& &          &              &      & S5      & 13,039    \\ 
			& &          &              &      & S7      & 12,682     \\ 
			\bottomrule
		\end{tabular}
	}
\end{table}
\vspace{-10pt}

\subsection{Benchmark Construction}
\label{subsec:supmat_bench_cons}

\begin{figure}
	\centering
         \begin{subfigure}[m]{1.0\linewidth}
		\begin{subfigure}[m]{0.21\textwidth}
			\includegraphics[width=\linewidth]{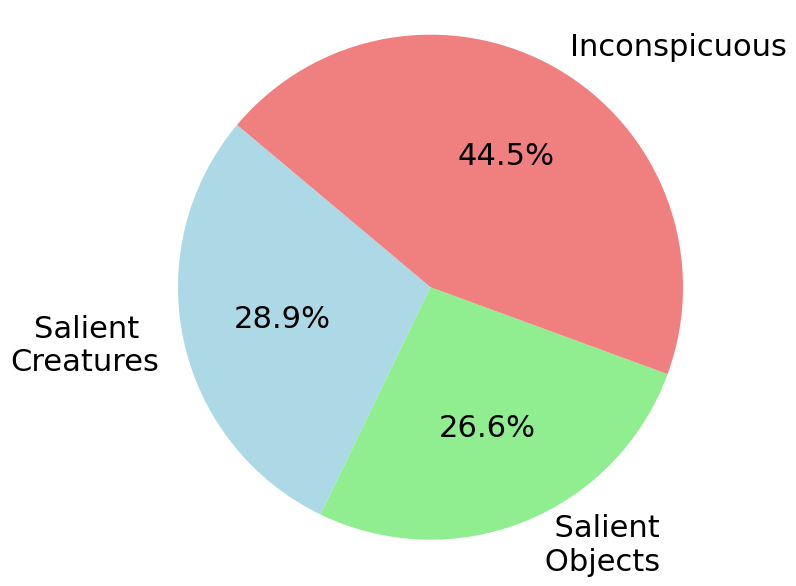}
            \end{subfigure}%
		\begin{subfigure}[m]{0.265\textwidth}
			\includegraphics[width=\linewidth]{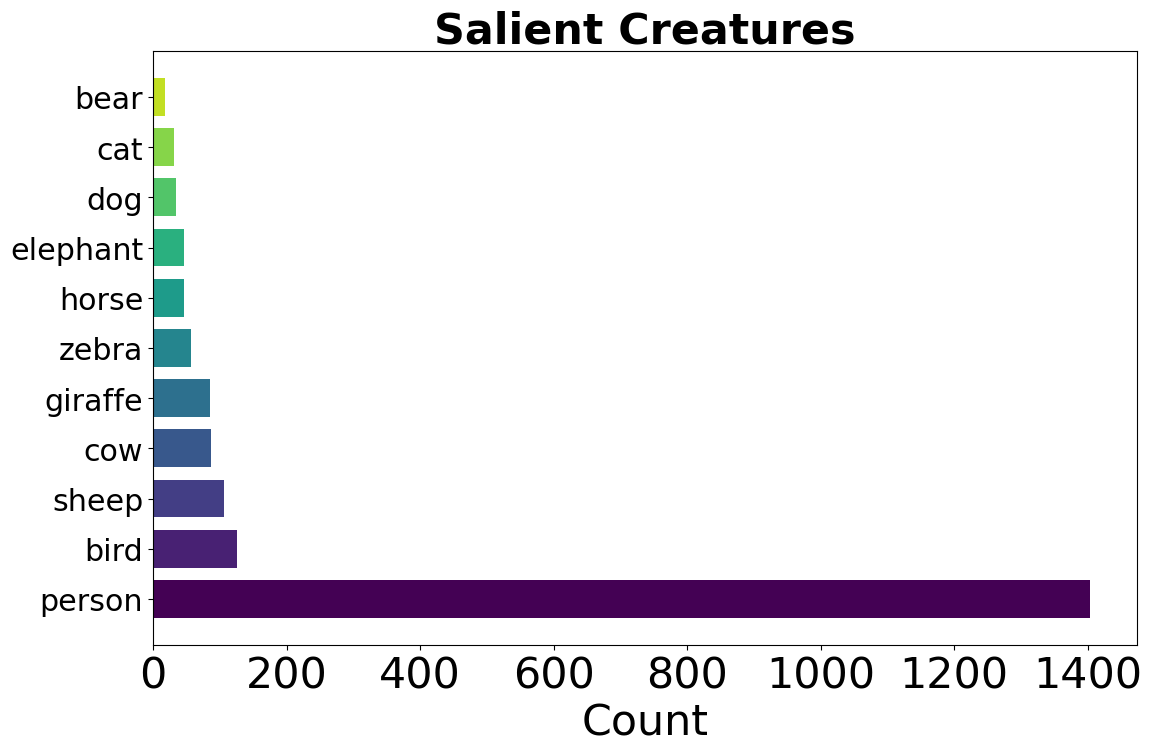}%
		\end{subfigure}%
		\begin{subfigure}[m]{0.265\textwidth}
			\includegraphics[width=\linewidth]{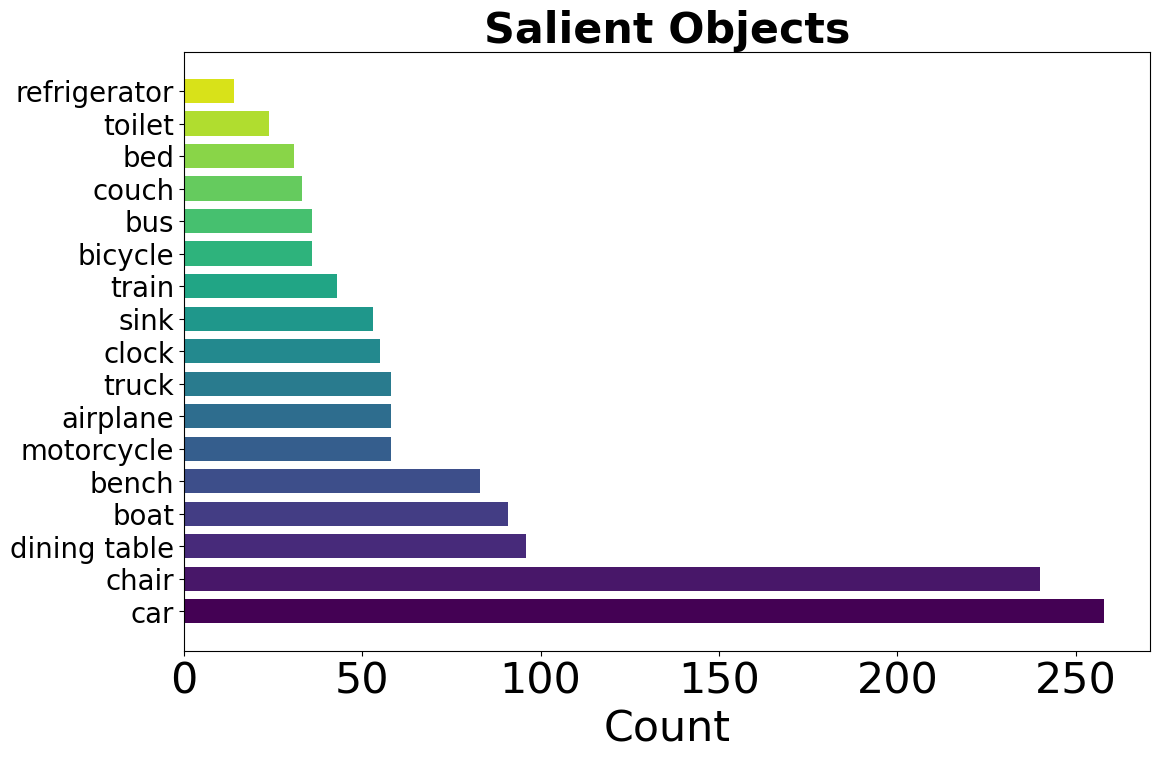}%
		\end{subfigure}%
		\begin{subfigure}[m]{0.265\textwidth}
			\includegraphics[width=\linewidth]{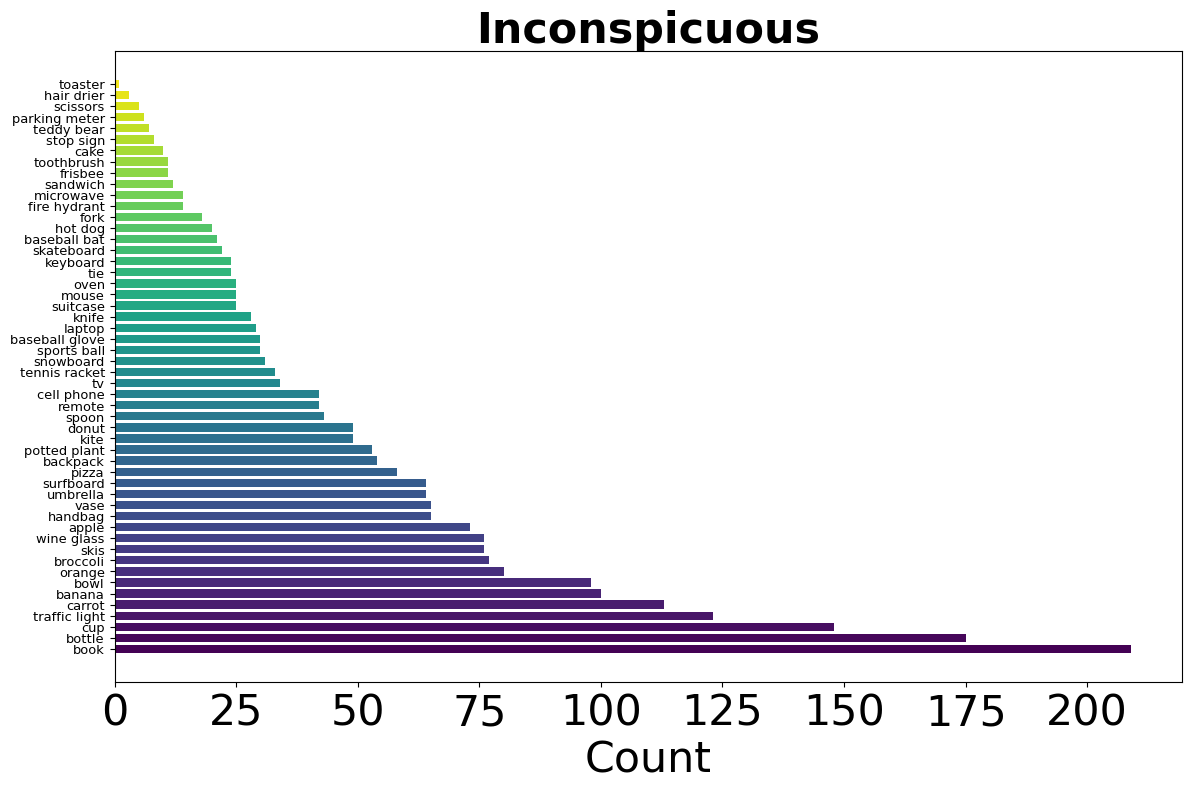}
		\end{subfigure}
          \caption{Statistics.}
	\label{fig:supmat_statistics_stats}
        \end{subfigure}\\[1em]
        \begin{subfigure}[m]{0.95\linewidth}
	\resizebox{1.0\linewidth}{!}
	{
		\begin{tabular}{ccp{12cm}}
			\toprule
			Category   & \quad & COCO classes (\# of classes)\\ 
			\midrule
			\textbf{A}  & &  \textbf{S} + \textbf{I} (80)\\
			\textbf{S}  & &  \textbf{SC} + \textbf{SO} (28)\\         
			\textbf{SC} & &  person, bird, cat, dog, horse, sheep, cow, elephant, bear, zebra, giraffe (11)\\
			\textbf{SO} & &  bicycle, car, motorcycle, airplane, bus, train, truck, boat, bench, chair, couch, bed, dining table, toilet, sink, refrigerator, clock (17)\\
			\textbf{I} & &  traffic light, fire hydrant, stop sign, parking meter, backpack, umbrella, handbag, tie, suitcase, frisbee, skis, snowboard, sports ball, kite, baseball bat, baseball glove, skateboard, surfboard, tennis racket, bottle, wine glass, cup, fork, knife, spoon, bowl, banana, apple, sandwich, orange, broccoli, carrot, hot dog, pizza, donut, cake, potted plant, tv, laptop, mouse, remote, keyboard, cell phone, microwave, oven, toaster, book, vase, scissors, teddy bear, hair drier, toothbrush   (52)\\
			\bottomrule
		\end{tabular}
	}
        \caption{Mapping between categories and COCO classes.}
	\label{fig:supmat_statistics_map}
        \end{subfigure}
	\caption{\textbf{\bench{} Statistics.} We illustrate the statistics in (\subref{fig:supmat_statistics_stats}) and the mapping relationships in (\subref{fig:supmat_statistics_map}) for the categories ontology used in \bench{}, \wrt to the original COCO classes. Please zoom in (\subref{fig:supmat_statistics_stats}) for details.}
	\label{fig:supmat_statistics}
\end{figure}

\bench{} extends NSD~\cite{allen2022massive} for the four subjects who finished all scanning sessions (S1, S2, S5, and S7). 
\cref{tab:supmat_nsd_dataset} outlines the characteristics of the benchmark. Each image viewed by the subject contains several captions and may include multiple bounding boxes (instances) of each class. 
According to their salience in images~\cite{uddin2015salience}, we group the 80 classes of COCO into 4 salience categories: Salient, Salient Creatures, Salient Objects, and Inconspicuous. Note that ‘Salient’ is the combination of Salient Creatures and Objects.
\cref{fig:supmat_statistics} shows the statistics and mapping of our categories, w.r.t. COCO classes. 
The inconspicuous (\textbf{I}) category accounts for the largest proportion, while the salient objects (\textbf{SO}) and creatures (\textbf{SC}) are roughly even. The classes within each category exhibit significant imbalance; for instance, the number of instances for the `person' class stands out in the salient category.

\vspace{-2.5mm}
\section{Additional Experiments}
\label{sec:supmat_results}

\subsection{Brain Captioning}
\label{subsec:supmat_captioning}

\begin{figure}
    \centering
    \begin{minipage}{0.172\textwidth}
        \scriptsize
        \centering
        \textbf{Image}\\
        \textit{\tiny(for reference only)}
    \end{minipage}
    \begin{minipage}{0.815\textwidth}
        \scriptsize
        \centering
        \textbf{Brain captioning}
    \end{minipage}\\
    \begin{minipage}{0.172\textwidth}
        \centering
        \includegraphics[width=\textwidth]{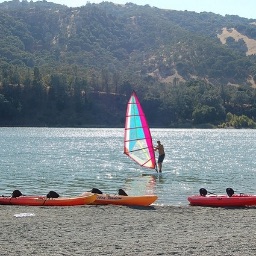}
    \end{minipage}
    \begin{minipage}{0.815\textwidth}
    \begin{tcolorbox}[colback=white!100,left=1pt,
                  right=1pt] {
        \centering
        \resizebox{\textwidth}{!}{
        \begin{tabular}{ll}
        SDRecon~\cite{takagi2023improving} & \textcolor[rgb]{0.8,0,0.8}{the sea with some trees in the fore, and mountains in the distance are red} \\ 
        BrainCap~\cite{ferrante2023brain} & \textcolor[rgb]{0.04453125, 0.54921875, 0.14453125}{a person is standing on a beach with a snowboard.} \\  
        OneLLM~\cite{han2024onellm} &  \textcolor[rgb]{0.9,0.8,0.5}{A group of people gathered on the beach flying kites.} \\ 
        \methodSubject{S1} & \textcolor[rgb]{0.97265625, 0.55, 0.27265625}{A group of people riding boards on top of a beach.} \\ 
        \method{} & \textcolor[rgb]{0.8,0,0}{A person is parasailing on a lake with mountains in the distance.} \\ 
        \end{tabular}}}
        \end{tcolorbox}
    \end{minipage} \\
    \begin{minipage}{0.172\textwidth}
        \centering
        \includegraphics[width=\textwidth]{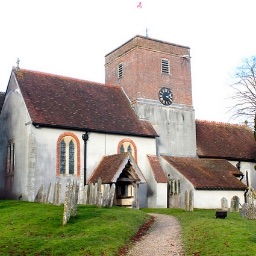}
    \end{minipage}
    \begin{minipage}{0.815\textwidth}
    \begin{tcolorbox}[colback=white!100,left=1pt,
                  right=1pt] {
        \centering
        \resizebox{\textwidth}{!}{
        \begin{tabular}{ll}
        SDRecon~\cite{takagi2023improving} & \textcolor[rgb]{0.8,0,0.8}{the city of london from an perspective} \qquad \qquad \qquad \qquad \qquad \qquad \qquad  \qquad \qquad  \\ 
        BrainCap~\cite{ferrante2023brain} & \textcolor[rgb]{0.04453125, 0.54921875, 0.14453125}{a corner of a building with a train station.} \\  
        OneLLM~\cite{han2024onellm} &  \textcolor[rgb]{0.9,0.8,0.5}{A kitchen is seen through an open door.} \\ 
        \methodSubject{S1} & \textcolor[rgb]{0.97265625, 0.55, 0.27265625}{A large building with a clock tower on top.} \\ 
        \method{} & \textcolor[rgb]{0.8,0,0}{A large building with a clock tower on top.} \\ 
        \end{tabular}}}
        \end{tcolorbox}
\end{minipage} \\
\begin{minipage}{0.172\textwidth}
        \centering
        \includegraphics[width=\textwidth]{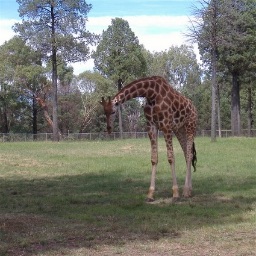}
    \end{minipage}
    \begin{minipage}{0.815\textwidth}
    \begin{tcolorbox}[colback=white!100,left=1pt,
                  right=1pt] {
        \centering
        \resizebox{\textwidth}{!}{
        \begin{tabular}{ll}
        SDRecon~\cite{takagi2023improving} & \textcolor[rgb]{0.8,0,0.8}{some animals in the wild area near to wildlife world} \qquad \qquad \qquad \qquad \qquad \qquad \\ 
        BrainCap~\cite{ferrante2023brain} & \textcolor[rgb]{0.04453125, 0.54921875, 0.14453125}{a large area of grass.} \\  
        OneLLM~\cite{han2024onellm} &  \textcolor[rgb]{0.9,0.8,0.5}{A man standing on a snowy slope skiing.} \\ 
        \methodSubject{S1} & \textcolor[rgb]{0.97265625, 0.55, 0.27265625}{A giraffe is standing in a grassy field.} \\ 
        \method{} & \textcolor[rgb]{0.8,0,0}{A giraffe is standing in a grassy field.} \\ 
        \end{tabular}}}
        \end{tcolorbox}
\end{minipage} \\
\begin{minipage}{0.172\textwidth}
        \centering
        \includegraphics[width=\textwidth]{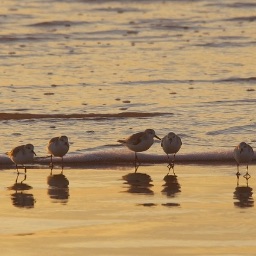}
    \end{minipage}
    \begin{minipage}{0.815\textwidth}
    \begin{tcolorbox}[colback=white!100,left=1pt,
                  right=1pt] {
        \centering
        \resizebox{\textwidth}{!}{
        \begin{tabular}{ll}
        SDRecon~\cite{takagi2023improving} & \textcolor[rgb]{0.8,0,0.8}{a beach with rocks on the shore and ocean} \qquad \qquad \qquad \qquad \qquad \qquad \qquad  \qquad \\ 
        BrainCap~\cite{ferrante2023brain} & \textcolor[rgb]{0.04453125, 0.54921875, 0.14453125}{a large body of water with some birds on it.} \\  
        OneLLM~\cite{han2024onellm} &  \textcolor[rgb]{0.9,0.8,0.5}{A foggy day in San Francisco with a bus and a streetlamp.} \\ 
        \methodSubject{S1} & \textcolor[rgb]{0.97265625, 0.55, 0.27265625}{A group of birds standing on top of a sandy beach.} \\ 
        \method{} & \textcolor[rgb]{0.8,0,0}{A flock of birds standing on a body of water.} \\ 
        \end{tabular}}}
        \end{tcolorbox}
\end{minipage} \\
\begin{minipage}{0.172\textwidth}
        \centering
        \includegraphics[width=\textwidth]{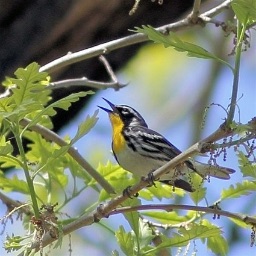}
    \end{minipage}
    \begin{minipage}{0.815\textwidth}
    \begin{tcolorbox}[colback=white!100,left=1pt,
                  right=1pt] {
        \centering
        \resizebox{\textwidth}{!}{
        \begin{tabular}{ll}
        SDRecon~\cite{takagi2023improving} & \textcolor[rgb]{0.8,0,0.8}{some of the amazingly colorful african animals, the only because its natural habitat is}  \\ 
        BrainCap~\cite{ferrante2023brain} & \textcolor[rgb]{0.04453125, 0.54921875, 0.14453125}{a close up of a person with a ball} \\  
        OneLLM~\cite{han2024onellm} &  \textcolor[rgb]{0.9,0.8,0.5}{A black and white sheep standing next to a green bush.} \\ 
        \methodSubject{S1} & \textcolor[rgb]{0.97265625, 0.55, 0.27265625}{A small bird is sitting on a tree branch.} \\ 
        \method{} & \textcolor[rgb]{0.8,0,0}{A bird is sitting on the branches of a tree.} \\ 
        \end{tabular}}}
        \end{tcolorbox}
\end{minipage} \\
\begin{minipage}{0.172\textwidth}
        \centering
        \includegraphics[width=\textwidth]{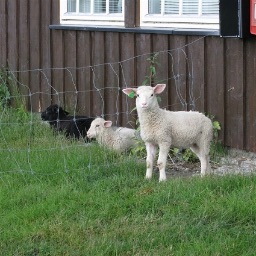}
    \end{minipage}
    \begin{minipage}{0.815\textwidth}
    \begin{tcolorbox}[colback=white!100,left=1pt,
                  right=1pt] {
        \centering
        \resizebox{\textwidth}{!}{
        \begin{tabular}{ll}
        SDRecon~\cite{takagi2023improving} & \textcolor[rgb]{0.8,0,0.8}{an abandoned house and two of her babies} \qquad \qquad \qquad \qquad \qquad \qquad \qquad \qquad \\ 
        BrainCap~\cite{ferrante2023brain} & \textcolor[rgb]{0.04453125, 0.54921875, 0.14453125}{a couple of animals that are in the grass.} \\  
        OneLLM~\cite{han2024onellm} &  \textcolor[rgb]{0.9,0.8,0.5}{A group of three sheep standing next to each other.} \\ 
        \methodSubject{S1} & \textcolor[rgb]{0.97265625, 0.55, 0.27265625}{Three sheep standing next to each other on a grassy field.} \\ 
        \method{} & \textcolor[rgb]{0.8,0,0}{Three sheep standing near each other in a field.} \\ 
        \end{tabular}}}
        \end{tcolorbox}
\end{minipage} \\
\begin{minipage}{0.172\textwidth}
        \centering
        \includegraphics[width=\textwidth]{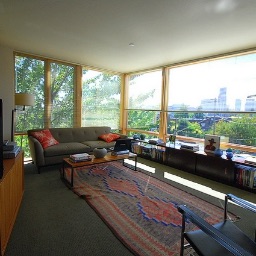}
    \end{minipage}
    \begin{minipage}{0.815\textwidth}
    \begin{tcolorbox}[colback=white!100,left=1pt,
                  right=1pt] {
        \centering
        \resizebox{\textwidth}{!}{
        \begin{tabular}{ll}
        SDRecon~\cite{takagi2023improving} & \textcolor[rgb]{0.8,0,0.8}{an empty room with couch in it} \qquad \qquad \qquad \qquad \qquad \qquad \qquad \qquad \qquad \qquad \qquad \\ 
        BrainCap~\cite{ferrante2023brain} & \textcolor[rgb]{0.04453125, 0.54921875, 0.14453125}{a room with a large window and a sink.} \\  
        OneLLM~\cite{han2024onellm} &  \textcolor[rgb]{0.9,0.8,0.5}{A large silver bed sitting in a room.} \\ 
        \methodSubject{S1} & \textcolor[rgb]{0.97265625, 0.55, 0.27265625}{A living room filled with furniture and a large window.} \\ 
        \method{} & \textcolor[rgb]{0.8,0,0}{A living room filled with furniture and a large window.} \\ 
        \end{tabular}}}
        \end{tcolorbox}
\end{minipage} \\
\begin{minipage}{0.172\textwidth}
        \centering
        \includegraphics[width=\textwidth]{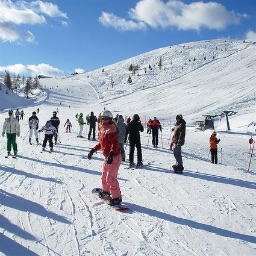}
    \end{minipage}
    \begin{minipage}{0.815\textwidth}
    \begin{tcolorbox}[colback=white!100,left=1pt,
                  right=1pt] {
        \centering
        \resizebox{\textwidth}{!}{
        \begin{tabular}{ll}
        SDRecon~\cite{takagi2023improving} & \textcolor[rgb]{0.8,0,0.8}{a man on a motorcycle riding across the ocean while another man standing on a ramp} \\ 
        BrainCap~\cite{ferrante2023brain} & \textcolor[rgb]{0.04453125, 0.54921875, 0.14453125}{a group of people on a field with a dog.} \\  
        OneLLM~\cite{han2024onellm} &  \textcolor[rgb]{0.9,0.8,0.5}{A man talking on a cell phone while skiing.} \\ 
        \methodSubject{S1} & \textcolor[rgb]{0.97265625, 0.55, 0.27265625}{A group of people riding skis on top of a snow covered slope.} \\ 
        \method{} & \textcolor[rgb]{0.8,0,0}{A group of people riding skis on top of a snow covered slope.} \\ 
        \end{tabular}}}
        \end{tcolorbox}
\end{minipage}
\caption{\textbf{Brain Captioning Comparison on S1.} Baselines for S1 include SDRecon~\cite{takagi2023improving}, BrainCap~\cite{ferrante2023brain}, and OneLLM~\cite{han2024onellm}. `\methodSubject{S1}' refers to our model trained only with subject S1, while `\method{}' denotes the model with cross-subject training.}
\label{fig:supmat_captioning_comparison}
\end{figure}

\condparagraph{More Results for S1.} In addition to the quantitative evaluation of brain captioning on S1 reported in the main paper, we report qualitative comparison in \cref{fig:supmat_captioning_comparison}. 
As shown, SDRecon~\cite{takagi2023improving} often produces incoherent and irrelevant descriptions, consistent with the lowest fluency and relevance metrics in the quantitative evaluation results.
OneLLM~\cite{han2024onellm} provides complete and coherent responses, but typically diverges from the content in the reference visual stimuli, exhibiting the lowest CLIP similarity scores~\cite{hessel2021clipscore} (CLIP-S and RefCLIP-S).
BrainCap~\cite{ferrante2023brain} yields better results compared to the above two methods, with fluency not being an issue but encountering challenges in content similarity with the reference.

Both of our methods, \method{} and \methodSubject{S1}, yield results with the most fluent and relevant descriptions. 
Here, `\methodSubject{Sx}' refers to our model trained with a single subject only and
`\method{}' is the unified brain decoding model with cross-subject training. %
When other methods fail to provide accurate descriptions, ours can accurately describe the scenes, such as parasailing, lake, mountain in row 1; building with clock tower in row 2; bird on a tree branch in row 5, skiing in the last row. 
Besides, baselines tend to provide approximate descriptions in contradiction to our method, such as for giraffe in row 3, sheep in row 6 (note \method{} even provides accurate counts), living room in the second last. 
Importantly, note that unlike the brain captioning SOTAs~\cite{han2024onellm,mai2023unibrain,takagi2023improving,ferrante2023brain}, we do \textit{not} use any ground truth captions during the training of the brain encoder.\\

\condparagraph{Results for S2, S5, S7.} \cref{tab:supmat_captioning} shows the brain captioning results on S2, S5, S7. We compare our single-subject model \methodSubject{Sx} and cross-subject model \method{} with two state-of-the-art methods SDRecon~\cite{takagi2023improving} and BrainCap~\cite{ferrante2023brain}.
Results for other subjects in OneLLM~\cite{han2024onellm} and UniBrain~\cite{mai2023unibrain} are unavailable: OneLLM~\cite{han2024onellm} solely trains with S1, and UniBrain~\cite{mai2023unibrain} has not been open-sourced, with no reported results for other subjects. The results on S2, S5, and S7 show consistent performance with those presented for S1 in the main paper. 
Remarkably, in all metrics and subjects, our methods perform better than the baselines~\cite{takagi2023improving,ferrante2023brain}.
The models using cross-subject training (\method{}) also generally perform better than those trained on a single subject (\methodSubject{Sx}).

\begin{figure}[!th]
	\centering
    \begin{minipage}{0.172\textwidth}
        \scriptsize
        \centering
        \textbf{Image}\\
        \textit{\tiny(for reference only)}
    \end{minipage}
    \begin{minipage}{0.815\textwidth}
        \scriptsize
        \centering
        \textbf{Brain captioning}
    \end{minipage}\\
\begin{minipage}{0.175\textwidth}
        \centering
        \includegraphics[width=\textwidth]{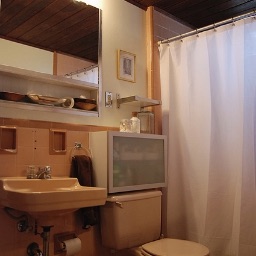}
    \end{minipage}
    \begin{minipage}{0.815\textwidth}
    \begin{tcolorbox}[colback=white!100,left=1pt,
                  right=1pt] {
        \centering
        \resizebox{\textwidth}{!}{
        \begin{tabular}{ll}
        COCO~\cite{lin2014microsoft} & {A bathroom with a vanity mirror sitting above a toilet next to a bathtub.}  \qquad \\ 
        Shikra-w/img~\cite{chen2023shikra} & {A bathroom with a toilet, sink and a television.} \\  
        S1 & {A bathroom with a toilet, sink and mirror.} \\ 
        S2 & {A bathroom with a sink, mirror and toilet.} \\ 
        S5 & {A kitchen with a stove, sink, and cabinets} \\ 
        S7 & {A bathroom with a toilet, sink and bathtub.} \\ 
        \end{tabular}}}
        \end{tcolorbox}
\end{minipage} \\
\begin{minipage}{0.175\textwidth}
        \centering
        \includegraphics[width=\textwidth]{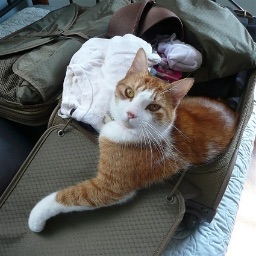}
    \end{minipage}
    \begin{minipage}{0.815\textwidth}
    \begin{tcolorbox}[colback=white!100,left=1pt,
                  right=1pt] {
        \centering
        \resizebox{\textwidth}{!}{
        \begin{tabular}{ll}
        COCO~\cite{lin2014microsoft} & {A picture of a cat and some luggage.}  \qquad \qquad \qquad \qquad \qquad \qquad \qquad \qquad \qquad \\ 
        Shikra-w/img~\cite{chen2023shikra} & {A cat sitting on a suitcase with clothes on a table.} \\  
        S1 & {A cat is sitting on top of a closed suitcase.} \\ 
        S2 & {A cat is laying down on a soft surface.} \\ 
        S5 & {A cat is laying on top of a bed.} \\ 
        S7 & {A cat laying on top of a bed in a room.} \\ 
        \end{tabular}}}
        \end{tcolorbox}
\end{minipage} \\
\begin{minipage}{0.175\textwidth}
        \centering
        \includegraphics[width=\textwidth]{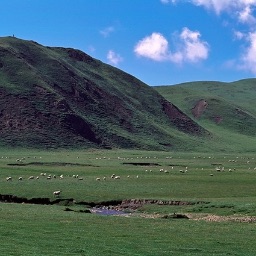}
    \end{minipage}
    \begin{minipage}{0.815\textwidth}
    \begin{tcolorbox}[colback=white!100,left=1pt,
                  right=1pt] {
        \centering
        \resizebox{\textwidth}{!}{
        \begin{tabular}{ll}
        COCO~\cite{lin2014microsoft} & {A large field of grass with sheep grazing on the land.}  \qquad \qquad \qquad \qquad \qquad \qquad \\ 
        Shikra-w/img~\cite{chen2023shikra} & {A herd of sheep graze in a lush green field.} \\  
        S1 & {A large mountain range filled with lots of trees.} \\ 
        S2 & {The image shows a great wilderness of mountains.} \\ 
        S5 & {A large mountain range is shown with a sky in the background.} \\ 
        S7 & {A large field with a mountain range in the background.} \\ 
        \end{tabular}}}
        \end{tcolorbox}
\end{minipage} \\
\begin{minipage}{0.175\textwidth}
        \centering
        \includegraphics[width=\textwidth]{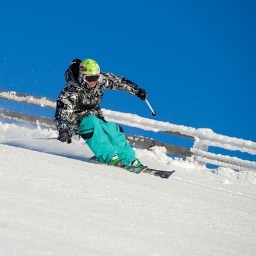}
    \end{minipage}
    \begin{minipage}{0.815\textwidth}
    \begin{tcolorbox}[colback=white!100,left=1pt,
                  right=1pt] {
        \centering
        \resizebox{\textwidth}{!}{
        \begin{tabular}{ll}
        COCO~\cite{lin2014microsoft} & {A man riding a snowboard down a hill.}  \qquad \qquad \qquad \qquad \qquad \qquad \qquad \qquad \\ 
        Shikra-w/img~\cite{chen2023shikra} & {A skier is going down a snowy hill.} \\ 
        S1 & {A person in a ski outfit skiing down a slope.} \\ 
        S2 & {A man riding a surfboard on top of a wave.} \\ 
        S5 & {A person on skis is skiing on a snowy slope.} \\ 
        S7 & {A person riding a snowboard on top of a snow covered slope.} \\ 
        \end{tabular}}}
        \end{tcolorbox}
\end{minipage} \\
\begin{minipage}{0.175\textwidth}
        \centering
        \includegraphics[width=\textwidth]{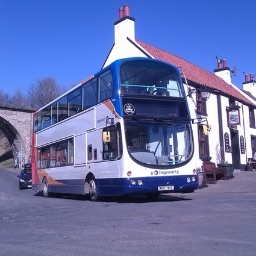}
    \end{minipage}
    \begin{minipage}{0.815\textwidth}
    \begin{tcolorbox}[colback=white!100,left=1pt,
                  right=1pt] {
        \centering
        \resizebox{\textwidth}{!}{
        \begin{tabular}{ll}
        COCO~\cite{lin2014microsoft} & {Double decker bus on the street next to buildings.}  \qquad \qquad \qquad \qquad \qquad \qquad \\ 
        Shikra-w/img~\cite{chen2023shikra} & {A double decker bus is parked outside a building.} \\ 
        S1 & {A transit bus riding down a street with buildings around.} \\ 
        S2 & {A passenger bus that is driving down the street.} \\ 
        S5 & {A large bus is traveling down the street.} \\ 
        S7 & {A bus driving down the street near another bus.} \\ 
        \end{tabular}}}
        \end{tcolorbox}
\end{minipage} \\
\begin{minipage}{0.175\textwidth}
        \centering
        \includegraphics[width=\textwidth]{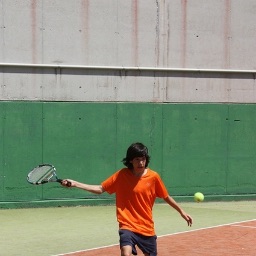}
    \end{minipage}
    \begin{minipage}{0.815\textwidth}
    \begin{tcolorbox}[colback=white!100,left=1pt,
                  right=1pt] {
        \centering
        \resizebox{\textwidth}{!}{
        \begin{tabular}{ll}
        COCO~\cite{lin2014microsoft} & {A person holding a tennis racket in their hand.} \qquad \qquad \qquad \qquad \qquad \qquad \\ 
        Shikra-w/img~\cite{chen2023shikra} & {A young man in an orange shirt playing tennis.} \\ 
        S1 & {A woman holding a tennis racquet on top of a tennis court.} \\ 
        S2 & {A woman holding a tennis racket on a tennis court.} \\ 
        S5 & {A woman standing on a tennis court holding a racket.} \\ 
        S7 & {A man holding a tennis racquet on a tennis court.} \\ 
        \end{tabular}}}
        \end{tcolorbox}
\end{minipage} \\
	\caption{\textbf{Brain Captioning Results on Different Subjects.} `COCO' is the ground truth caption in the COCO dataset~\cite{lin2014microsoft}. We excerpt the first caption if there are multiple captions for the same image. Shikra-w/img~\cite{chen2023shikra} is the result using the ground truth images as input. Results for all four subjects (S1, S2, S5, and S7) are from our cross-subject~\method{} model.
 }
	\label{fig:supmat_captioning_subjects}
 \vspace{2em}
\end{figure}

\condparagraph{Subject Comparison.} \cref{fig:supmat_captioning_subjects} shows brain captioning results on different subjects. The first lines are ground truth captions from COCO~\cite{lin2014microsoft} for comparison. 
When an image comes with multiple ground truth captions, we always select the one. The second lines display the results from Shikra~\cite{chen2023shikra} using images as input, which can be seen as an approximate upper bound of performance for our method. 
Our method shares the same adapter and the finetuned LLM with Shikra in most experiments, but uses a different encoder (brain encoder instead of image encoder) and input modality (brain response instead of image).
The following lines show the results for different subjects (S1, S2, S5, and S7).  Results for these subjects are all from a single \method{} model with subject-specific training using only the brain responses as input.
Our method achieves comparable quality and relevance even when compared to ground truth, or the image captioning method using images as input.

\setlength{\tabcolsep}{4pt}
\setlength{\fboxrule}{0pt} 
\setlength{\fboxsep}{2pt}
\begin{table}[th]
\caption{\textbf{Brain Captioning}. `\methodSubject{Sx}' refers to our model trained with a single subject  only, while \method{}' denotes the model with cross-subject training.
The results of S1 have been presented in the main paper and are listed here for completeness.
The colors represent the \tgold{best}, \tsilver{second-best}, and \tbronze{third-best} performance.
Note that both our models are consistently surpassing the baselines and, furthermore, our cross-subject model (\method{}) is almost always better than its single-subject counterpart.
}
\label{tab:supmat_captioning}
\vspace{-2.5mm}
\centering
\resizebox{\textwidth}{!}{
\begin{tabular}{@{}l|c|cccccccccc@{}}
\toprule
Method & Eval & BLEU1 & BLEU2 & BLEU3 & BLEU4 & METEOR & ROUGE & CIDEr  & SPICE  & CLIP-S & RefCLIP-S\\
\midrule
SDRecon~\cite{takagi2023high} & \multirow{4}{*}{S1} & 36.21 & 17.11 & 7.72 & 3.43 & 10.03 & 25.13  & 13.83 & 5.02 & 61.07 & 66.36   \\
BrainCap~\cite{ferrante2023brain} & ~ & \bronze{55.96} & \bronze{36.21} & \bronze{22.70} & \bronze{14.51} & \bronze{16.68} & \bronze{40.69} & \bronze{41.30} & \bronze{9.06} & \bronze{64.31} & \bronze{69.90} \\
\methodSubject{S1} & ~ & \silver{57.63} & \silver{38.02} & \silver{25.00}  & \silver{16.76}   & \silver{18.41}   &  \silver{42.15}  & \silver{51.93} & \silver{11.83} & \silver{66.44} & \silver{72.12}   \\
\method{} & ~   & \gold{59.44} & \gold{40.48} & \gold{27.66} & \gold{19.03} & \gold{19.45} & \gold{43.71}  & \gold{61.06} & \gold{12.79} &  \gold{67.78}   &    \gold{73.54}  \\
\midrule
SDRecon~\cite{takagi2023improving} & \multirow{4}{*}{S2} & 34.71 & 15.87 & 6.72 & 3.02 & 9.60 & 24.22 & 13.38 & 4.58 & 59.52 & 65.30 \\
BrainCap~\cite{ferrante2023brain} & ~ & \bronze{53.80} & \bronze{34.16} & \bronze{20.86} & \bronze{13.03} & \bronze{15.90} & \bronze{39.96} & \bronze{35.60} & \bronze{8.47} & \bronze{62.48} & \bronze{68.19} \\
\methodSubject{S2}  & ~ & \silver{57.18} & \silver{37.76} & \silver{25.06} & \silver{17.18} & \silver{18.11} & \silver{41.85} & \silver{50.62} & \silver{11.50} & \silver{64.87} & \silver{71.06} \\
\method{}  & ~ & \gold{59.37} & \gold{40.47} & \gold{27.14} & \gold{18.41} & \gold{19.17} & \gold{43.86} & \gold{55.93} & \gold{12.08} & \gold{66.46} & \gold{72.36} \\
\midrule
SDRecon~\cite{takagi2023improving} & \multirow{4}{*}{S5} & 34.96 & 16.39 & 7.36 & 3.49 & 9.93 & 24.77 & 13.85 & 5.19 & 60.83 & 66.30 \\
BrainCap~\cite{ferrante2023brain} & ~  &\bronze{55.28} & \bronze{35.71} & \bronze{22.62} & \bronze{14.62} & \bronze{16.45} & \bronze{40.87} & \bronze{41.05} & \bronze{9.24} & \bronze{63.89} & \bronze{69.64} \\
\methodSubject{S5}  & ~ & \silver{58.99} & \silver{39.88} & \silver{27.03} & \silver{18.73} & \silver{19.04} & \silver{43.30} & \silver{57.09} & \silver{12.70} & \silver{66.48} & \silver{72.69} \\
\method{}  & ~ & \gold{60.36} & \gold{41.27} & \gold{27.92} & \gold{19.03} & \gold{20.04} & \gold{44.81} & \gold{61.32} & \gold{13.19} & \gold{68.39} & \gold{74.11} \\
\midrule
SDRecon~\cite{takagi2023improving} & \multirow{4}{*}{S7} & 34.99 & 16.10 & 7.06 & 3.26 & 9.54 & 24.33 & 13.01 & 4.74 & 58.68 & 64.59 \\
BrainCap~\cite{ferrante2023brain} & ~ & \bronze{54.25} & \bronze{34.47} & \bronze{21.67} & \bronze{14.00} & \bronze{15.94} & \bronze{40.02} & \bronze{37.49} & \bronze{8.57} & \bronze{62.52} & \bronze{68.48} \\
\methodSubject{S7}  & ~ & \silver{55.71} & \silver{36.24} & \silver{23.62} & \silver{15.75} & \silver{17.51} & \silver{40.64} & \silver{47.07} & \silver{11.26} & \silver{63.66} & \silver{70.09} \\
\method{}  & ~ & \gold{57.20} & \gold{38.30} & \gold{25.49} & \gold{17.13} & \gold{18.29} & \gold{42.16} & \gold{52.73} & \gold{11.63} & \gold{65.90} & \gold{71.83} \\
\bottomrule
\end{tabular}
}
\end{table}
\setlength{\tabcolsep}{1.4pt}

\subsection{Brain Grounding}
\label{subsec:supmat_grouning}

Results of acc@0.3 and acc@0.7 on different subject (S1, S2, S5, and S7) and different categories (A, S, SC, SO, and I) are reported in \cref{tab:supmat_brain_grounding_all}. 
Results of acc@0.5 are reported in the main paper. The IoU values remain consistent with those presented in the main paper.

Generally, the cross-subject model outperforms the single-subject models.
These results are obtained using the grounding prompt `\texttt{Locate <expr> in <image> and provide its coordinates, please.}', where \texttt{<expr>} is the expression. 
In practice, to obrain the evaluation results shown in~\cref{tab:supmat_brain_grounding_all}, we assume there are several `known' concepts (\texttt{<expr>}) in the given image and formulate the task as detecting queried objects and returning their coordinates. This is known as the REC task in MLLMs. The example supported task prompts are shown in the second row of~\cref{tab:supmat_prompt}.

\begin{figure}[!th]
\vspace{-1.5em}
	\centering
	\includegraphics[width=\linewidth]{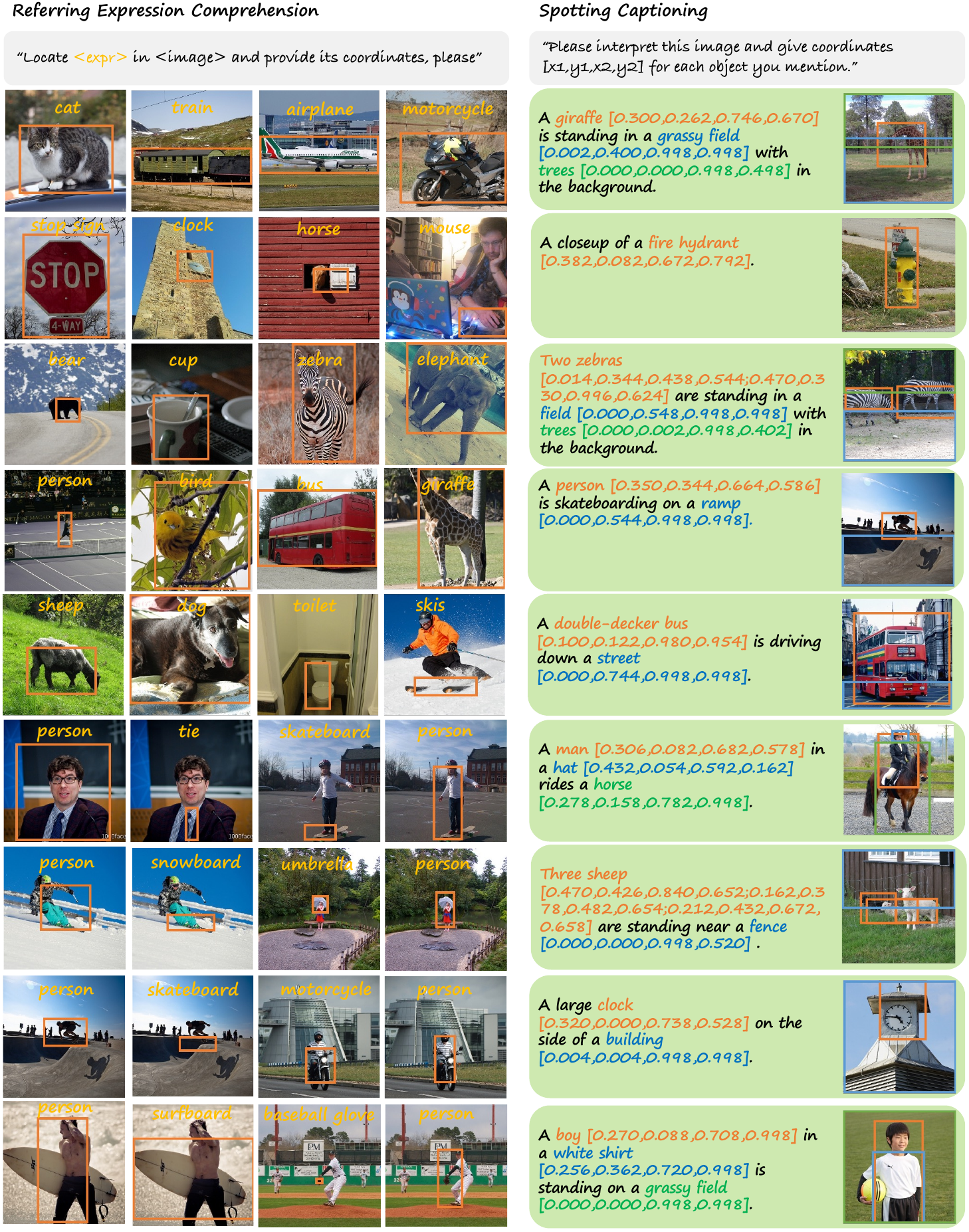}
	\caption{\textbf{Brain Grounding Results using Different Task Prompts.} The shown results are from our \method{} using brain responses as input. Reference images are visual stimuli for input brain responses and are just used here for visualization. The tags \texttt{<expr>} in the REC prompt are depicted in the corresponding reference images. The bounding boxes are outputted as text responses, and we visualize the outputs in the reference images.
    The tags \texttt{<expr>} in the REC prompt are displayed in the reference images with \textcolor{orange}{color}. The concepts and their coordinates in the Spotting Captioning responses are color-coded to match the bounding box color in the visualizations.
 }
	\label{fig:supmat_grounding_task}
\end{figure}

\begin{figure}[!th]
	\centering
	\includegraphics[width=\linewidth]{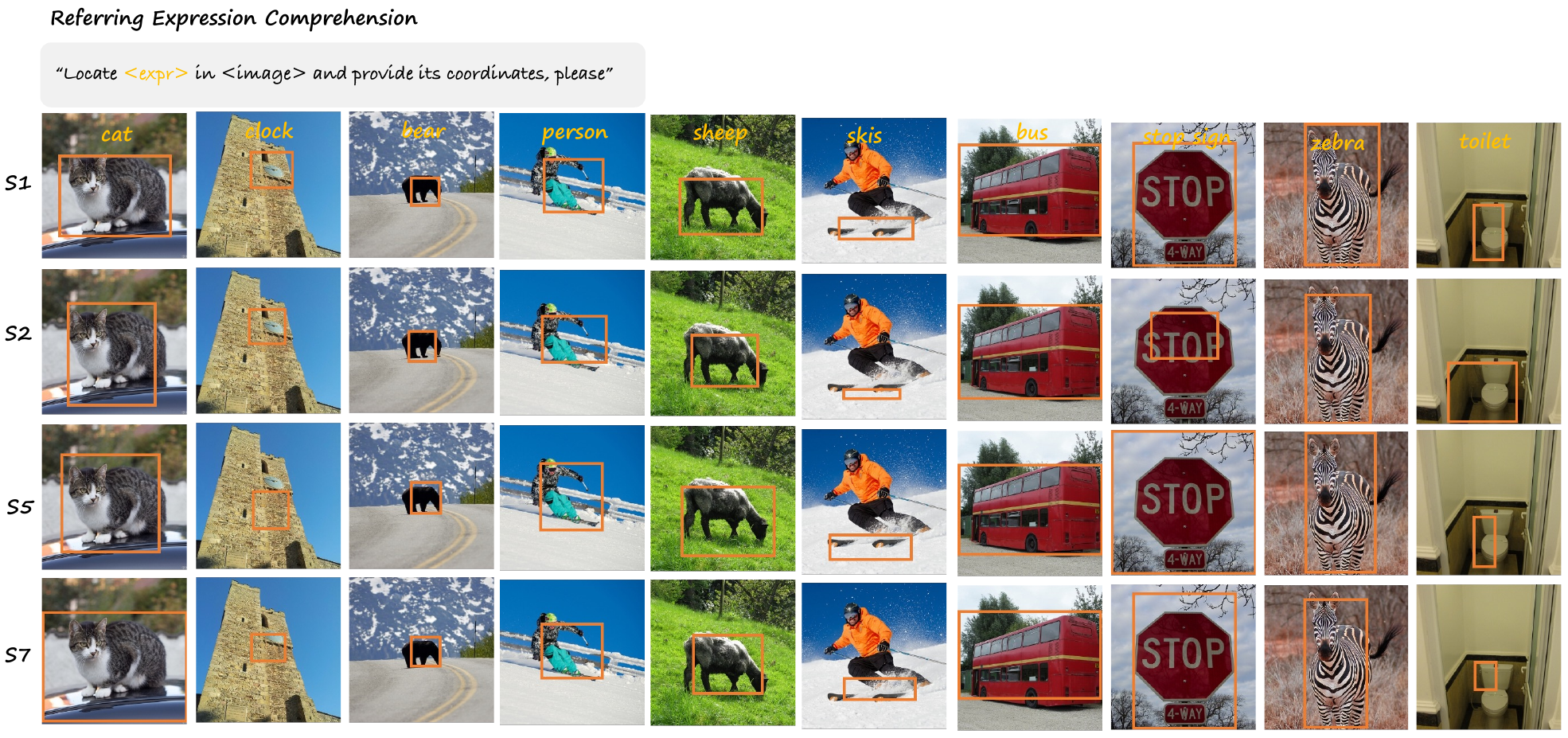}
	\caption{\textbf{Brain Grounding Result on Different Subjects.} All results are from our \method{} using brain responses as input. Reference images here are for visualization.
 }
	\label{fig:supmat_grounding_sub}
\end{figure}

\condparagraph{Brain Grounding without Priors.}
Interestingly, our method can also provide descriptions and coordinates for brain signals without `prior knowledge' of their contents. This task is referred to as `Spotting Captioning' in~\cite{chen2023shikra}, but it is a grounding-related task, as the primary goal is to describe the image (brain responses in our case) and identify the mentioned objects or regions using points or boxes.
In addition to the prompts shown in~\cref{tab:supmat_prompt}, we can also utilize a wide range of instructions such as 
`{\small \texttt{Please interpret this image and give coordinates [x1,y1,x2,y2] for each object you mention}}' and `{\small \texttt{Provide a detailed description of the image using around 100-500 words, including the objects, attributes, and spatial locations depicted in the picture}}'. 
The qualitative results of brain grounding, using various task prompts and across different subjects, are provided in~\cref{fig:supmat_grounding_task} and~\cref{fig:supmat_grounding_sub}, respectively. The bounding boxes are outputted as text responses, and we visualize the outputs in the reference images. The tags \texttt{<expr>} in the REC prompt are depicted in the corresponding reference images with \textcolor{orange}{color}. The concepts and coordinates in the responses of the Spotting Captioning task are depicted using the same color as the bounding box color in the visualizations. 

\setlength{\tabcolsep}{4pt}
\begin{table}[t]
\caption{\textbf{Brain Grounding Results of acc@0.3 and acc@0.7 on Different Subjects}. 
The accuracy with threshold $m$ is abbreviated as acc@m.
The IoU values remain consistent with those presented in the main paper, corresponding to each subject and category.
`\methodSubject{Sx}' refers to our model trained with a single subject only, while `\method{}' denotes the model with cross-subject training. Results of acc@0.5 are reported in the main paper. 
The best results per subject is in \tgold{color}.
}
\label{tab:supmat_brain_grounding_all}
\centering
\resizebox{\textwidth}{!}{
\begin{tabular}{@{}l|c|cc|cc|cc|cc|cc@{}}
\toprule
\multirow{2}{*}{Method} & \multirow{2}{*}{Eval}  & \multicolumn{2}{c|}{All (A)} & \multicolumn{2}{c|}{Salient (S)}  & \multicolumn{2}{c|}{Salient Creatures (SC)} & \multicolumn{2}{c|}{ Salient Objects (SO)}  & \multicolumn{2}{c}{Inconspicuous (I)} \\
~ & ~ & acc@0.3 & acc@0.7  & acc@0.3   & acc@0.7  & acc@0.3  & acc@0.7 & acc@0.3  & acc@0.7  & acc@0.3  & acc@0.7 \\
\midrule
\methodSubject{S1} & \multirow{2}{*}{S1} & 24.22 & 5.75 & 36.26 & 9.08 & 43.71 & 10.00 & 28.15 & 8.09 & 9.20 & 1.58 \\
\method{} & ~  & \gold{30.47} & \gold{8.47} & \gold{44.45} & \gold{13.55} & \gold{55.86} & \gold{16.14} & \gold{32.04} & \gold{10.73} & \gold{13.01} & \gold{2.14} \\
\midrule
\methodSubject{S2} & \multirow{2}{*}{S2} & 26.00 & 6.53 & 39.09 & 10.35 & 47.43 & 11.43 & \gold{30.02} & \gold{9.18} & 9.67 & 1.77 \\
\method{} & ~ & \gold{29.60} & \gold{7.94} & \gold{42.96} & \gold{12.58} & \gold{55.14} & \gold{16.00} & 29.70 & 8.86 & \gold{12.92} & \gold{2.14} \\
\midrule
\methodSubject{S5} & \multirow{2}{*}{S5} & 26.04 & 5.99 & 39.09 & 9.23 & 45.57 & 9.14 & 32.04 & \gold{9.33} & 9.76 & 1.95 \\
\method{} & ~ & \gold{30.05} & \gold{7.28} & \gold{44.75} & \gold{11.47} & \gold{56.14} & \gold{14.29} & \gold{32.35} & 8.40 & \gold{11.71} & \gold{2.04} \\
\midrule
\methodSubject{S7} & \multirow{2}{*}{S7} & 25.47 & {5.21} & 37.97 & 8.12 & 46.43 & 7.86 & 28.77 & 8.40 & {9.85} & 1.58 \\
\method{} & ~ & \gold{28.32} & \gold{7.03} & \gold{42.07} & \gold{10.80} & \gold{53.14} & \gold{12.86} & \gold{30.02} & \gold{8.55} & \gold{11.15} & \gold{2.32} \\
\bottomrule
\end{tabular}
}
\end{table}
\setlength{\tabcolsep}{1.4pt}

\subsection{Brain Retrieval}
\label{subsec:supmat_retrieval}

\begin{figure}[!t]
    \centering
    \resizebox{1.0\linewidth}{!}{%
    \setlength{\tabcolsep}{2pt}%
    \begin{tabular}{cc}
	\multicolumn{1}{c}{\tiny{Reference~~Top 1~~~~~Top~2~~~~~Top~3~~~~~Top~4~~~~Top 5~~~~}} & \multicolumn{1}{c}{\tiny{Reference~~Top 1~~~~~Top~2~~~~~Top~3~~~~~Top~4~~~~Top 5~~~~}}\\
    \begin{subfigure}[b]{0.485\textwidth}
        \includegraphics[width=\textwidth]{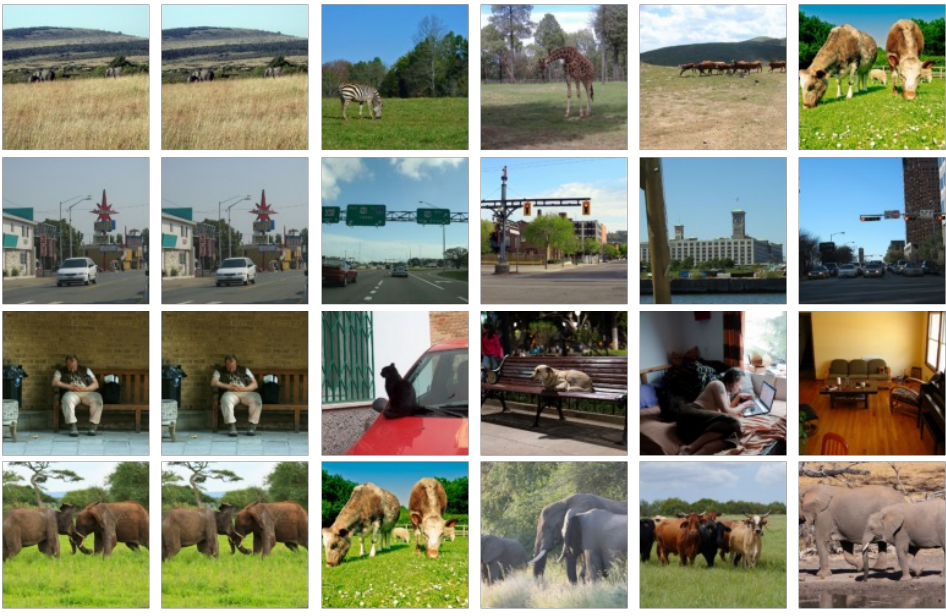}
    \end{subfigure}&
    \begin{subfigure}[b]{0.485\textwidth}
        \includegraphics[width=\textwidth]{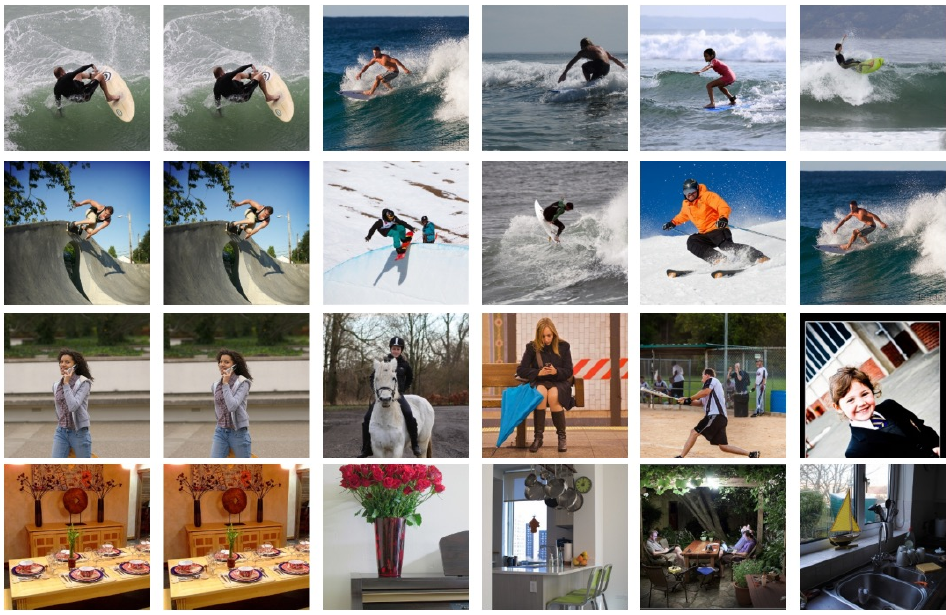}
    \end{subfigure}\\
    \cmidrule{1-2}
    \begin{subfigure}[b]{0.485\textwidth}
           \includegraphics[width=\textwidth]{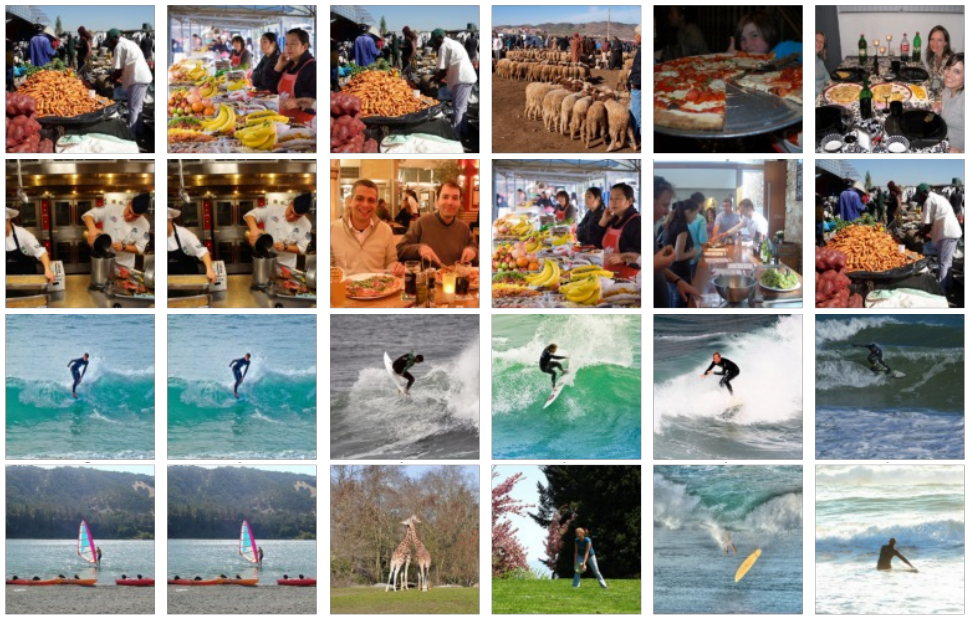}
    \caption{MindEye~\cite{scotti2023reconstructing}}
       \end{subfigure}&
    \begin{subfigure}[b]{0.485\textwidth}
           \includegraphics[width=\textwidth]{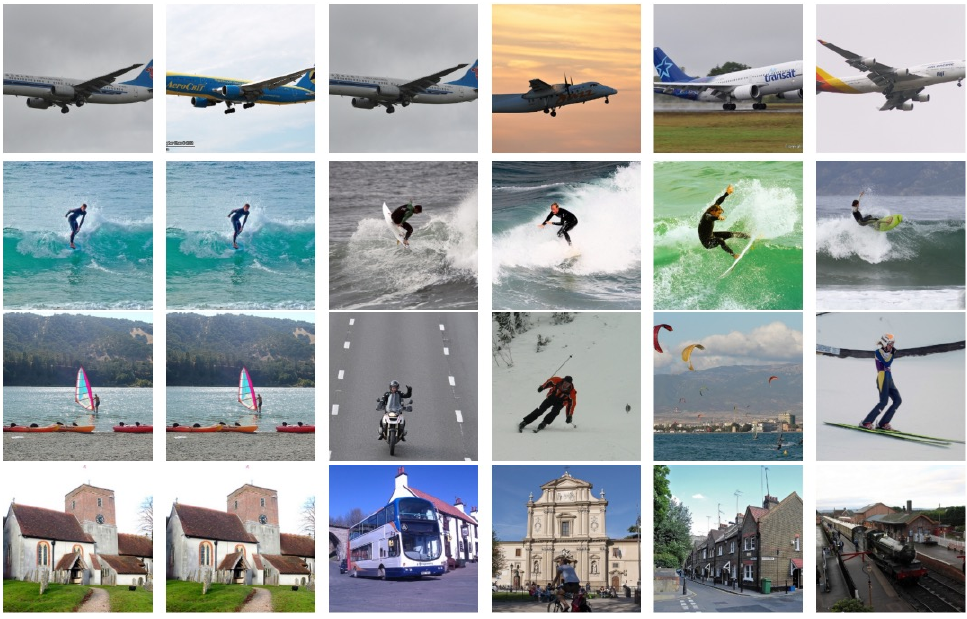}
           \caption{\method{}}
       \end{subfigure}\\
\end{tabular}%
}
\caption{\textbf{Brain Retrieval Results.} 
We show the forward and backward retrieval results of MindEye~\cite{scotti2023reconstructing} and our \method{}.
The forward retrieval~\cite{lin2022mind} (\textbf{top}) is to find the correct image embedding given a brain embedding. Conversely, the backward retrieval~(\textbf{bottom}) aims to locate the correct brain embedding given an image embedding.
}
\label{tab:supmat_retrieval}
\end{figure}

\cref{tab:supmat_retrieval} shows the forward and backward retrieval results using MindEye~\cite{scotti2023reconstructing} and our \method{}.
The images displayed in the top row are the reference image and the top 5 retrieval images obtained from the \textit{forward} retrieval~\cite{lin2022mind}. This process is to find the correct paired CLIP image embedding given a brain embedding.
Similarly, the bottom row are the reference image and the top 5 retrieval images obtained from the \textit{backward} retrieval process, which aims to locate the correct brain embedding given an image embedding.

\subsection{Visual Decoding}
\label{subsec:supmat_reconstruction}

\begin{figure}
\centering
\renewcommand{\arraystretch}{0.8}
\setkeys{Gin}{width=0.22\linewidth}
\setlength{\tabcolsep}{1.2pt}
\footnotesize
{
\resizebox{\textwidth}{!}{
\begin{tabular}[t]{ccccccccc}
\parbox[t]{2mm}{\multirow{1}{*}{\rotatebox[origin=c]{90}{Reference\hspace{-6em}}}} & \includegraphics{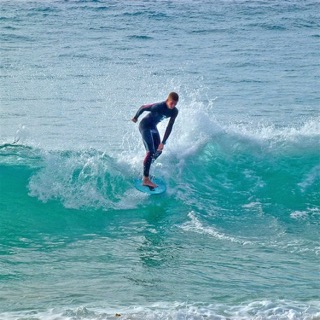} & \includegraphics{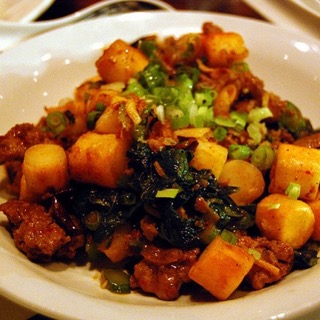} & \includegraphics{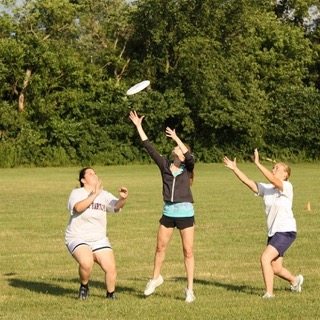} &
\includegraphics{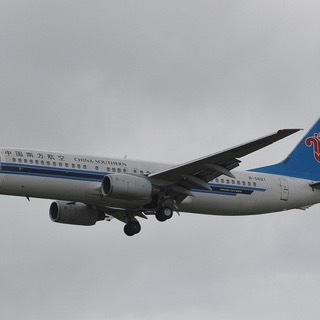} & \includegraphics{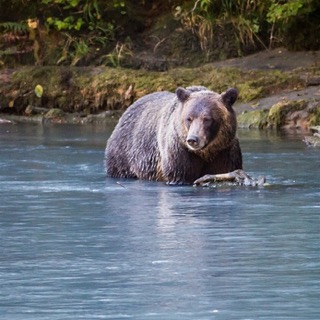} & \includegraphics{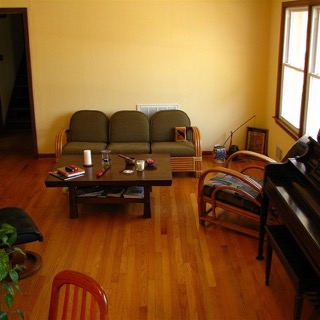} &
\includegraphics{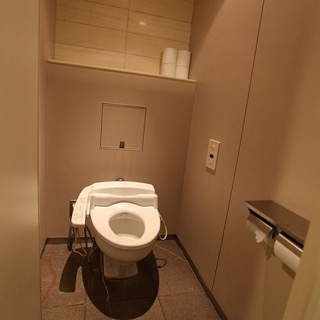} & \includegraphics{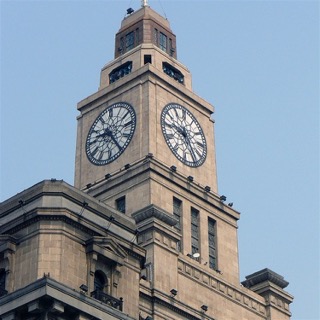} \\
\parbox[t]{2mm}{\multirow{1}{*}{\rotatebox[origin=c]{90}{SD~\cite{rombach2022high}\hspace{-6em}}}} & \includegraphics{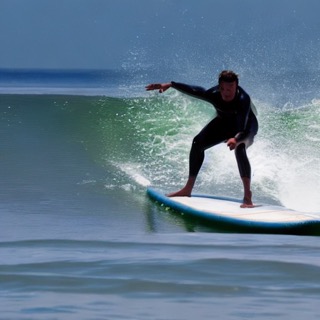} & \includegraphics{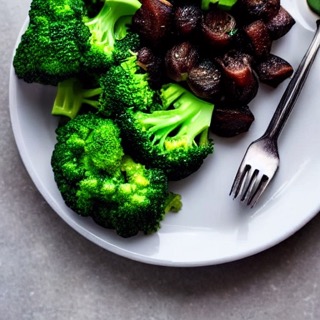} & \includegraphics{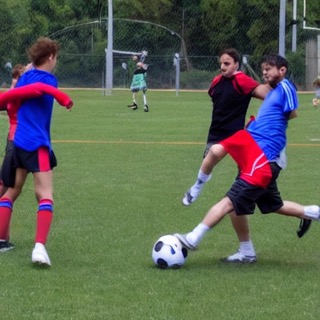} &
\includegraphics{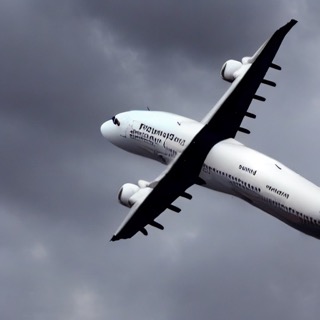} & \includegraphics{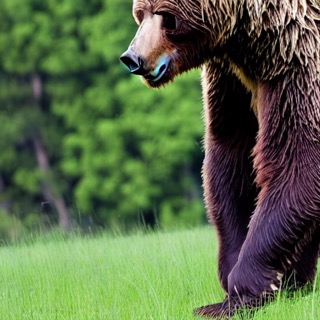} & \includegraphics{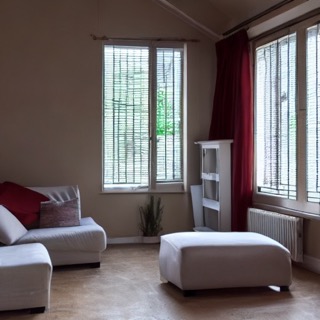} &
\includegraphics{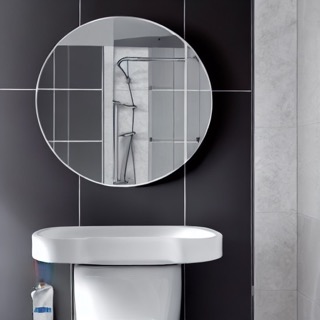} & \includegraphics{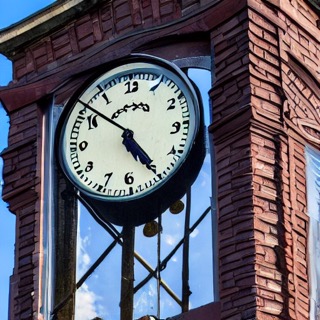} \\
\parbox[t]{2mm}{\multirow{1}{*}{\rotatebox[origin=c]{90}{SDXL~\cite{podell2023sdxl}\hspace{-6em}}}} & \includegraphics{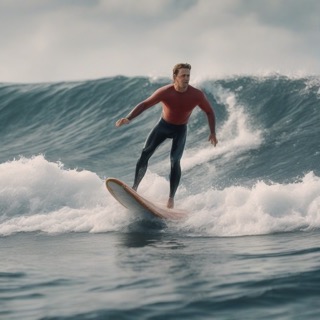} & \includegraphics{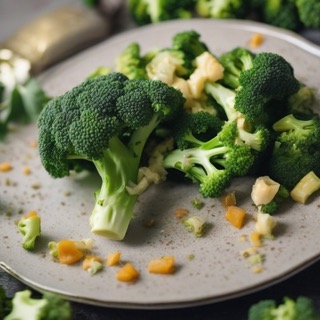} & \includegraphics{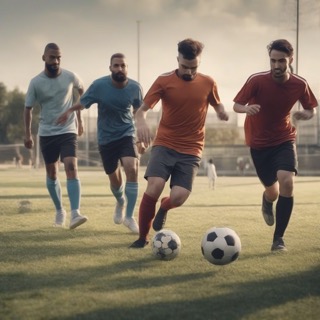} &
\includegraphics{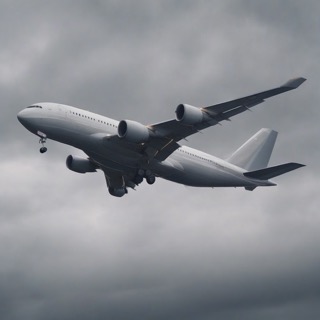} & \includegraphics{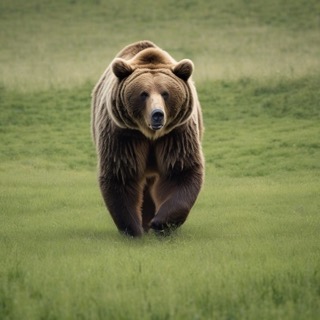} & \includegraphics{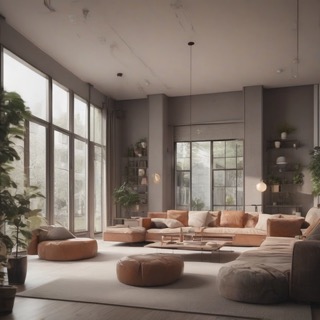} &
\includegraphics{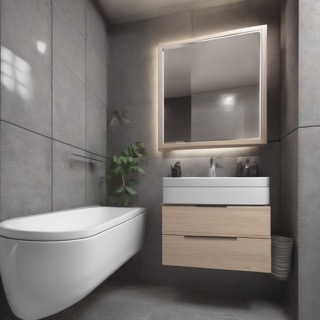} & \includegraphics{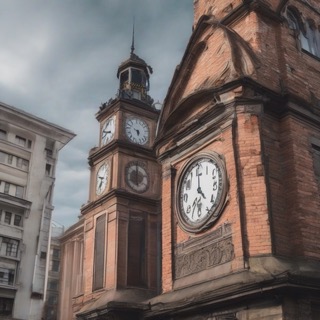} \\
\parbox[t]{2mm}{\multirow{1}{*}{\rotatebox[origin=c]{90}{Kandinsky3~\cite{arkhipkin2023kandinsky}\hspace{-6em}}}} & \includegraphics{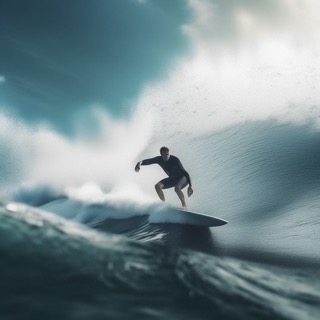} & \includegraphics{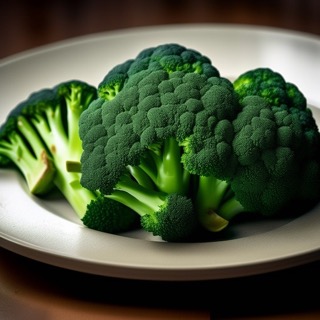} & \includegraphics{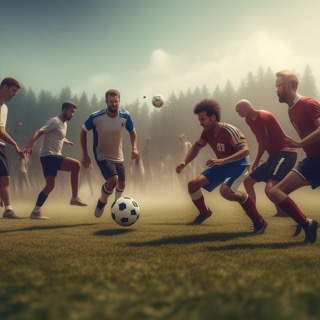} &
\includegraphics{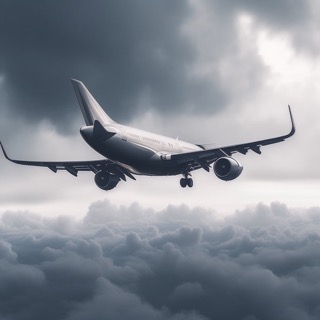} & \includegraphics{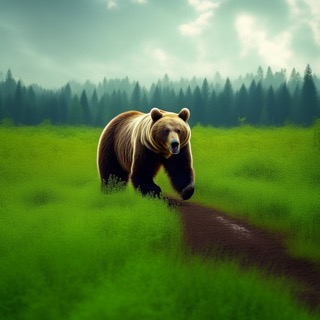} & \includegraphics{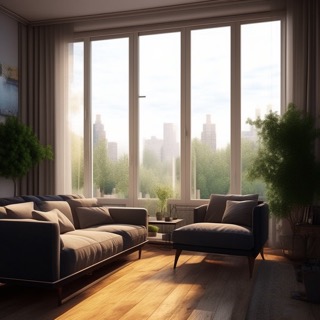} &
\includegraphics{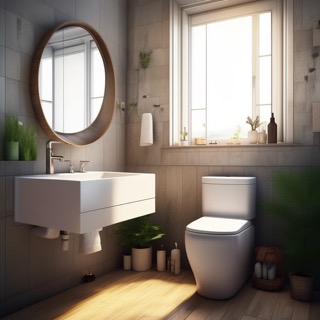} & \includegraphics{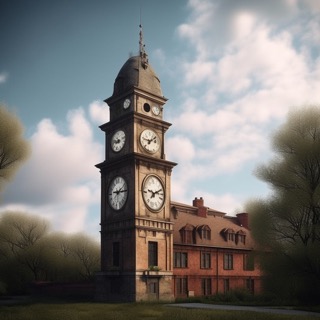} \\
\parbox[t]{2mm}{\multirow{1}{*}{\rotatebox[origin=c]{90}{GLIGEN~\cite{li2023gligen}\hspace{-6em}}}} & \includegraphics{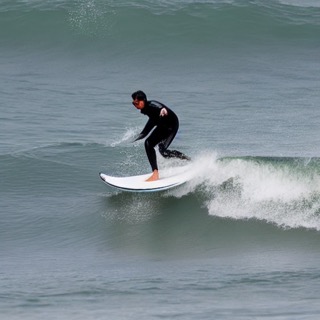} & \includegraphics{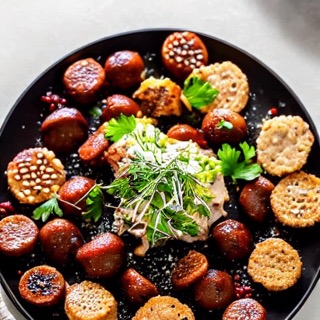} & \includegraphics{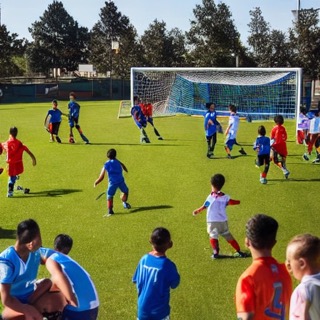} &
\includegraphics{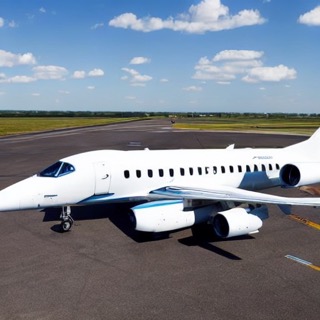} & \includegraphics{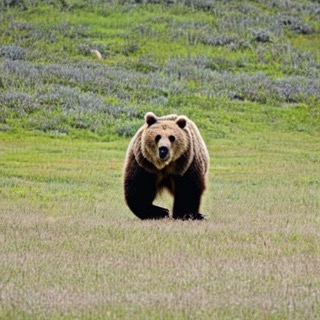} & \includegraphics{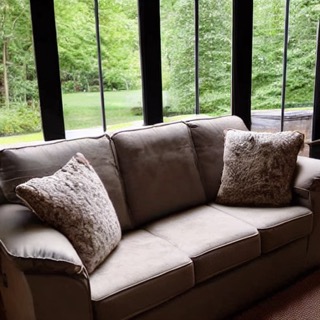} &
\includegraphics{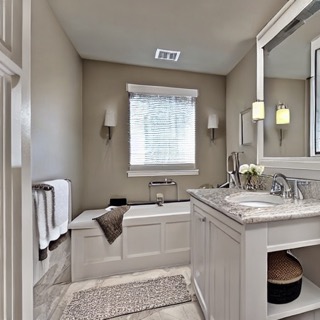} & \includegraphics{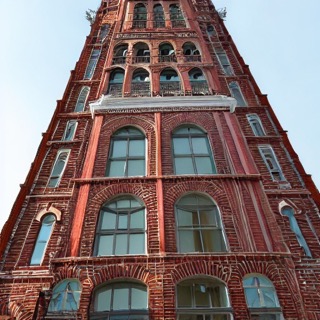} \\
\parbox[t]{2mm}{\multirow{1}{*}{\rotatebox[origin=c]{90}{VD~\cite{xu2022versatile}\hspace{-6em}}}} & \includegraphics{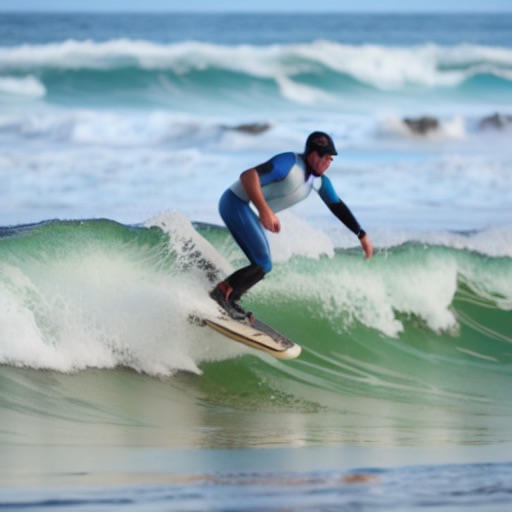} & \includegraphics{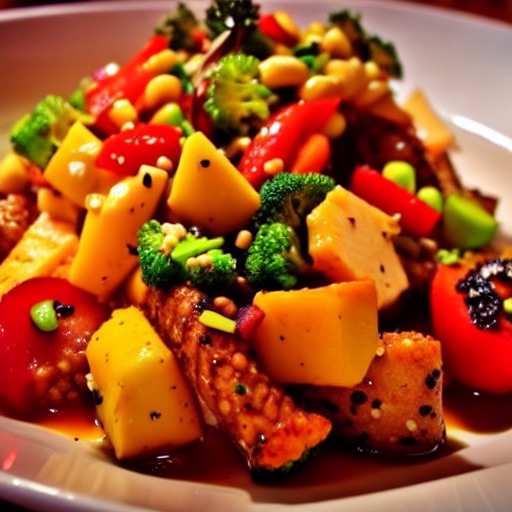} & \includegraphics{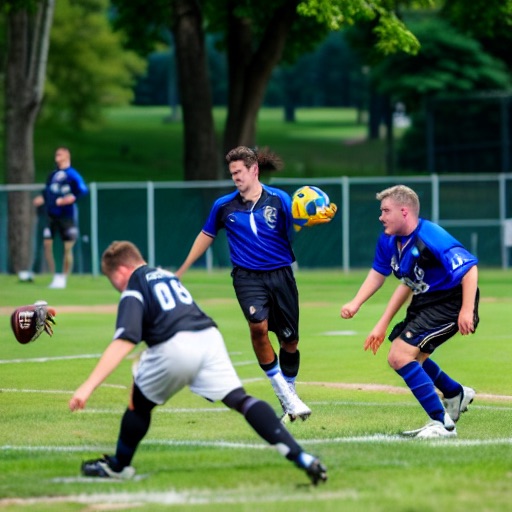} &
\includegraphics{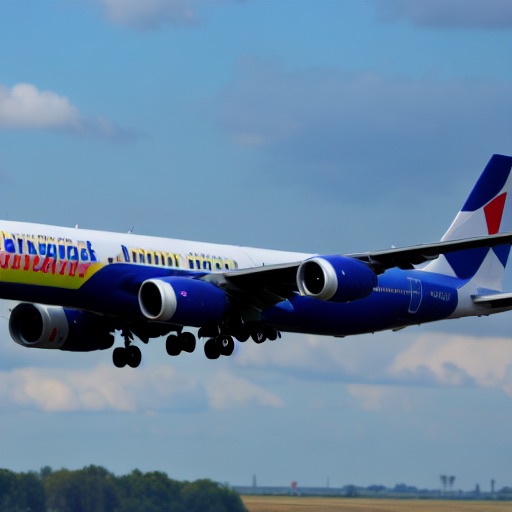} & \includegraphics{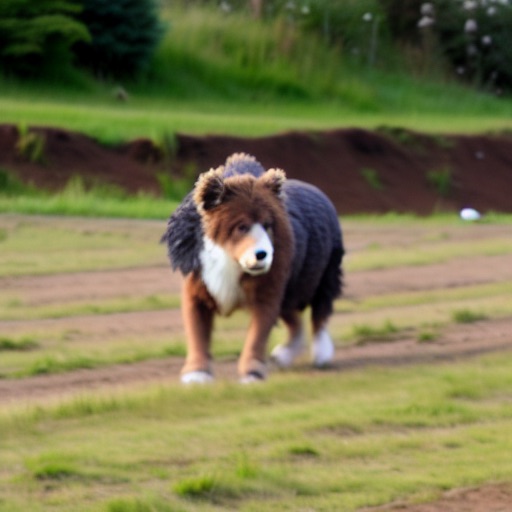} & \includegraphics{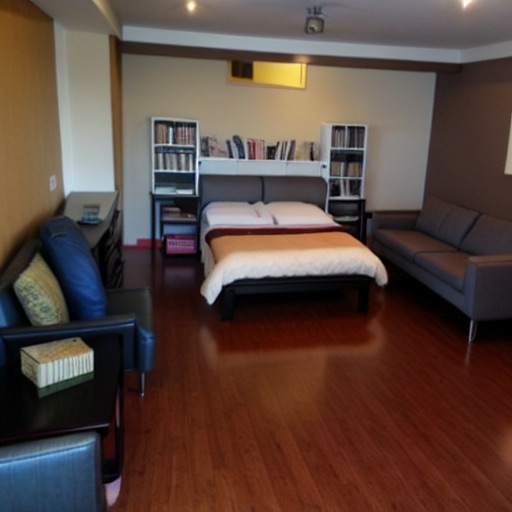} &
\includegraphics{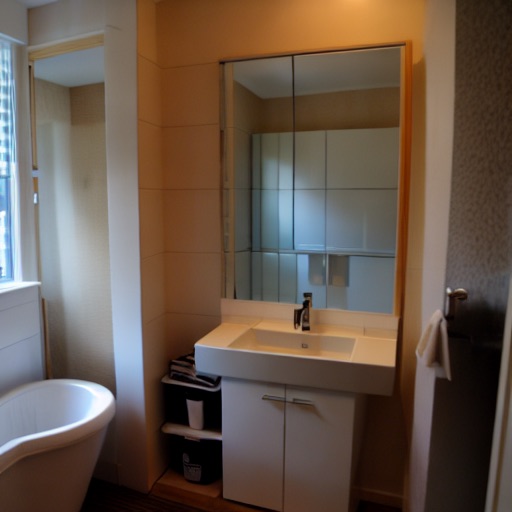} & \includegraphics{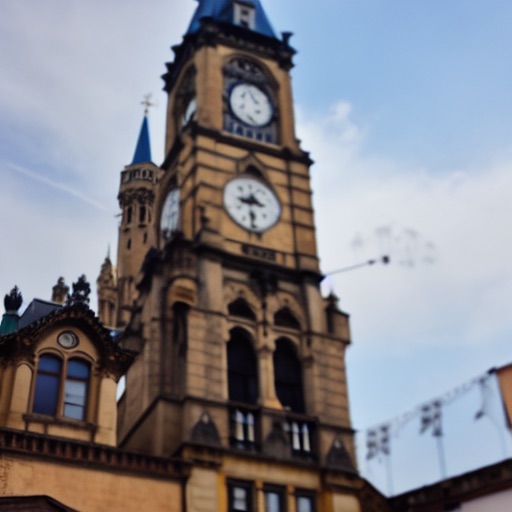} \\
\parbox[t]{2mm}{\multirow{1}{*}{\rotatebox[origin=c]{90}{VD-1 \hspace{-6em}}}} & \includegraphics{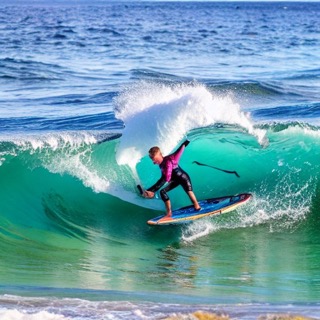} & \includegraphics{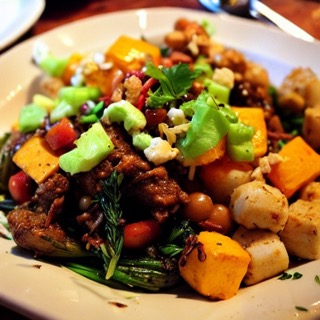} & \includegraphics{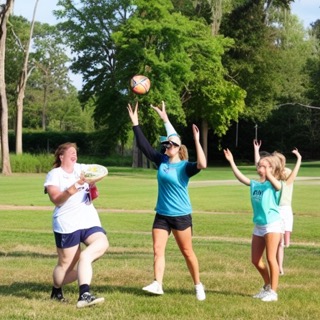} &
\includegraphics{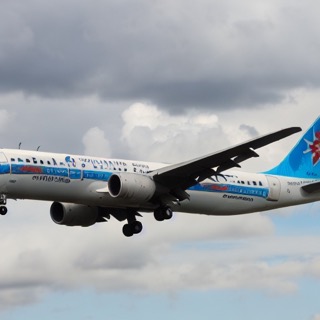} & \includegraphics{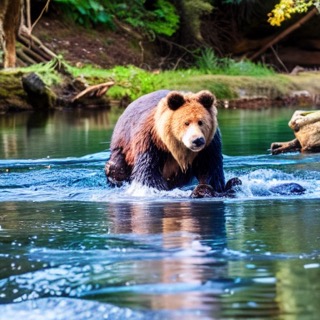} & \includegraphics{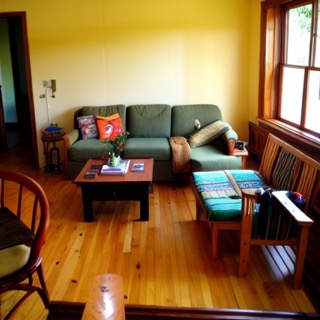} &
\includegraphics{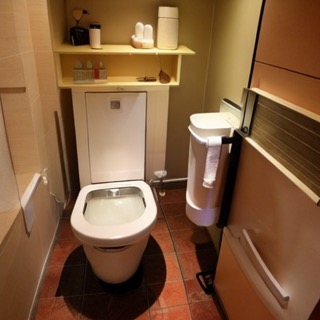} & \includegraphics{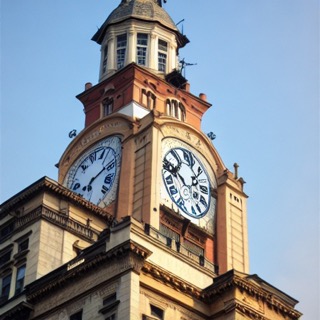} \\
\parbox[t]{2mm}{\multirow{1}{*}{\rotatebox[origin=c]{90}{VD-2 \hspace{-6em}}}} & \includegraphics{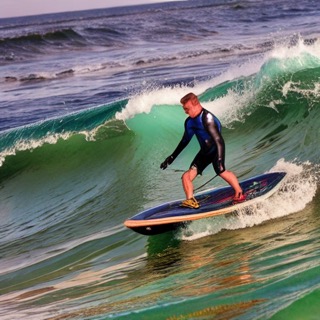} & \includegraphics{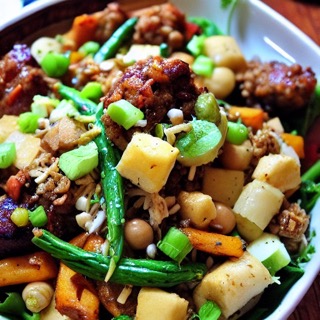} & \includegraphics{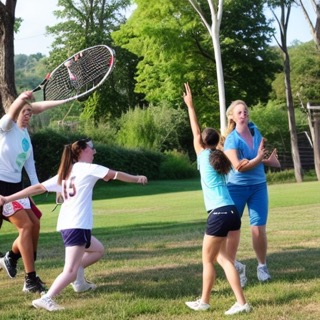} &
\includegraphics{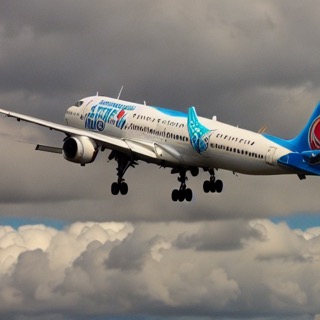} & \includegraphics{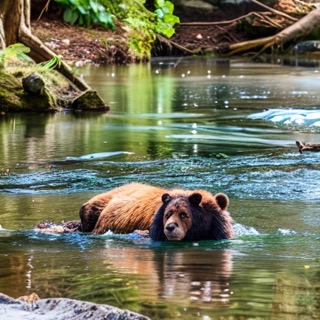} & \includegraphics{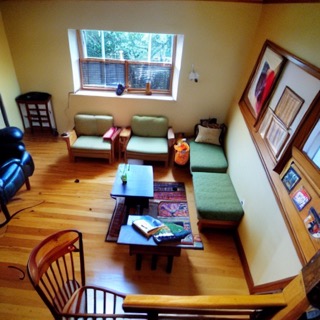} &
\includegraphics{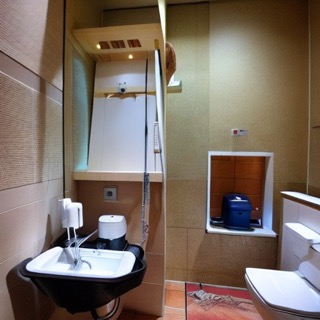} & \includegraphics{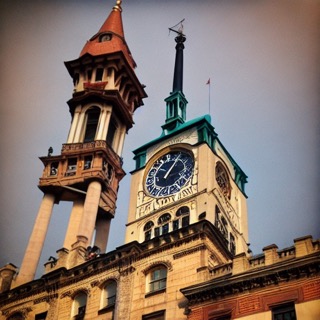} \\
\parbox[t]{2mm}{\multirow{1}{*}{\rotatebox[origin=c]{90}{VD-3\hspace{-6em}}}} & \includegraphics{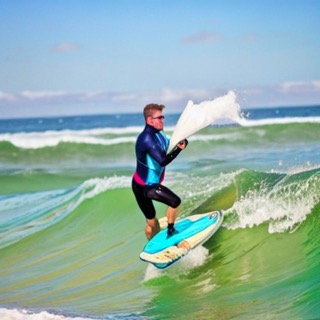} & \includegraphics{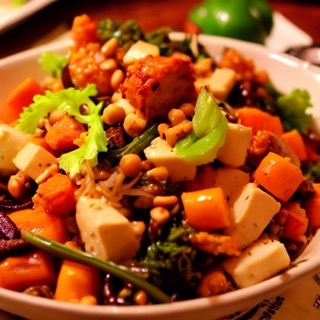} & \includegraphics{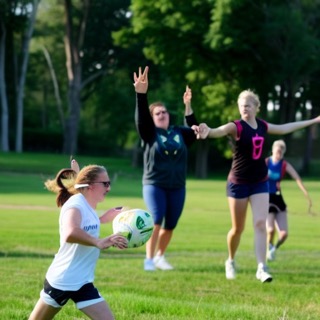} &
\includegraphics{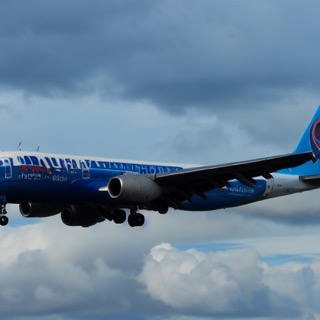} & \includegraphics{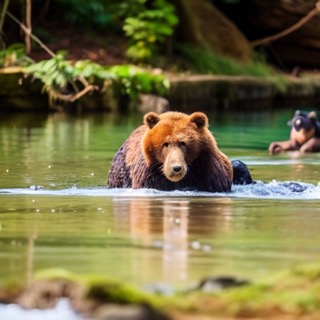} & \includegraphics{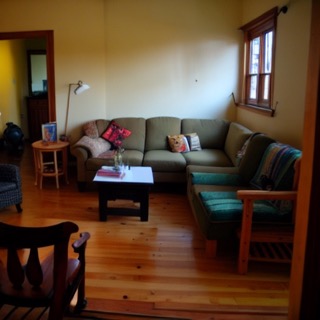} &
\includegraphics{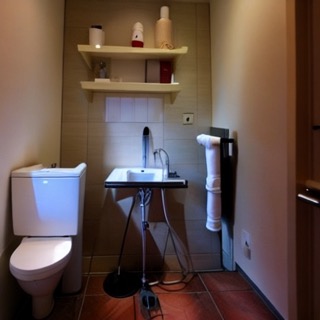} & \includegraphics{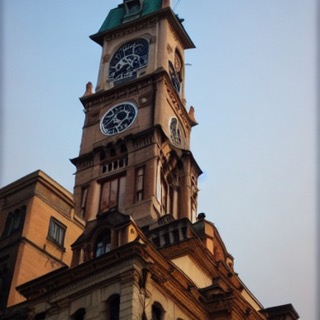} \\
\end{tabular}
}
\caption{\textbf{\method{} Visual Decoding with Various Image Generation Models.} 
We use textual and visual outputs predicted by our \method{} model as cues for the final image reconstruction, using a variety of pretrained image generation models. These models include text-to-image SD~\cite{rombach2022high}, SD-XL~\cite{podell2023sdxl}, and Kandinsky3~\cite{arkhipkin2023kandinsky}, a layout-to-image GlIGEN~\cite{li2023gligen}, and a multiple-condition VD~\cite{xu2022versatile}. Given that our method is based on VD~\cite{xu2022versatile}, we further analyze the effects of its inputs. VD-1 is using ground truth image and text, along with the latent code decoded from the ground truth image. VD-2 is similar but uses randomly sampled latent codes. VD-3 uses ground truth image and text, as well as predicted low-level images from Brain-Diffuser~\cite{ozcelik2023brain}.
}
\label{fig:supmat_visual_decoding_different_model}
}
\end{figure}

Despite that our method is not specifically designed for the task of visual decoding (fMRI-to-image reconstruction), our predicted textual and visual cues can be used for the final image reconstruction by using a variety of pretrained image generation models. To be specific, we use three text-to-image models SD~\cite{rombach2022high}, SD-XL~\cite{podell2023sdxl} and Kandinsky~\cite{arkhipkin2023kandinsky}, a layout-to-image model GlIGEN~\cite{li2023gligen}, and a multiple-condition model Versatile Diffusion (VD)~\cite{xu2022versatile}. The quantitative evaluation results are in~\cref{tab:supmat_visual_decoding_different_model}. Besides the common visual decoding metrics~\cite{ozcelik2023brain}, we further evaluate the image quality using FID~\cite{heusel2017gans} and Clean-FID (CFID)~\cite{parmar2022aliased}, as well as the image similarity using LPIPS~\cite{zhang2018unreasonable}.
The qualitative comparison is illustrated in~\cref{fig:supmat_visual_decoding_different_model}.

Given that our method is VD-based~\cite{xu2022versatile}, we further analyze the effects of its inputs, as shown in the last three rows in~\cref{tab:supmat_visual_decoding_different_model}.
For the dual context mode (generating an image conditioned on both text and image) of VD~\cite{xu2022versatile}, three parts can be used as inputs for the visual decoding task: the input text ($t$), the input image ($i$), and a latent code $z$ that can either be encoded from a given image or randomly sampled. We test the upper limits of performance for different inputs. Specifically, VD-1 is using ground truth image and text, along with the latent code decoded from the ground truth image. VD-2 is similar but uses randomly sampled latent codes. VD-3 uses ground truth image and text, as well as predicted low-level images from Brain-Diffuser~\cite{ozcelik2023brain}. As shown in~\cref{tab:supmat_visual_decoding_different_model}, latent codes $z$ in the latent diffusion-based image generation models play a crucial role in the final visual decoding performance. Simply using predicted text (SD, SDXL, and Kandinsky3) or adding predicted image embedding~(VD-2) is not sufficient for reliable visual decoding results. 
The predicted bounding boxes (GLIGEN) are intended to function as low-level constraints for the final reconstruction. However, they do not perform well on low-level metrics due to their unreliable nature.

\begin{figure}[th]
    \centering
    \includegraphics[width=\textwidth]{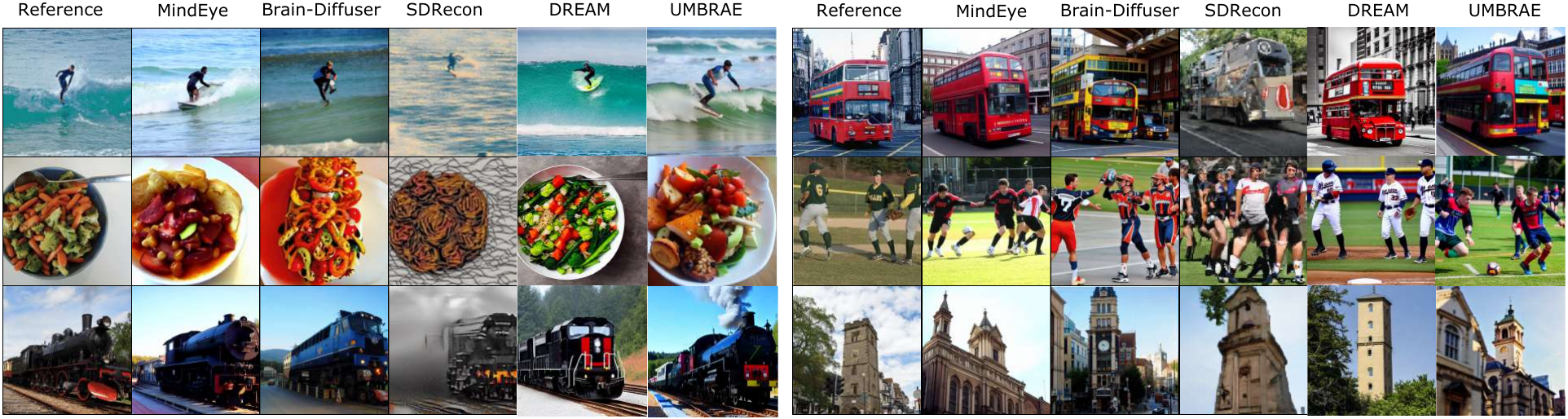}
    \caption{\revise{\textbf{Visual Decoding Comparison between \method{} and the literature on NSD.} All reconstructed images are from S1.
    }}
    \label{fig:supmat_comparison_vd_compressed}
\end{figure}

\begin{figure}[t]
\centering
\renewcommand{\arraystretch}{0.8}
\setkeys{Gin}{width=0.22\linewidth}
\setlength{\tabcolsep}{1.2pt}
\footnotesize
{
\resizebox{\textwidth}{!}{
\begin{tabular}[t]{ccccccccc}
\parbox[t]{2mm}{\multirow{1}{*}{\rotatebox[origin=c]{90}{Reference\hspace{-6em}}}} & \includegraphics{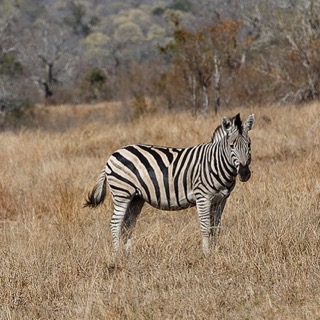} & \includegraphics{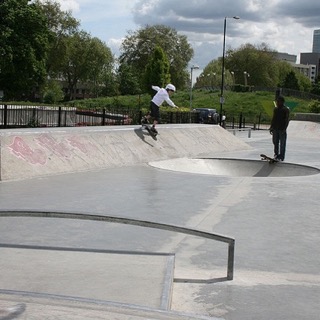} & \includegraphics{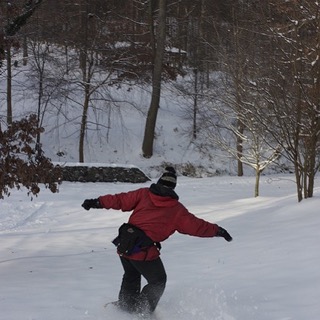}& \includegraphics{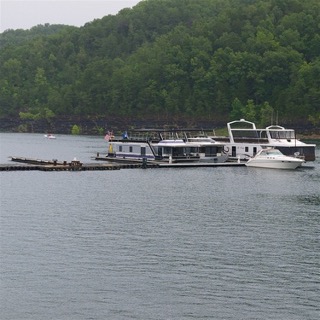} &
\includegraphics{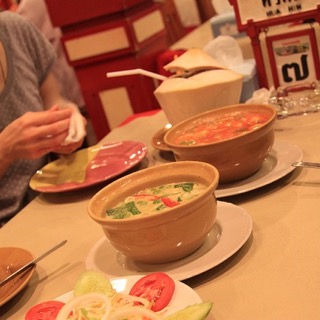} & \includegraphics{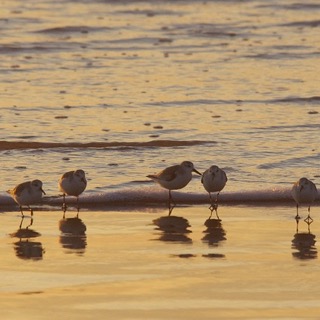} & \includegraphics{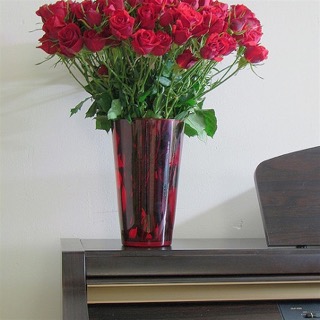} &
\includegraphics{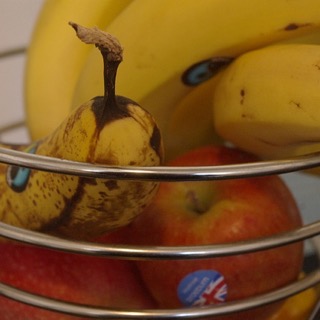} \\
\parbox[t]{2mm}{\multirow{1}{*}{\rotatebox[origin=c]{90}{S1\hspace{-6em}}}} & \includegraphics{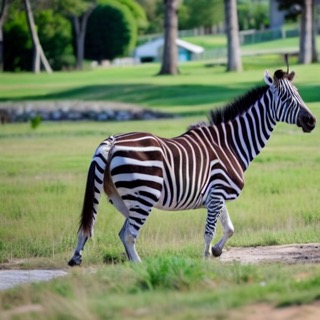} & \includegraphics{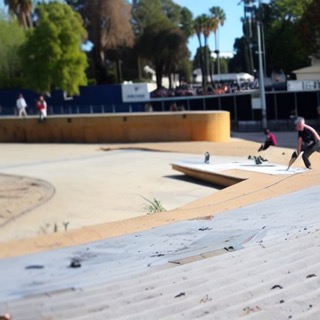} & \includegraphics{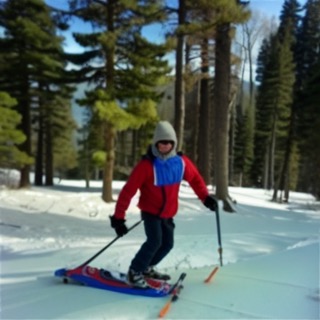}& \includegraphics{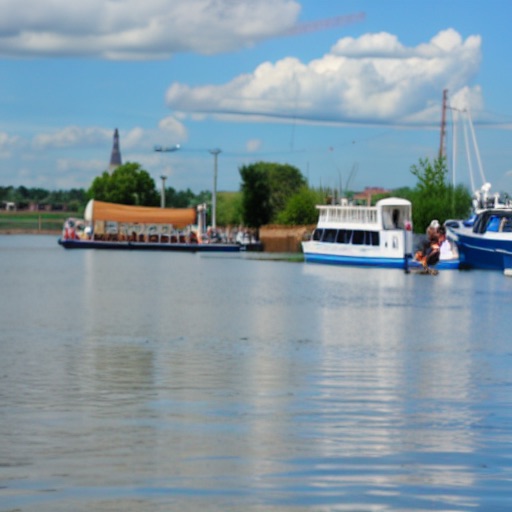} &
\includegraphics{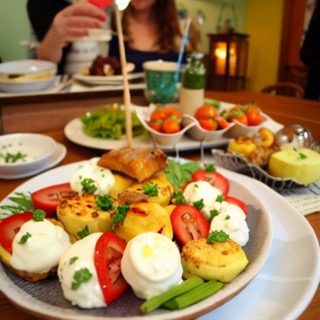} & \includegraphics{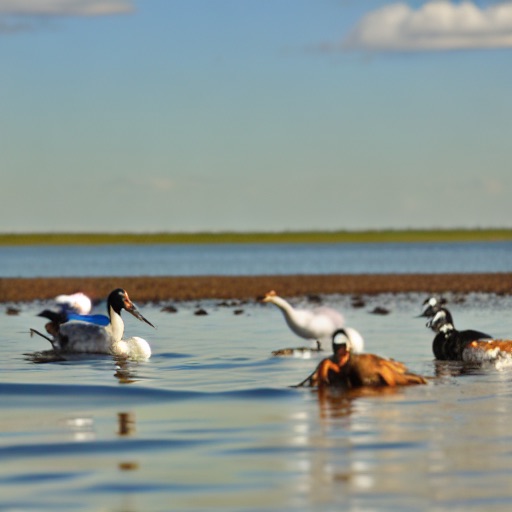} & \includegraphics{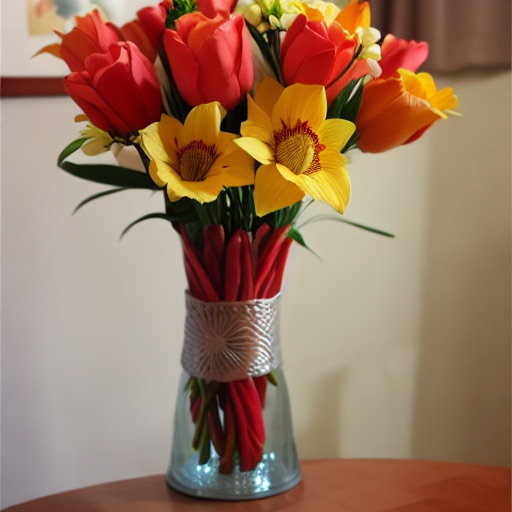} &
\includegraphics{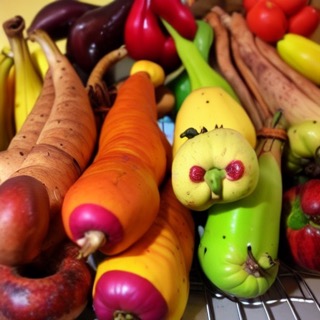} \\
\parbox[t]{2mm}{\multirow{1}{*}{\rotatebox[origin=c]{90}{S2\hspace{-6em}}}} & \includegraphics{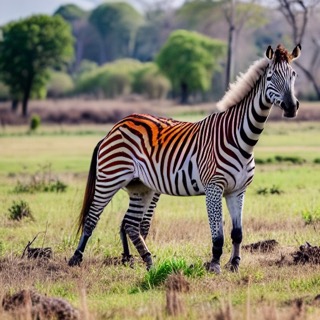} & \includegraphics{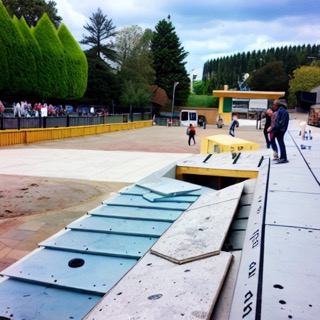} & \includegraphics{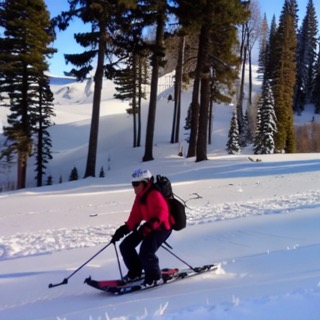}& \includegraphics{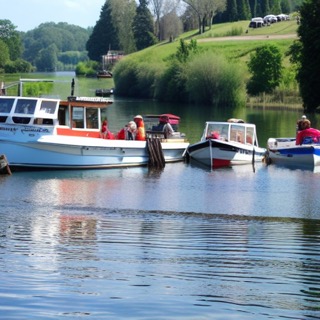} &
\includegraphics{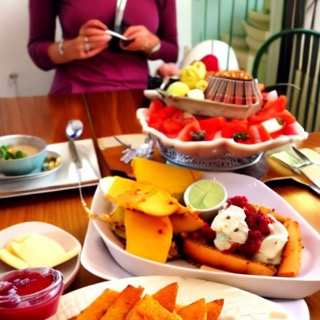} & \includegraphics{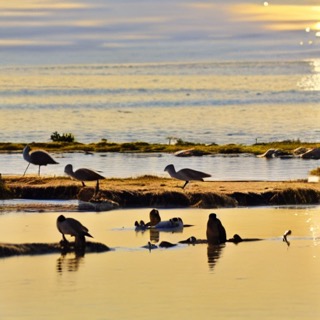} & \includegraphics{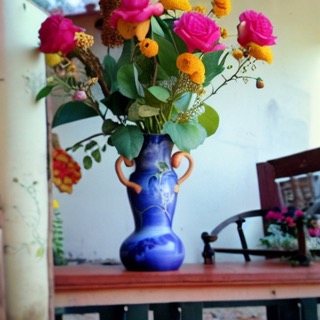} &
\includegraphics{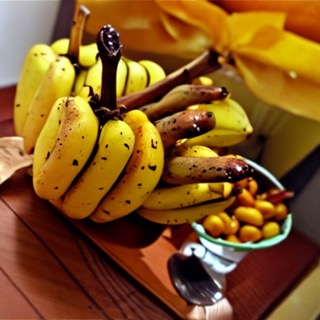} \\

\parbox[t]{2mm}{\multirow{1}{*}{\rotatebox[origin=c]{90}{S5\hspace{-6em}}}} & \includegraphics{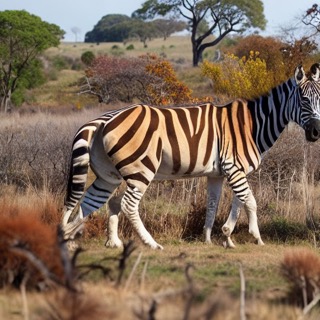} & \includegraphics{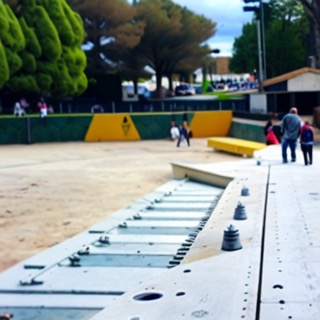} & \includegraphics{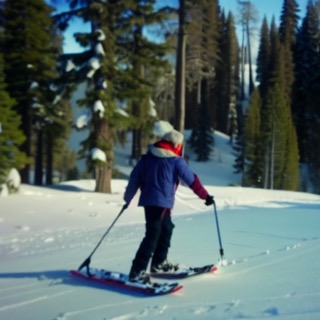}& \includegraphics{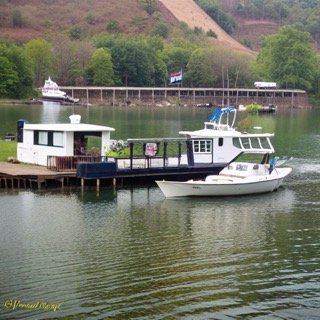} &
\includegraphics{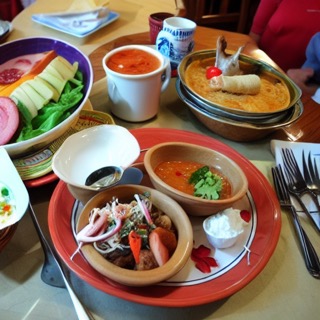} & \includegraphics{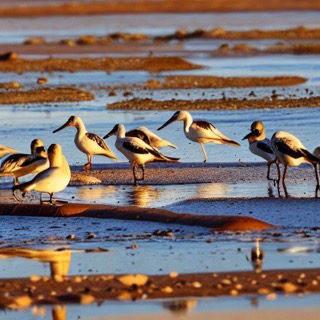} & \includegraphics{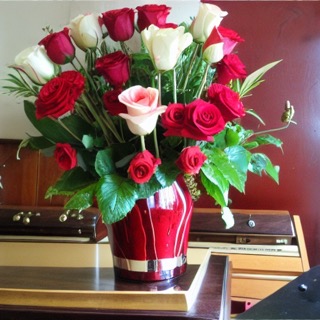} &
\includegraphics{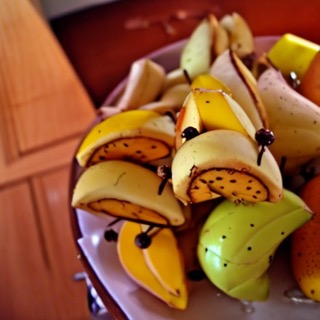} \\
\parbox[t]{2mm}{\multirow{1}{*}{\rotatebox[origin=c]{90}{S7\hspace{-6em}}}} & \includegraphics{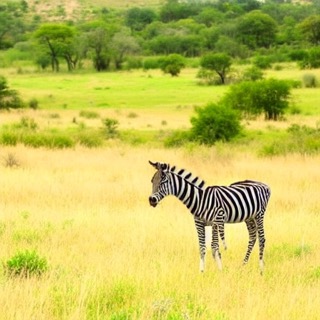} & \includegraphics{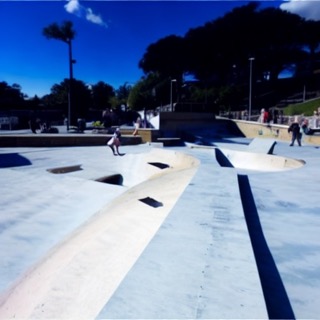} & \includegraphics{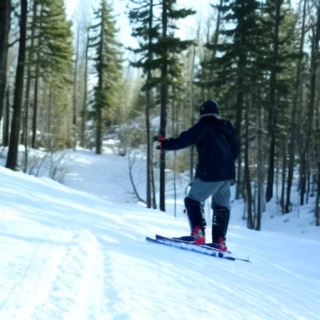}& \includegraphics{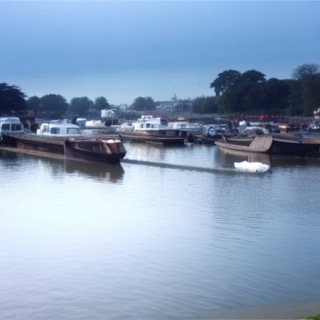} &
\includegraphics{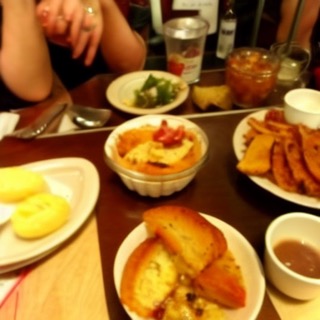} & \includegraphics{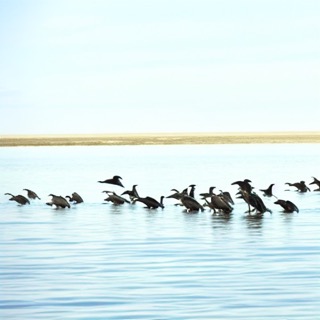} & \includegraphics{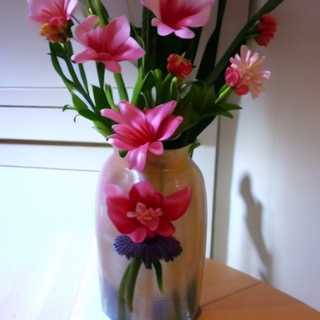} &
\includegraphics{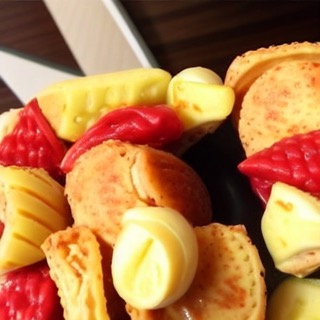} \\
\end{tabular}
}
\caption{\textbf{Subject-Specific Visual Decoding Results.} Qualitative comparison of \method{} reconstruction for the four subjects (S1, S2, S5, and S7) of NSD~\cite{allen2022massive}.
}
\label{fig:supmat_other_subject}
}
\end{figure}

\cref{fig:supmat_other_subject} displays a qualitative comparison of individual subjects, while the quantitative evaluation of \method{} on all four subjects can be found in \cref{tab:supmat_other_subject}.
\revise{\cref{fig:supmat_comparison_vd_compressed} shows comparison on visual decoding between our method and the literature.}

\vspace{-10pt}

\vspace{-10pt}
\begin{table}[!th]
    \centering
    \caption{\textbf{\method{} Visual Decoding for S1 with Various Image Generation Models.} 
    Despite that our method is not specifically designed for fMRI-to-image reconstruction (visual decoding), the predicted textual and visual outputs can serve as cues for the final image reconstruction using a variety of pretrained image generation models. These models include text-to-image SD~\cite{rombach2022high}, SD-XL~\cite{podell2023sdxl}, and Kandinsky~\cite{arkhipkin2023kandinsky}, a layout-to-image GlIGEN~\cite{li2023gligen}, and a multiple-condition Versatile Diffusion (VD)~\cite{xu2022versatile}. We further analyze the effects of VD's inputs in the last three rows. The colors represent the \tgold{best}, \tsilver{second-best}, and \tbronze{third-best} performance.
    }
\label{tab:supmat_visual_decoding_different_model}
    \resizebox{\textwidth}{!}{
    \begin{tabular}{l|cccc|cccc|ccc}
    \toprule
        \multirow{2}{*}{Image Generation}  & \multicolumn{4}{c|}{Low-Level} & \multicolumn{4}{c|}{High-Level} & \multirow{2}{*}{FID $\downarrow$} & \multirow{2}{*}{CFID $\downarrow$} & \multirow{2}{*}{LPIPS $\downarrow$}  \\
	  ~ & PixCorr $\uparrow$ & SSIM $\uparrow$ & AlexNet(2) $\uparrow$ & AlexNet(5) $\uparrow$ & Inception $\uparrow$ & CLIP $\uparrow$  & EffNet-B $\downarrow$ & SwAV $\downarrow$ & ~ & ~ & ~ \\ 
        \midrule
        SD~\cite{rombach2022high} & - & .292 & 71.7\% & 83.5\% & 85.6\% & 86.1\% & .786 & .535  & 95.62 & 89.04 & 0.79\\ 
        SDXL~\cite{podell2023sdxl} & .070 & \silver{.336} & 73.6\% & \bronze{86.9\%} & \silver{87.2\%} & 86.3\% & \bronze{.769} & \bronze{.475}  & \bronze{85.67} & \bronze{82.41} & \silver{0.76} \\ 
        Kandinsky3~\cite{arkhipkin2023kandinsky} & \silver{.110} & \bronze{.328} & \bronze{75.9\%} & 85.4\% & 85.8\% & \bronze{86.4\%} & .789 & .514  &  89.98 & 88.01   & 0.78\\ 
        GLIGEN~\cite{li2023gligen} & \bronze{.078} & .255 & \silver{78.6\%} & \silver{90.1\%} & \bronze{86.5\%} & \silver{87.5\%} & \silver{.766} & \silver{.473}  & \silver{73.70} & \silver{67.21} & \bronze{0.77} \\ 
        VD~\cite{xu2022versatile} & \gold{.293} & \gold{.345} & \gold{95.8\%} & \gold{97.2\%} & \gold{92.6\%} & \gold{93.9\%} & \gold{.690} & \gold{.391}  & \gold{67.47} & \gold{63.81} & \gold{0.73} \\ 
        \midrule
        VD-1 (GT-ti-z) & \gold{.641} & \gold{.402} & \gold{99.9996\%} & \gold{99.9985\%} & \gold{99.5\%} & \gold{99.98\%} & \gold{.390} & \gold{.187}  & \gold{54.33} & \gold{44.74} & \gold{0.53} \\ 
        VD-2 (GT-ti-rand-z) & \bronze{.098} & \bronze{.244} & \bronze{91.0\%} & \bronze{98.9\%} & \bronze{98.8\%} & \silver{99.93\%} & \bronze{.504} & \silver{.261}  & \silver{59.12} & \bronze{53.30} & \silver{0.69} \\ 
        VD-3 (GT-ti-pred-z) & \silver{.327} & \silver{.352} & \silver{98.975\%} & \silver{99.705\%} & \silver{98.9\%} & \bronze{99.90\%} & \silver{.497} & \silver{.261}  & \bronze{62.96} & \silver{52.48}  & \bronze{0.71} \\ 
        \bottomrule
    \end{tabular}
    }
\end{table}

\vspace{-2.5mm}

\begin{table}[!th]
	\centering
	\caption{\textbf{Subject-Specific Visual Decoding Evaluation.} Quantitative evaluation of the \method{} reconstruction for the four subjects (S1, S2, S5, and S7) of NSD~\cite{allen2022massive}.}
	\label{tab:supmat_other_subject}
	\resizebox{\linewidth}{!}
        {
		\begin{tabular}{c|cccc|cccc|ccc}
			\toprule
			\multirow{2}{*}{Subject}  & \multicolumn{4}{c|}{Low-Level} & \multicolumn{4}{c|}{High-Level} & \multirow{2}{*}{FID $\downarrow$} & \multirow{2}{*}{CFID $\downarrow$} & \multirow{2}{*}{LPIPS $\downarrow$} \\
			~ & PixCorr $\uparrow$ & SSIM $\uparrow$ & AlexNet(2) $\uparrow$ & AlexNet(5) $\uparrow$ & Inception $\uparrow$ & CLIP $\uparrow$  & EffNet-B $\downarrow$ & SwAV $\downarrow$ & ~ & ~ & ~ \\
			\midrule
			S1 & .293 & .345 & 95.8\% & 97.2\% & 92.6\% & 93.9\% & .690 & .391  & 67.47 & 63.81 & 0.73 \\
			S2 & .283 & .353 & 96.2\% & 97.3\% & 90.8\% & 93.0\% &.705 & .396 & 70.00 & 66.78 & 0.74 \\
			S5 & .277 & .337 & 95.8\% & 97.7\% & 93.7\% & 94.8\% & .689 & .380 & 67.70 & 64.52 & 0.74  \\
			S7 & .279  & .329 & 95.0\% & 96.9\% & 90.5\% & 92.1\% & .713 & .405 & 70.07 & 66.50  & 0.75  \\
			\bottomrule
		\end{tabular}
	}
\end{table}
\vspace{-2.5mm}

\vspace{-2.5mm}
\section{Additional Analyses}
\label{sec:supmat_analysis}

Here, we provide a couple of additional studies to further evaluate the per-subject performance (\cref{sec:supmat_analysis_subj}), the weakly-supervised adaptation (\cref{sec:supmat_analysis_weak}). 
\revise{We analyze different sampling strategies in cross-subject training (\cref{subsec:sampling_strategies}) and present a joint grounding-decoding evaluation (\cref{subsec:groundingdecod}).}
We also build on MLLMs capacities to explore other tasks (\cref{subsec:other_tasks}) and finally, demonstrate the model-agnostic characteristics of our method (\cref{sec:supmat_analysis_llava}).

\subsection{Subject-Specific Analysis}
\label{sec:supmat_analysis_subj}

In the main paper, we primarily present the results from two types of our models: the cross-subject training model \method{} and the single-subject training model \methodSubject{Sx}. The former is trained using all training data from subjects \{S1,~S2,~S5,~S7\}, whereas \methodSubject{Sx} trains only on data from a specific subject.

Instead, here we explore training the cross-subject models with various combinations of different subjects (\eg, training on \{S1,~S2,~S5\} and testing on S1). These pretrained models are then utilized in the weakly-supervised subject adaptation experiments. The corresponding results are detailed in \cref{tab:supmat_training_sub_combination}. 
It is interesting to note that while training with more than one subject is always improving performance, the best performance are not always achieved when using all subjects available. An example is the S7 evaluation which performs better when \method{} is trained only on \{S1,~S5,~S7\} (\ie, not using S2 data). We conjecture this could relate to some subjects having more similar brain activities patterns than others. 

\begin{table}[t]
\setlength{\fboxrule}{0pt} 
\setlength{\fboxsep}{2pt}
\setlength{\tabcolsep}{4pt}
\caption{\textbf{\method{} Subject Analysis.} {We report the per-subject evaluation of \method{} when varying the set of training subjects (col `Train').}
For example, `S1' denotes our method is trained only on data from S1, \textit{i.e.} this corresponds to \methodSubject{S1}. Similarly, `S1,S2,S5,S7' means training using data from all subjects, and corresponding to the model \method{}.
`S1,S2,S5' means the model is trained with samples from \{S1,~S2,~S5\}. 
`Eval' means evaluation on a certain subject.
{We note that better performance are almost always achieved when training on more than one subject.} 
The colors represent the \tgold{best}, \tsilver{second-best}, and \tbronze{third-best} performance. 
}
\vspace{-2.5mm}
\label{tab:supmat_training_sub_combination}
\centering
\resizebox{\textwidth}{!}
{
\begin{tabular}{@{}cc|cccccc|cccc@{}}
\toprule
\multicolumn{2}{c|}{\method{} setting} & \multicolumn{6}{c|}{Captioning} & \multicolumn{4}{c}{Grounding} \\
Train & Eval &  BLEU1  & ROUGE & CIDEr & SPICE & CLIP-S & RefCLIP-S  &  acc@0.5 (A)       & IoU (A)      & acc@0.5 (S)       & IoU (S)  \\
\midrule
 S1 & \multirow{5}{*}{S1} & 57.63 & 42.15 & 51.93 & 11.83 & \silver{66.44} & 72.12 & 13.72 & 17.56 & 21.52 & 25.14\\
 S1,S2,S5 & ~ & \gold{59.75} & \bronze{43.53} & \silver{57.36} & \gold{12.79} & \bronze{66.39} & \silver{72.63} & \silver{15.63} & \bronze{19.30} & \bronze{23.60} & \bronze{27.15}  \\
 S1,S2,S7 & ~ & {58.31} & 42.67 & {55.03} & 12.20 & 65.72 & 71.95 & 14.76 & 18.76 & 22.93 & 26.58 \\
 S1,S5,S7 & ~ & \bronze{59.23} & \silver{43.70} & \bronze{57.00} & \bronze{12.69} & {66.25} & \bronze{72.42} & \bronze{15.58} & \silver{19.34} & \silver{24.05} & \silver{27.58}  \\
  S1,S2,S5,S7 & ~ & \silver{59.44} & \gold{43.71} & \gold{61.06} & \gold{12.79} & \gold{67.78} & \gold{73.54} & \gold{18.93} & \gold{21.28} & \gold{30.23} & \gold{30.18} \\
\midrule
 S2 & \multirow{5}{*}{S2} & 57.18 & 41.85 & 50.62 & 11.50 & 64.87 & 71.06 & 15.21 & 18.68 & 23.60  & 26.59 \\
 S1,S2,S5 & ~ & \silver{57.91} & \silver{42.43} & \silver{52.80} & \silver{11.91} & \silver{65.37} & \silver{71.57} & \silver{16.04} & \silver{19.29} & \silver{25.09} & \bronze{27.43} \\
 S1,S2,S7 & ~ & 57.30 & 41.98 & 51.17 & 11.42 & 64.53 & 70.85 & \bronze{15.42} & \bronze{19.27} & \bronze{24.05} & \silver{27.60} \\
 S2,S5,S7 & ~ & \bronze{57.69} & \bronze{42.29} & \bronze{51.77} & \bronze{11.72} & \bronze{65.08} & \bronze{71.27} & 15.25 & 18.95 & 23.08 & 26.70 \\
  S1,S2,S5,S7 & ~ & \gold{59.37} & \gold{43.86} & \gold{55.93} & \gold{12.08} & \gold{66.46} & \gold{72.36} & \gold{18.27} & \gold{20.77} & \gold{28.22} & \gold{29.19} \\
\midrule
 S5 & \multirow{5}{*}{S5} & 58.99 & 43.30 & 57.09 & 12.70 & 66.48 & 72.69 & 14.72 & 18.45 & 22.93 & 26.34 \\
 S1,S2,S5 & ~ & \bronze{59.63} & {43.32} & \silver{60.00} & \silver{13.25} & \bronze{67.10} & 73.16 & \bronze{15.01} & \silver{18.90} & \silver{23.60} & \silver{26.97} \\
 S1,S5,S7 & ~ & \silver{60.02} & \silver{43.50} & \bronze{59.67} & \gold{13.31} & \silver{67.24} & \silver{73.39} & 14.84 & 18.68 & 22.86 & 26.34 \\
 S2,S5,S7 & ~ & 59.43 & \bronze{43.45} & 59.22 & 12.71 & \bronze{67.10} & \bronze{73.24} & \silver{15.05} & \bronze{18.82} & \bronze{23.16} & \bronze{26.52} \\
  S1,S2,S5,S7 & ~ & \gold{60.36} & \gold{44.81} & \gold{61.32} & \bronze{13.19} & \gold{68.39} & \gold{74.11} & \gold{18.19} & \gold{20.85} & \gold{28.74} & \gold{30.02} \\ 
 \midrule
 S7 & \multirow{5}{*}{S7} & 55.71 & 40.64 & 47.07 & 11.26 & 63.66 & 70.09 & 13.60 & 17.83 & 21.07 & 25.19 \\
 S1,S2,S7 & ~ & \bronze{56.72} & 41.43 &  49.78 & 11.37 & 64.21 & 70.62 & \bronze{14.43} & \bronze{18.26} & 21.82 & 25.64  \\
 S1,S5,S7 & ~ & \gold{57.83} & \silver{42.09} & \gold{53.53} & \gold{11.88} & \silver{64.92} & \silver{71.29} & \silver{14.76} & \silver{18.68} & \silver{22.49} & \silver{26.26}  \\
 S2,S5,S7 & ~ & 56.29 & \bronze{41.64} & \bronze{51.23} & \bronze{11.52} & \bronze{64.46} & \bronze{70.84} & \bronze{14.43} & {18.22} & \bronze{22.11} & \bronze{25.68} \\
 S1,S2,S5,S7 & ~ & \silver{57.20} & \gold{42.16} & \silver{52.73} & \silver{11.63} & \gold{65.90} & \gold{71.83} & \gold{16.74} & \gold{19.63} & \gold{25.69} & \gold{27.90} \\
\bottomrule
\end{tabular}
}
\end{table}

\subsection{Weakly-Supervised Adaptation}
\label{sec:supmat_analysis_weak}

As explained in the main paper, our method enables weakly-supervised subject adaptation and can train a model for a new subject in a data-efficient manner. 
{In other words, \method{} can accommodate a new subject with only a portion of the total training data. This is crucial considering the challenges in obtaining brain modality data. In the main paper, we presented the results of pretraining on \{S1, S2, S5\} and adapting to S7 by finetuning with varying amounts \{5\%, 10\%, 20\%, 30\%, 50\%, 80\%, 100\%\} of S7 training data.}

\begin{figure}
    \centering
	\begin{subfigure}{1.0\linewidth}
		    \resizebox{1.0\linewidth}{!}{%
			\setlength{\tabcolsep}{2pt}%
			\begin{tabular}{ccccc}
				\multicolumn{3}{c}{\scriptsize\textbf{Captioning}} & \multicolumn{2}{c}{\scriptsize\textbf{Grounding}}\\
				\cmidrule(lr){1-3}\cmidrule(lr){4-5}
				\begin{subfigure}[b]{0.18\textwidth}
					\includegraphics[width=\textwidth]{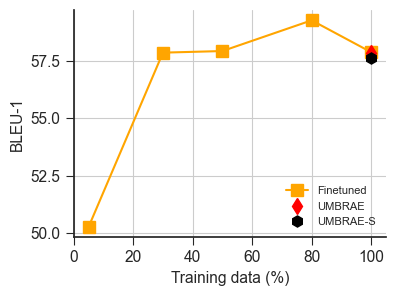}
				\end{subfigure}&%
				\begin{subfigure}[b]{0.18\textwidth}
					\includegraphics[width=\textwidth]{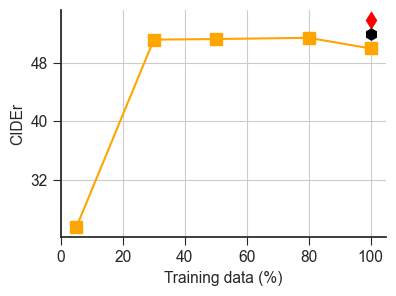}
				\end{subfigure}&
				\begin{subfigure}[b]{0.18\textwidth}
					\includegraphics[width=\textwidth]{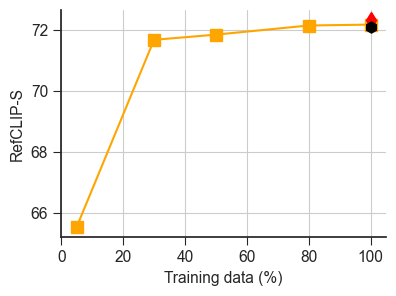}
				\end{subfigure}&%
				\begin{subfigure}[b]{0.18\textwidth}
					\includegraphics[width=\textwidth]{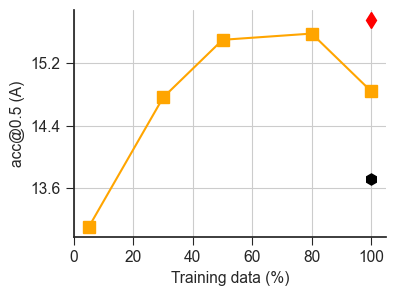}
				\end{subfigure}&%
				\begin{subfigure}[b]{0.18\textwidth}
					\includegraphics[width=\textwidth]{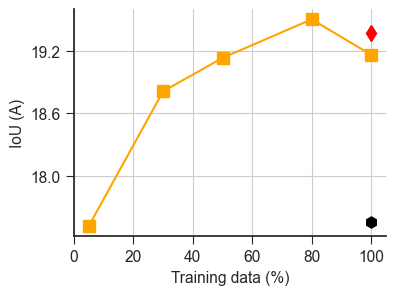}
				\end{subfigure}\\%
				\begin{subfigure}[b]{0.18\textwidth}
					\includegraphics[width=\textwidth]{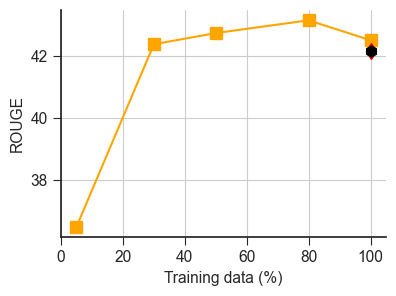}
				\end{subfigure}&%
				\begin{subfigure}[b]{0.18\textwidth}
					\includegraphics[width=\textwidth]{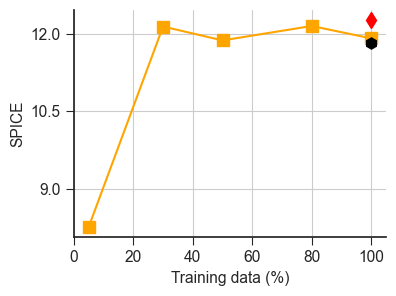}
				\end{subfigure}&%
				\begin{subfigure}[b]{0.18\textwidth}
					\includegraphics[width=\textwidth]{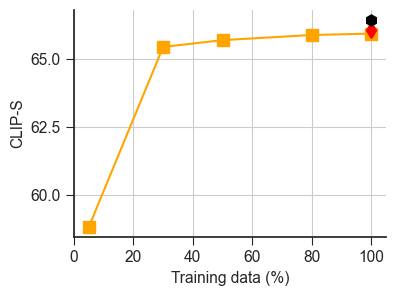}
				\end{subfigure}&%
				\begin{subfigure}[b]{0.18\textwidth}
					\includegraphics[width=\textwidth]{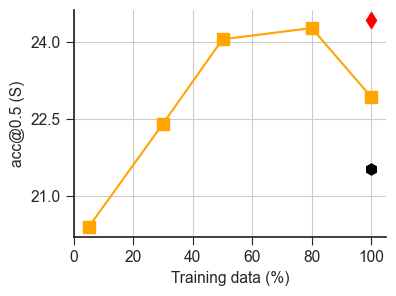}
				\end{subfigure}&%
				\begin{subfigure}[b]{0.18\textwidth}
					\includegraphics[width=\textwidth]{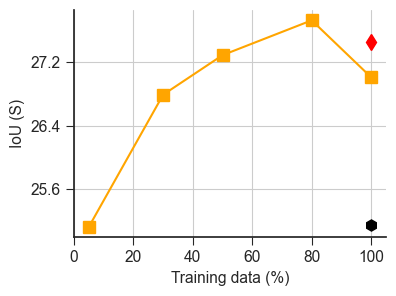}
				\end{subfigure}
			\end{tabular}%
		}
		\caption{Adaptation to S1 (trained on S2, S5, and S7)}
		\label{fig:few_shot_s1}
	\end{subfigure}
	\begin{subfigure}{1.0\linewidth}
		\resizebox{1.0\linewidth}{!}{%
			\setlength{\tabcolsep}{2pt}%
			\begin{tabular}{ccccc}
				\cmidrule(lr){1-3}\cmidrule(lr){4-5}
				\begin{subfigure}[b]{0.18\textwidth}
					\includegraphics[width=\textwidth]{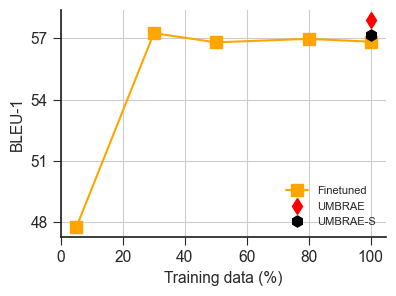}
				\end{subfigure}&%
				\begin{subfigure}[b]{0.18\textwidth}
					\includegraphics[width=\textwidth]{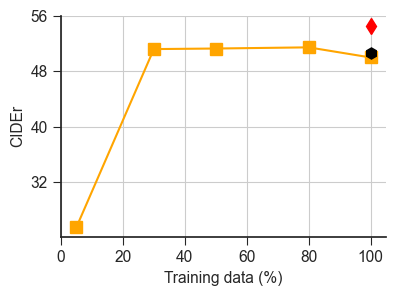}
				\end{subfigure}&
				\begin{subfigure}[b]{0.18\textwidth}
					\includegraphics[width=\textwidth]{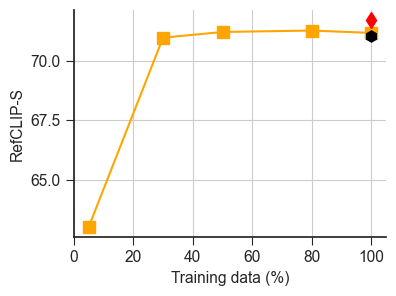}
				\end{subfigure}&%
				\begin{subfigure}[b]{0.18\textwidth}
					\includegraphics[width=\textwidth]{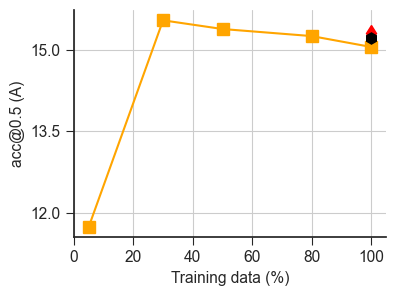}
				\end{subfigure}&%
				\begin{subfigure}[b]{0.18\textwidth}
					\includegraphics[width=\textwidth]{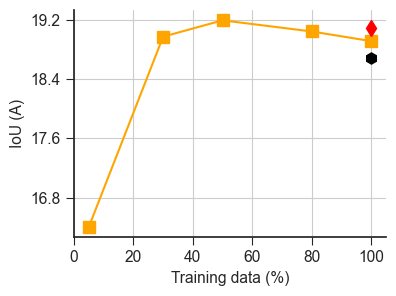}
				\end{subfigure}\\%
				\begin{subfigure}[b]{0.18\textwidth}
					\includegraphics[width=\textwidth]{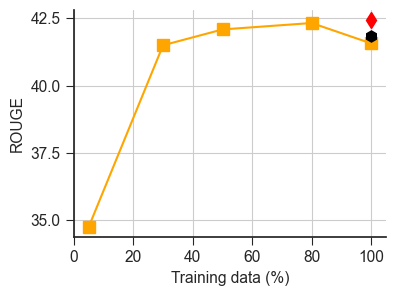}
				\end{subfigure}&%
				\begin{subfigure}[b]{0.18\textwidth}
					\includegraphics[width=\textwidth]{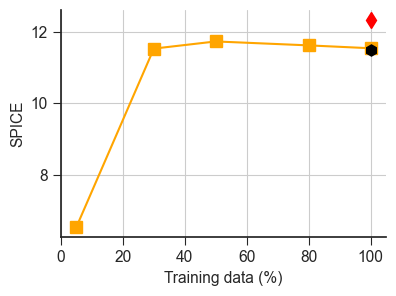}
				\end{subfigure}&%
				\begin{subfigure}[b]{0.18\textwidth}
					\includegraphics[width=\textwidth]{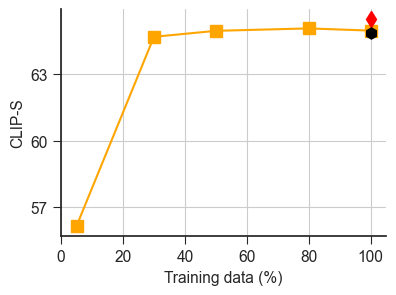}
				\end{subfigure}&%
				\begin{subfigure}[b]{0.18\textwidth}
					\includegraphics[width=\textwidth]{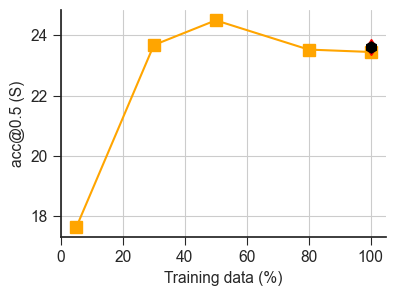}
				\end{subfigure}&%
				\begin{subfigure}[b]{0.18\textwidth}
					\includegraphics[width=\textwidth]{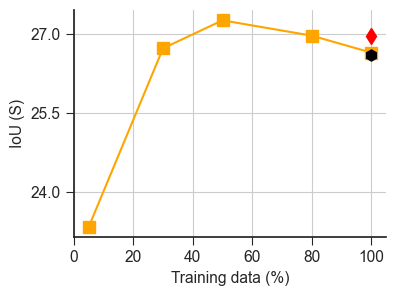}
				\end{subfigure}
			\end{tabular}%
		}
		\caption{Adaptation to S2 (trained on S1, S5, and S7)}
		\label{fig:few_shot_s2}
	\end{subfigure}
	\begin{subfigure}{1.0\linewidth}
		\resizebox{1.0\linewidth}{!}{%
			\setlength{\tabcolsep}{2pt}%
			\begin{tabular}{ccccc}
				\cmidrule(lr){1-3}\cmidrule(lr){4-5}
				\begin{subfigure}[b]{0.18\textwidth}
					\includegraphics[width=\textwidth]{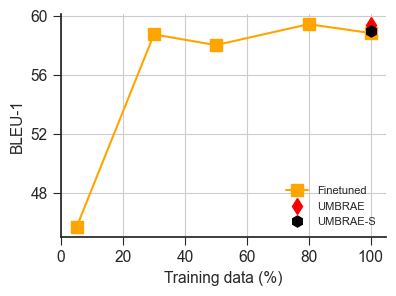}
				\end{subfigure}&%
				\begin{subfigure}[b]{0.18\textwidth}
					\includegraphics[width=\textwidth]{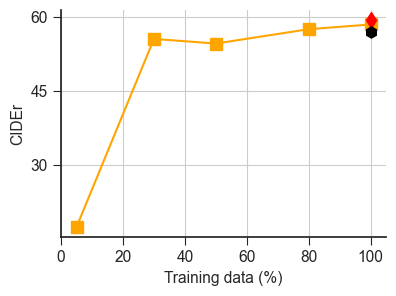}
				\end{subfigure}&
				\begin{subfigure}[b]{0.18\textwidth}
					\includegraphics[width=\textwidth]{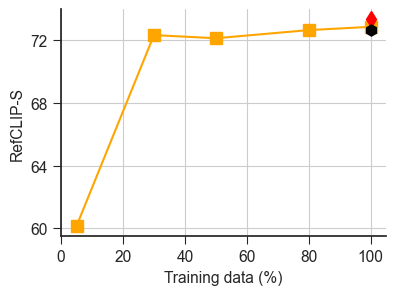}
				\end{subfigure}&%
				\begin{subfigure}[b]{0.18\textwidth}
					\includegraphics[width=\textwidth]{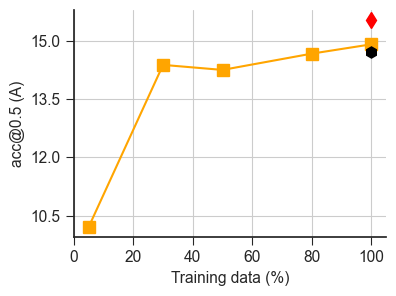}
				\end{subfigure}&%
				\begin{subfigure}[b]{0.18\textwidth}
					\includegraphics[width=\textwidth]{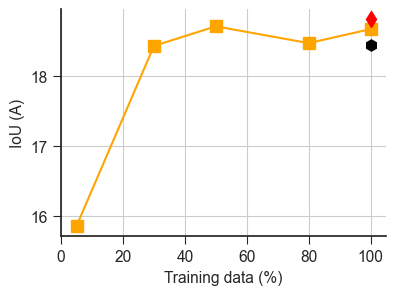}
				\end{subfigure}\\%
				\begin{subfigure}[b]{0.18\textwidth}
					\includegraphics[width=\textwidth]{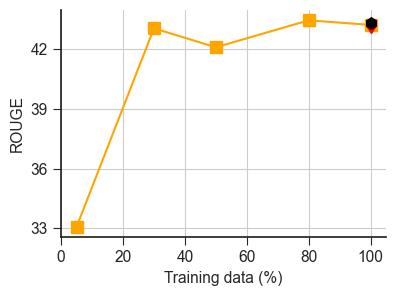}
				\end{subfigure}&%
				\begin{subfigure}[b]{0.18\textwidth}
					\includegraphics[width=\textwidth]{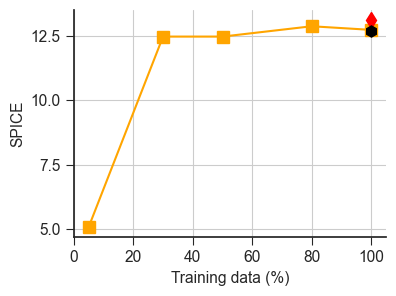}
				\end{subfigure}&%
				\begin{subfigure}[b]{0.18\textwidth}
					\includegraphics[width=\textwidth]{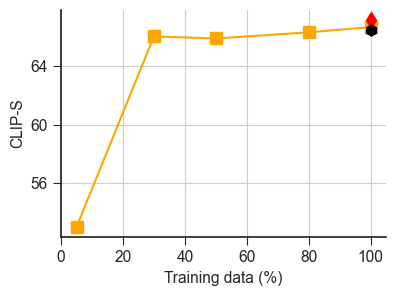}
				\end{subfigure}&%
				\begin{subfigure}[b]{0.18\textwidth}
					\includegraphics[width=\textwidth]{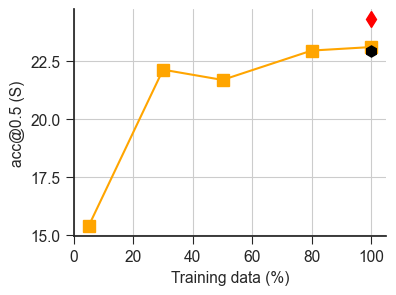}
				\end{subfigure}&%
				\begin{subfigure}[b]{0.18\textwidth}
					\includegraphics[width=\textwidth]{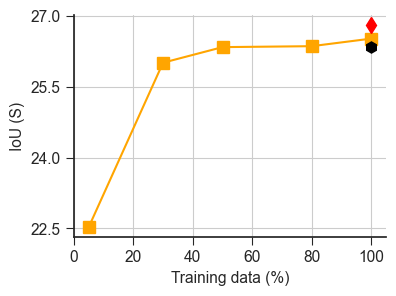}
				\end{subfigure}
			\end{tabular}%
		}
		\caption{Adaptation to S5 (trained on S1, S2, and S7)}
		\label{fig:few_shot_s5}
	\end{subfigure}
	\caption{\textbf{Weakly-Supervised Subject Adaptation}. We plot our performance for adaptation to a given subject (S1 in \subref{fig:few_shot_s1}, S2 in \subref{fig:few_shot_s2}, S5 in \subref{fig:few_shot_s5}). For adapting to a given subject Sx, we first train \method{} on all other subjects and then finetune the pretrained model with varying ratios of training data from the target subject.
	}
    \label{fig:supmat_few_shot}
\end{figure}

{In~\cref{fig:supmat_few_shot}, we present the adaptation to other subjects, being: \textbf{\cref{fig:few_shot_s1}} training on \{S2,~S5,~S7\}, adaptation to S1; \textbf{\cref{fig:few_shot_s2}} training on \{S1,~S5,~S7\}, adaptation to S2; \textbf{\cref{fig:few_shot_s5}} training on \{S1,~S2,~S7\}, adaptation to S5. 
In the above mentioned figures, we report performance when finetuning both the tokenizer and the encoder, which was proven to provide the best performance in the main paper. 
As for adaptation to S7 (cf. main paper), compared to the single-subject model `\methodSubject{Sx}', our `Finetuned' adaptation on S1, S2, and S5 achieves comparable performance using only {30\%} of the data.}

\subsection{\revise{Sampling Strategies in Cross-Subject Training}}
\label{subsec:sampling_strategies}

\begin{figure}[tp]
  \centering
  \begin{minipage}{0.28\linewidth}
    \centering
    \includegraphics[width=\linewidth]{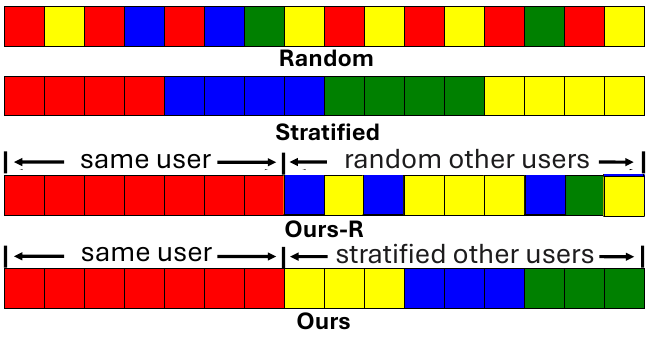}
  \end{minipage}
  \begin{minipage}{0.70\linewidth}
    \centering
    \resizebox{\textwidth}{!}{
    \setlength{\tabcolsep}{0.3em}
    \renewcommand{\arraystretch}{1.0}
    \begin{tabular}{@{}l|cccc|cccc@{}}
    \toprule
    Sampling  & \multicolumn{4}{c|}{Captioning} & \multicolumn{4}{c}{Grounding} \\ 
    ~ & BLEU1  & CIDEr & METER & RefCLIPS & acc@0.5 (A) & IoU (A) & acc@0.5 (S) & IoU (S) \\
    \midrule
    Random & 51.84 & 35.37 & 15.52 & 67.26 & 13.49 & 17.39 & 21.56 & 25.39 \\
    Stratified & 58.91 & 55.83 & 18.94 & 72.69 & 17.31 & 20.34 & 27.08 & 29.03 \\
    Ours-R  & 58.02 & 55.02 & 18.68 & 72.09 & 15.22 & 18.84 & 23.58 & 26.74 \\
    Ours & \textbf{59.09} & \textbf{57.76} & \textbf{19.24} & \textbf{72.96} & \textbf{18.03} &\textbf{20.63} & \textbf{28.22} & \textbf{29.32} \\
    \bottomrule
    \end{tabular}
    }
    \end{minipage}
    \captionof{table}{\revise{\textbf{Comparison of Sampling Strategies.}
    `Random' means all subjects are randomly sampled, while `Stratified' ensures that data from the four subjects are equal in number within a batch. `Ours-R' and `Ours' are the same when sampling from the selected dominant subject but differ for the remaining three subjects.}
    }
    \label{tab:sampling_strategy}
\end{figure}

\revise{There are different sampling strategies of subjects and data samples in the cross-subject training.
Our batch of $B$ samples is made of $\theta\!\times{}\!B$ samples from the \textit{same user} chosen according to users frequencies, while the remaining samples are then sampled from \textit{other users}. 
This is illustrated on the left of \cref{tab:sampling_strategy} using, for simplicity, four users, $\theta\!=\!0.44$ and $B\!=\!16$.
`Random' means all subjects are randomly sampled, while `Stratified' ensures that data samples from the four subjects are equal in number within a batch. `Ours-R' and `Ours' are the same for the dominant subject but differ in the sampling strategies for the remaining three subjects.
Using a dominant subject per batch helps the model to learn intra-subject variations while being exposed to other subjects patterns to enhance inter-subject discrimination and alleviate catastrophic forgetting.
\cref{tab:sampling_strategy} reports average metrics across users for `Random', `Stratified', `Ours-R', and `Ours' (using $\theta\!=\!0.50$, $B\!=\!256$).
Ours outperforms all other sampling strategies.}

\subsection{\revise{Joint Grounding-Decoding Evaluation}}
\label{subsec:groundingdecod}

\begin{figure}[thbp]
  \centering
  \begin{subfigure}{\linewidth}
    \centering
    \includegraphics[width=0.98\linewidth]{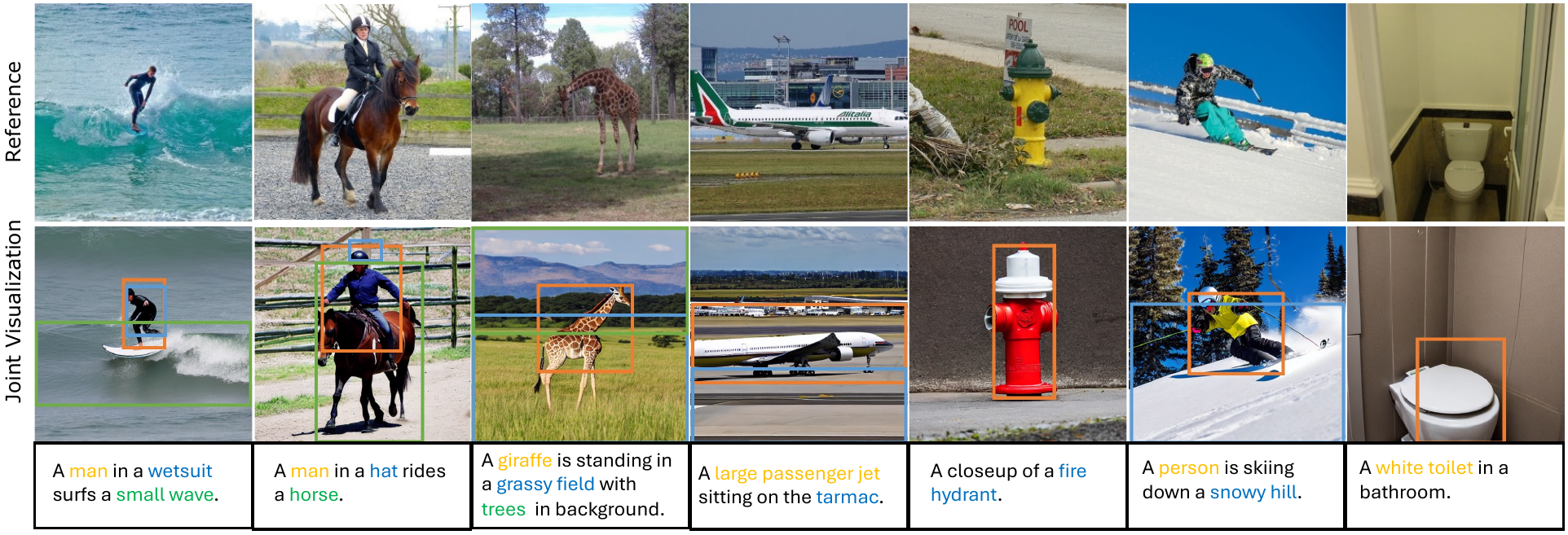}
  \end{subfigure}
  \caption{\revise{\textbf{Joint Visualization of Grounding and Reconstruction.} 
  The first row presents reference images. 
  The second row displays reconstructed images generated using decoded texts and groundings from the third row. 
  The coordinates in the third row for each mentioned object are omitted and visualized in color in the second row.
  Reconstruction may be inaccurate even with correct bounding boxes.}
  }
  \label{fig:groundingdecod}
\end{figure}

\revise{We visualize grounding results on the reference images (ground truth) to better assess their performance. \cref{fig:groundingdecod} further shows grounding and reconstruction simultaneously to highlight their synergy. Results demonstrate that the two tasks are correlated. In some cases, although grounding is correct, reconstruction is inaccurate (\eg~surfer, giraffe, and skier are well located but misorriented).
The first row presents reference images. 
The second row displays reconstructed images, which are generated using the decoded texts and groundings from the third row as inputs. 
The third row illustrates the spotting captioning results, where the coordinates for each mentioned object are omitted and instead visualized in color within the corresponding generated images shown in the second row.
}

\subsection{Other Supported Tasks}
\label{subsec:other_tasks}

\begin{figure}[!t]
	\centering
	\includegraphics[width=0.95\linewidth]{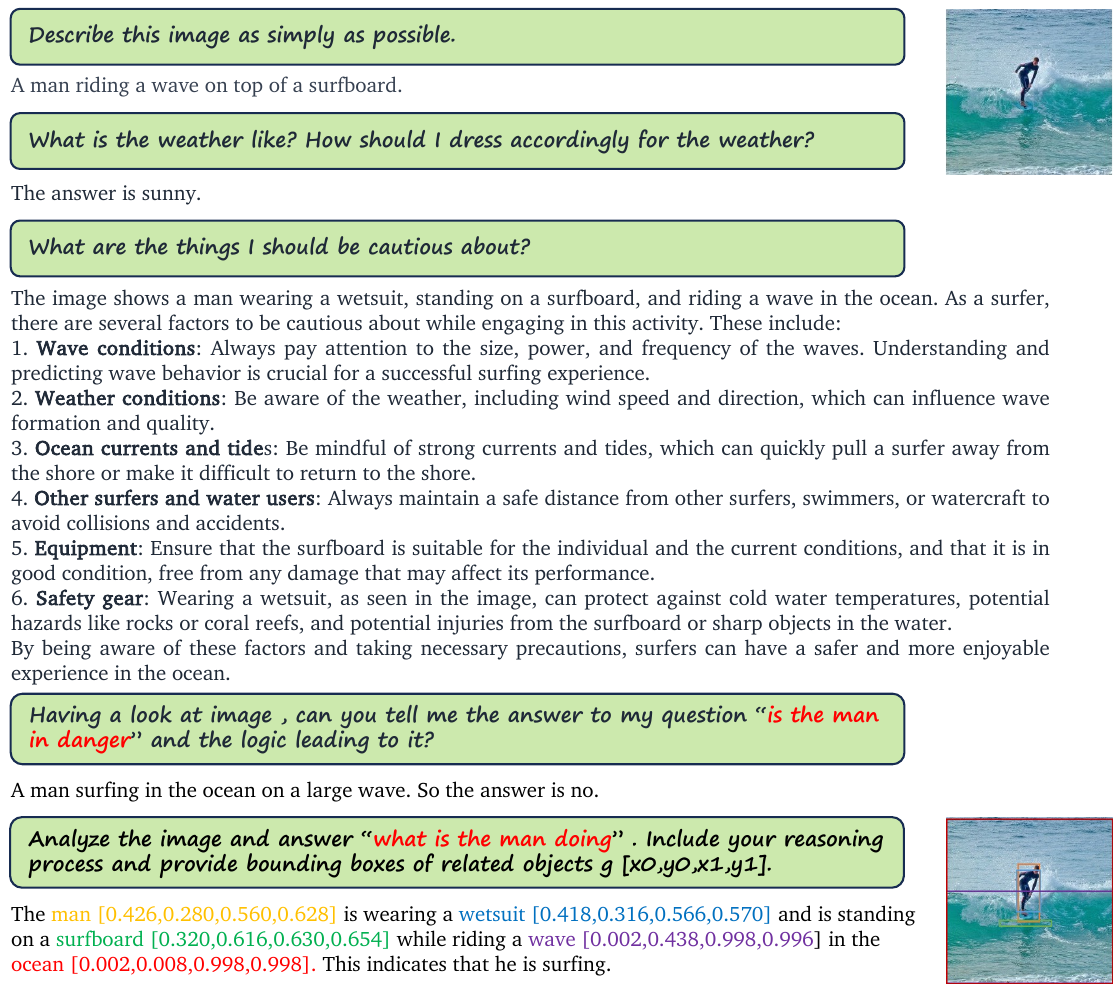}
	\caption{\textbf{Example Results of Other Tasks.} The additional supported tasks listed in~\cref{tab:supmat_prompt} include Q$\rightarrow$A, Q$\rightarrow$CA, and Q$\rightarrow$$\text{C}^\text{Box}$A.
 }
	\label{fig:supmat_other_task}
\end{figure}

As we build on MLLM, we can explore a large variety of tasks. \cref{tab:supmat_prompt} lists the supported tasks, which can be categorized into three groups: captioning, grounding, and QA.
We have presented the brain captioning results in~\cref{subsec:supmat_captioning} and the brain grounding (both REC and Spotting Captioning) results in~\cref{subsec:supmat_grouning}.
This section presents the additional QA tasks, including Q$\rightarrow$A, Q$\rightarrow$CA, and Q$\rightarrow$$\text{C}^\text{Box}$A.
Example results are shown in~\cref{fig:supmat_other_task}.

\subsection{Model-Agnostic Performance}
\label{sec:supmat_analysis_llava}
We employ the pretrained adapter and LLM from Shikra~\cite{chen2023shikra} due to its simultaneous support for both captioning and grounding. 
However, \method{} being model-agnostic, it  can accommodate any image encoders (\eg, CLIP~\cite{radford2021learning} and DINO~\cite{caron2021emerging}), LLMs (\eg, LLaMA~\cite{touvron2023llama} and Vicuna~\cite{chiang2023vicuna}), and MLLMs (\eg, Shikra~\cite{chen2023shikra} and LLaVA~\cite{liu2023visual}) according to specific needs.

\begin{figure}[!t]
	\centering
	\includegraphics[width=0.95\linewidth]{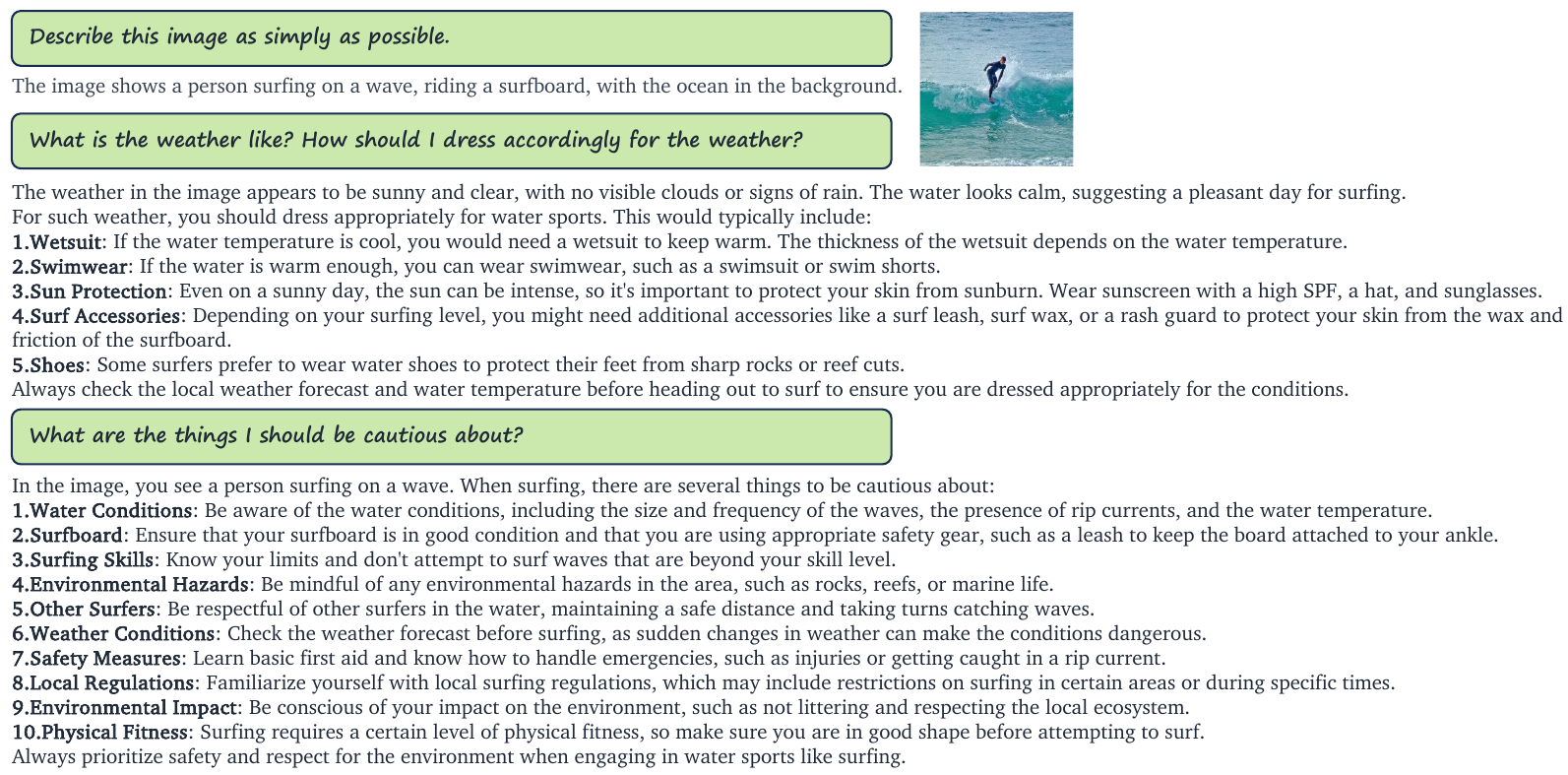}
	\caption{\textbf{\method{} with LLaVA as the MLLM.} Here, we utilize \method{} though, replacing our default use of Shikra~\cite{chen2023shikra}, with LLaVA~\cite{liu2023visual} as the MLLM. The responses show that LLaVA exhibits superior instruction-following capabilities in conversation, detailed description, and complex reasoning compared to Shikra (The corresponding results from Shikra are in~\cref{fig:supmat_other_task}).
 }
	\label{fig:supmat_model_agnostic}
\end{figure}

To demonstrate this capacity, in~\cref{fig:supmat_model_agnostic}, we provide qualitative results of `\method{} with LLaVA' which replaces Shikra with the nonetheless popular LLaVA~\cite{liu2023visual} as the MLLM. These example responses show that LLaVA~\cite{liu2023visual} exhibits superior instruction-following capabilities in conversation, detailed description, and complex reasoning compared to Shikra~\cite{chen2023shikra}. However, LLaVA~\cite{liu2023visual} lacks the capability of spatial localization.

\section{Discussion}
\label{sec:supmat_discussion}

\subsection{Method Limitations}
\label{subsec:limitation}

\begin{figure}[!th]
	\centering
	\includegraphics[width=\linewidth]{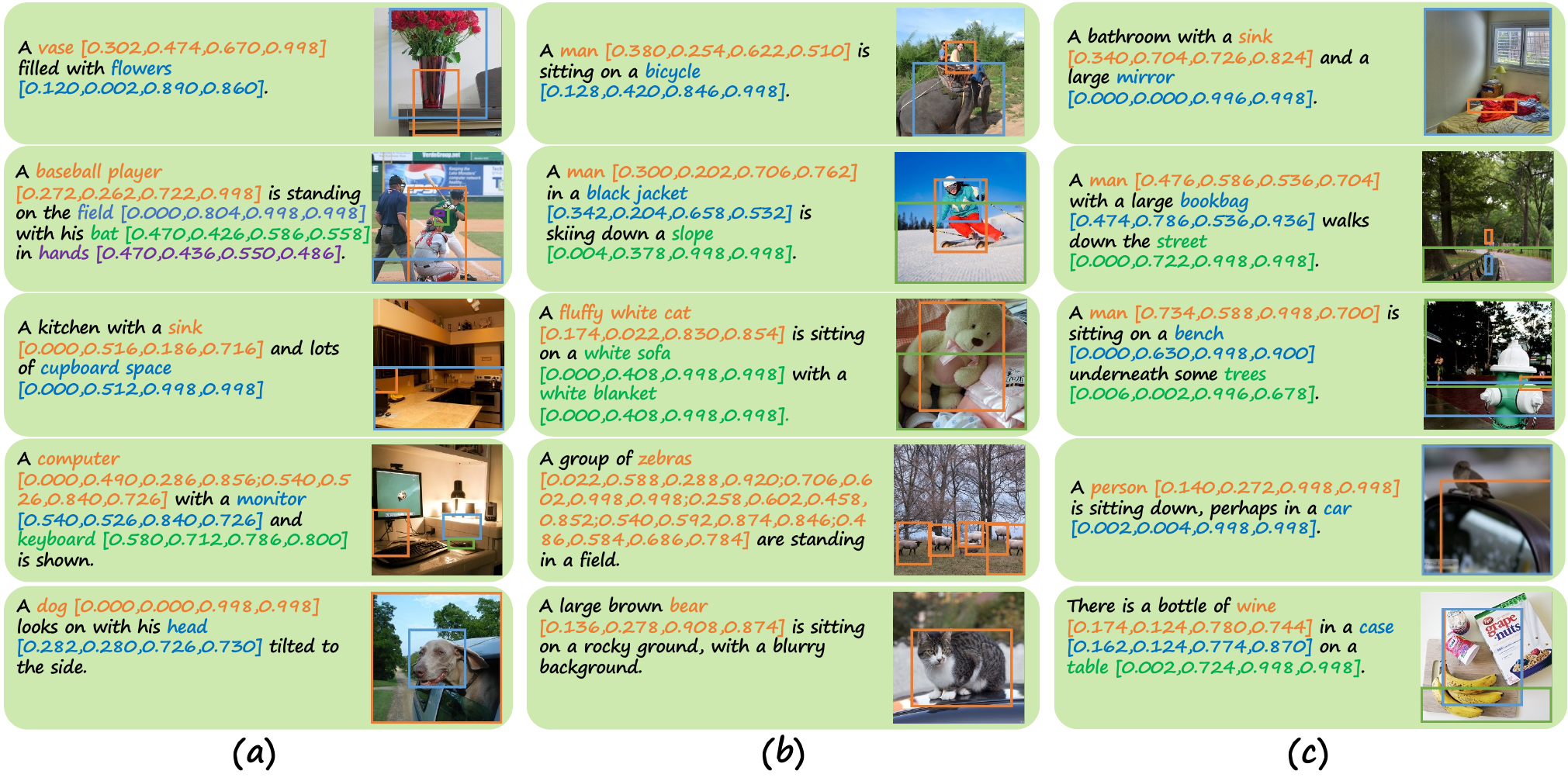}
	\caption{\textbf{Method Limitation.} The failure cases can be categorized into: (a) correct semantic recognition but inaccurate spatial localization; (b) accurate spatial localization but semantic errors; (c) errors in both semantic recognition and spatial localization.
 }
	\label{fig:supmat_limitation}
\end{figure}

Our goal is to decode brain signals into multimodal explanations, with a primary emphasis on semantics and positions, achieving both concept recognition and spatial localization. While results demonstrate decent performance, there are instances where it falls short. 
Apart from previously mentioned challenges with inconspicuous objects, other failure cases can be categorized into three types: (a) correct semantic recognition but inaccurate spatial localization, (b) accurate spatial localization but semantic errors, and (c) errors in both semantic recognition and spatial localization. 
We use spotting captioning to illustrate failure cases in~\cref{fig:supmat_limitation}, as this task outputs both concepts and locations.

Moreover, our method relies on pretrained MLLMs, inheriting their advantages while also facing common shortcomings associated with large models, including biases, hallucinations, generation of inappropriate content, and potential ethical concerns. Our method is also constrained by the quality of the captured brain responses in NSD~\cite{allen2022massive} in two ways. Firstly, there are inherent inaccuracies introduced during data collection. NSD is captured using non-invasive neuroimaging techniques, where participants' compliance is necessary to avoid disruption in decoding caused by head movement or distraction. Secondly, the experimental images are sourced from COCO~\cite{lin2014microsoft}, which limits our method to natural scenes similar to those found in the COCO dataset~\cite{lin2014microsoft}.

\subsection{Potential Negative Impact}
\label{subsec:supmat_negative_impact}

Our method relies on pretrained models as its foundation. While benefiting from the remarkable capabilities provided by LLMs~\cite{chen2023shikra,han2024onellm,touvron2023llama}, they also pose challenges and concerns that prompt broader societal impacts. These include potential biases in the training data, the generation of inaccurate or inappropriate content, and ethical considerations associated with their utilization. The inaccurate interpretation from our method may also lead to misunderstandings about the information contained within brain signals.

\vspace{-0.1in}

\end{appendix}

\bibliographystyle{splncs04}
\bibliography{reference}
\end{document}